%% file: main.tex
\documentclass[3p]{elsarticle}
\biboptions{sort&compress}

\usepackage{amssymb}

\usepackage{amsmath,amsfonts}
\usepackage{graphicx}
\graphicspath{{./fig/}}
\usepackage{subfig}
\usepackage{tikz}
\usetikzlibrary{arrows, shapes, tikzmark, calc, matrix, math, patterns}
\usetikzlibrary{overlay-beamer-styles}
\usetikzlibrary{plotmarks}
\usepackage{pgfplots}
\usepackage{multirow}
\usepackage{array}
\usepackage{bm}
\usepackage{comment} 
\usepackage{dblfloatfix}
\usepackage{markdown}
\usepackage{colortbl}
\usepackage{enumitem}
\usepackage{hyperref}

\newcommand{\nn}{\ensuremath{\mathcal{N}}}

\newcommand{\mfpShort}{MF predictor }
\newcommand{\baseg}{basic genomes }

\newcommand{\axg}{auxiliary genomes }

\newcommand{\basegone}{Basic genome }
\newcommand{\axgone}{Auxiliary genome }

\newcommand{\bg}{\ensuremath{g(\pmb{x})}}
\newcommand{\bx}{\ensuremath{\pmb{x}}}
\newcommand{\bG}{\ensuremath{G(\pmb{x})}}
\newcommand{\hp}{\ensuremath{\pmb{\theta}}}

\newcommand{\revise}[1]{{#1}}

\newcommand{\subc}[1]{({\textit #1})}

\journal{}

\begin{document}

\begin{frontmatter}



\title{Mosaic Flows: A Transferable Deep Learning Framework 
	for Solving PDEs on \revise{Unseen} Domains}


\author[inst1]{Hengjie Wang\fnref{1}}

\affiliation[inst1]{organization={Lawrence Berkeley National Laboratory},
            city={Berkeley},
            state={CA},
            postcode={94720}, 
            country={USA}}

\author[inst2]{Robert Planas\fnref{1}}
\author[inst2]{Aparna Chandramowlishwaran}
\author[inst2]{Ramin Bostanabad\fnref{fn2}}

\affiliation[inst2]{organization={University of California-Irvine},
            city={Irvine},
            state={CA},
            postcode={92697}, 
            country={USA}}
\fntext[fn1]{Equal Contribution}
\fntext[fn2]{Corresponding author. Email: Raminb@uci.edu}

\begin{abstract}
Physics-informed neural networks (PINNs) are increasingly employed to replace/augment traditional numerical methods in solving partial differential equations (PDEs). \revise{While state-of-the-art PINNs have many attractive features, they approximate a specific realization of a PDE system and hence are problem-specific. That is, the model needs to be re-trained each time the boundary conditions (BCs) and domain shape/size change. This limitation prohibits the application of PINNs to realistic or large-scale engineering problems especially since the costs and efforts associated with their training are considerable.}

\revise{We introduce} a transferable framework for solving boundary value problems (BVPs) via deep neural networks which can be trained once and used forever for various \revise{unseen domains and BCs}.
\revise{We first introduce \emph{genomic flow network} (GFNet), a neural network that can infer the solution of a BVP across arbitrary BCs on a small square domain called \emph{genome}.}
Then, we propose \emph{mosaic flow} (MF) predictor, a novel iterative algorithm that assembles the GFNet's inferences for BVPs on large domains with \revise{unseen sizes/shapes and BCs} while preserving the spatial regularity of the solution.
We demonstrate that our framework can estimate the solution of Laplace and Navier-Stokes equations in domains of unseen shapes and \revise{BCs} that are, respectively, $1200$ and $12$ times larger than \revise{the training domains}. 
Since our framework eliminates the need to re-train models for unseen domains and \revise{BCs}, it demonstrates up to 3 orders-of-magnitude speedups compared to the state-of-the-art.
\end{abstract}


\begin{highlights}
\item We introduce a transferable framework for solving boundary value problems (BVPs) via deep neural networks which can be trained once and used forever for various domains of unseen sizes, shapes, and boundary conditions.

\item We introduce \emph{genomic flow network} (GFNet) which can solve a BVP with arbitrary boundary conditions on a square domain. 
We design GFNet's architecture based on the characteristics of the PDE and show that such a design significantly improves accuracy. 
\revise{For instance, GFNet is twice more accurate than the conventional fully connected architecture in approximating the solution of the Laplace equation.
For learning the NS equations, we further enforce the input BC in GFNet and reduce the generalization error by nearly 70\% compared to DeepONet and Fourier Neural Operator (FNO).}

\item We develop \emph{mosaic flow} (MF) predictor, a novel iterative algorithm that assembles GFNet's inferences to predict the solution of a PDE system for domains with unseen sizes, shapes, and boundary conditions. 
We show that MF predictor can scale up GFNet's inferences by $\sim\!\!1200\times$ and $12\times$ in the case of, respectively, Laplace and NS equations.
\revise{Moreover, MF predictor is independent of the neural network model used as GFNet. 
We compare our model with DeepONet and FNO and demonstrate that it reduces the error by nearly 6 times when employed by MF predictor to predict unseen flow features.}

\item With GFNet and MF predictor, we eliminate the need to re-train a neural network for unseen domains or BCs. This transferability delivers 1-3 orders of magnitude speedup compared to the start-of-the-art Physics-Informed neural network (PINN) for solving the Laplace and NS equations while achieving comparable or better accuracy.

\end{highlights}


\end{frontmatter}


\input{intro}    
\input{related}  
\input{DL_model} 
\input{eval}     
\input{conc}     
\input{appendix}

%



 \bibliographystyle{elsarticle-num} 
 \bibliography{main}





\end{document}

%% file: intro.tex
\vspace{-4mm}
\section{Introduction}\label{intro}
Partial differential equations (PDEs) are ubiquitously used in sciences and engineering to model physical phenomena. 
Two notable PDEs that have far-reaching applications are the Navier-Stokes (NS) equations which model the motion of viscous fluids \cite{fox2020fox} and Laplace equation which extensively appears in fluid dynamics, electrostatics, and steady-state heat transfer \cite{RN1440}. 
Solving such PDEs on large domains with arbitrary initial and boundary conditions (ICs and BCs) relies on numerical methods such as finite element (FE) \cite{belytschko2013nonlinear} and finite difference (FD) \cite{Kreyszig2010}. 
While these methods are extremely powerful, they are computationally expensive. 
Because of these high costs, researchers face the challenge of drawing conclusions using a limited number of time-consuming studies. 
For instance, NASA's CFD vision 2030 \cite{Slotnick2014} estimates that achieving a 24-hour turnaround time on a wall-modeled large eddy simulation of a full-wing at Reynolds $Re=10^7$ requires 180 PFlops/s which is only achievable on the most powerful supercomputer today. 
Such high costs make it infeasible to conduct compute-intensive studies such as uncertainty quantification (UQ) or airfoil shape optimization. 
To address this challenge, particularly in large-scale and inverse problems, simulations are augmented with inexpensive surrogates. 

Although a wide range of surrogates such as Gaussian processes (GPs) \cite{kennedy2000predicting, Robert2021, Rasmussen2006} and trees \cite{Alpaydin2014, RN3} are available, most recent applications of surrogate modeling employ deep neural networks (DNNs) \cite{RN1163, RN1267, RN1070, RN248, RN1024, RN1311, RN1434, RN1384, RN1268, RN1419, RN773}. 
The increasing use of DNNs in emulation is largely \revise{because they have extremely high learning capacity} which enables them to distill highly nonlinear and hidden features and relations from a training dataset without explicit instructions. In addition, DNNs are highly scalable and versatile. For instance, while DNNs can naturally build regressors/interpolators from large and high-dimensional data, GPs rely on numerical methods such as low-rank matrix approximation \cite{Rasmussen2006} to handle datasets with more than $\sim5000$ samples \cite{RN940, RN1270}. Or, unlike trees, DNNs can easily handle both classification and regression/interpolation problems with high accuracy \cite{RN1070}.

A major driver in using DNNs for surrogating PDE-governed systems is automatic differentiation (AD) \cite{RN998} which allows \revise{the easy, systematic embedding of a system's governing equations} in a network's training stage to build physics informed neural networks (PINNs) \cite{Raissi2019}. 
This inductive bias not only reduces the reliance on training data obtained from traditional solvers, but also increases the adherence of the DNN to the physics of the problem and dispenses with discretization errors (since AD is exact \cite{RN1441}).
To demonstrate such a physics-informed training process, consider the $2D$ boundary value problem (BVP):
\begin{equation}
  \begin{aligned}
    \nabla^2u(\pmb{x}) &= 0,  \qquad\; \pmb{x}\in\Omega \\
    u(\pmb{x}) &= g(\pmb{x}), \quad \pmb{x}\in\partial\Omega
  \end{aligned}
  \label{eqn:ex_laplace}
\end{equation}
where $\pmb{x}=[x, y]$, $\nabla^2$ is the Laplacian operator ($\partial^2/\partial x^2 + \partial^2/\partial y^2$), $\Omega$ denotes the domain, and $g(\pmb{x})$ is the boundary value function. When training a DNN that \revise{approximates} $u(\pmb{x})$, the loss function can be designed as:
\begin{equation}
  \begin{aligned}
   L(\hp) &= L_1(\hp) + \alpha L_2(\hp) + \beta L_3(\hp), \\
   L_1(\hp) &= \frac{1}{N_1}\sum^{N_1}_{i=1} 
               \left ( u(\pmb{x}_i) - \nn(\pmb{x}_i|\hp) \right )^2, \\
   L_2(\hp) &= \frac{1}{N_2}\sum^{N_2}_{j=1}
                  \left ( \nabla^2\nn(\pmb{x}_i|\hp) \right )^2, \\
   L_3(\hp) &= |\hp|^2 \\
  \end{aligned}
  \label{eqn:ex_pinn_loss}
\end{equation}
where $\hp$ are the parameters of the network (weights and biases), $\nn$ is the DNN \revise{approximation}, $L_1(\hp)$ is the network’s error in predicting the available data on the solution (these data can be on $\partial\Omega$ or in $\Omega$), $L_2(\hp)$ is the residual error which enforces $\nn$ to satisfy the Laplace equation on some randomly selected points (aka collocation points) in $\Omega$, $L_3(\hp)$ denotes Tikhonov regularization that prevents overfitting, and $\alpha$ and $\beta$ are constants that control the contributions of, respectively, $L_2(\hp)$ and $L_3(\hp)$ to $L(\hp)$. $L_2(\hp)$ in Equation \ref{eqn:ex_pinn_loss} involves partial derivatives which can be analytically calculated via AD.

Training a DNN that surrogates the solution of PDEs can be a time-consuming and challenging task primarily because (1) obtaining training data via high-fidelity simulations is expensive, (2) designing an efficient network architecture is an iterative process, (3) optimizing the network’s parameters (such as weights and biases) is demanding, (4) calculating the residual loss is computationally costly because each time a differential operator is applied to the network, it almost doubles the number of computations associated with backpropagation \cite{RN1070, RN1069}, and (5) choosing the optimizer’s parameters (such as batch size and learning rate) as well as balancing the contributions of different loss components to the overall loss (e.g., $\alpha$ and $\beta$ in Equation \ref{eqn:ex_pinn_loss}) are iterative processes. 
These challenges are exacerbated when training DNNs that need many parameters to accurately \revise{approximate} the solution of complex PDEs (such as the NS equations) in large domains.

We argue that the benefits of DNNs that \revise{approximate} PDE-governed systems outweigh the high costs and challenges associated with their training if the DNNs are \textit{transferable}. Unfortunately, existing methods \cite{Jagtap2020, jiang2020meshfreeflownet, RN1188, li2020neural, RN1169} fail in this regard, i.e., there is currently no mechanism for estimating the solution of PDEs with a \textit{pre-trained} DNN.
For example, a PINN that \revise{approximates} $u(\pmb{x})$ in Equation \ref{eqn:ex_laplace} would be largely useless if $\Omega$ and $g(\pmb{x})$ change. 

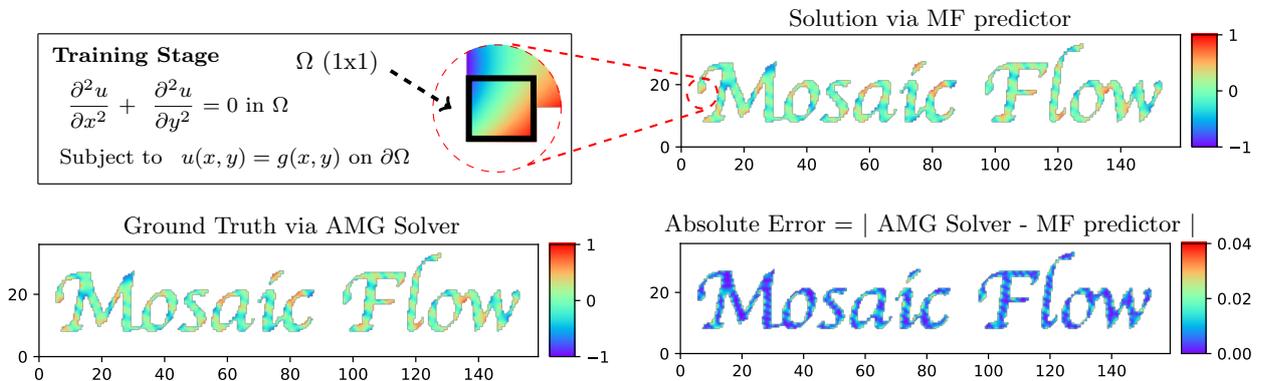
\begin{figure}[!b]
\vspace{-1em}
	\centering
	\input{./fig/Logo_SC_Figure.tex}
	\vspace{-1em}
	\caption[Logo SC]{Transferable learning of a PDE: We train our network in a small domain of size $1\times1$ for a wide range of boundary conditions. We then use the network to solve the Laplace equation in a domain that is $1200\times$ larger. Prediction time is almost $3$ hours on a single NVIDIA V100 GPU.}
	\label{fig:sclogo}
	\vspace{-1em}
\end{figure}

To bridge this gap, we develop a novel framework that builds a transferable DNN surrogate that solves PDEs in unseen domains with arbitrary BCs (see Figure \ref{fig:sclogo}). 
We present our idea in the context of flow prediction (i.e., surrogating the solution of the NS equations) but note that the framework is readily applicable to other PDE types such as the Laplace and Poisson equations.

For on-the-fly prediction of the flow in an unseen large domain with unseen BCs, we first decompose the domain into small subdomains called \textit{genomes}. 
Then, we predict the flow in each genome with a pre-trained DNN, called \emph{genomic flow network} (GFNet), such that the iterative assembly of these genome-wise predictions \revise{approximates} the flow in the large domain. 
We denote this assembled flow as a \textit{mosaic}. 
More specifically, this paper makes the following contributions:

\begin{itemize}
    \item We introduce GFNet which can solve a BVP with arbitrary BCs on a small square domain called \emph{genome}. 
    We design GFNet's architecture based on the characteristics of the PDE and show that such a design significantly improves accuracy. 
    \revise{For instance, our GFNet is twice more accurate than the conventional fully connected  (FC) architecture for solving the Laplace equation (Section \ref{sec:laplace}).
    For learning the NS equations, we further enforce the input BC in GFNet and reduce the generalization error by nearly 70\% compared to DeepONet \cite{Lu2019} and Fourier Neural Operator (FNO) \cite{li2020neural} (Section \ref{sec:comparison}).
    }
    \item We develop \emph{mosaic flow} (MF) predictor, \revise{a novel iterative algorithm based on the \emph{continuous Schwarz alternating algorithm} \cite{schwarz1972gesammelte} that iteratively assembles GFNet's inferences to predict the solution of the PDE system for unseen domains spanned by genomes with unseen BCs. 
    We show that MF predictor can scale up GFNet's inferences to domains $\sim\!\!1200\times$ and $12\times$ larger in the case of, respectively, Laplace and NS equations (Section \ref{sec:eval}).
    Moreover, MF predictor is independent of the neural network model used as GFNet.
    We compare our model with DeepONet \cite{Lu2019} and FNO \cite{li2020neural} and demonstrate that it reduces the error by nearly 6 times when employed by MF predictor to predict unseen flow features.
    }
    \item With GFNet and MF predictor, we eliminate the need to re-train a DNN for unseen domains or BCs. 
    This transferability delivers 1-3 orders of magnitude speedup compared to the start-of-the-art PINN for solving the Laplace and NS equations while achieving comparable or better accuracy (Section \ref{sec:eval}). 
    \item Our GFNets are not tailored to specific applications and hence can be used by others. This reusability saves energy and benefits researchers with limited access to GPUs.
\end{itemize}


%% file: fig/Logo_SC_Figure.tex
\begin{tikzpicture}
\def \y{2.7cm} 
\def \yunits{1.25} 

\def \ydis{-0.7} 
\def \ydisp{0.3} 
\def \xdisp{0.4}
\def \xtrim{0.1}
\def \LCD{-0.7}

\def \xsoldis{-0.1}

\node[anchor=south west,inner sep=0] at (-6.7 -\xsoldis ,\y) {\includegraphics[width=0.5\textwidth]{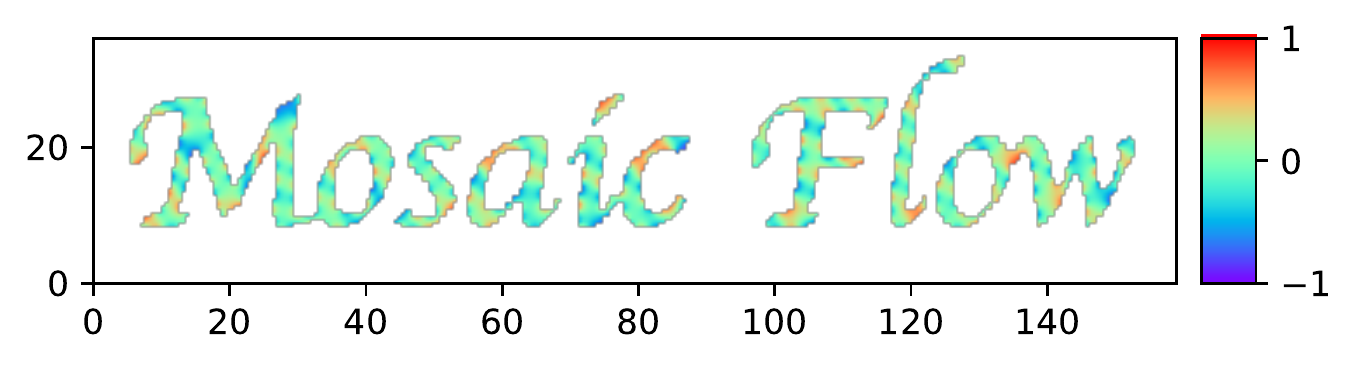}};

\node[anchor=south west,inner sep=0] at (1.75-\xsoldis,5.48) {\includegraphics[width=0.5\textwidth]{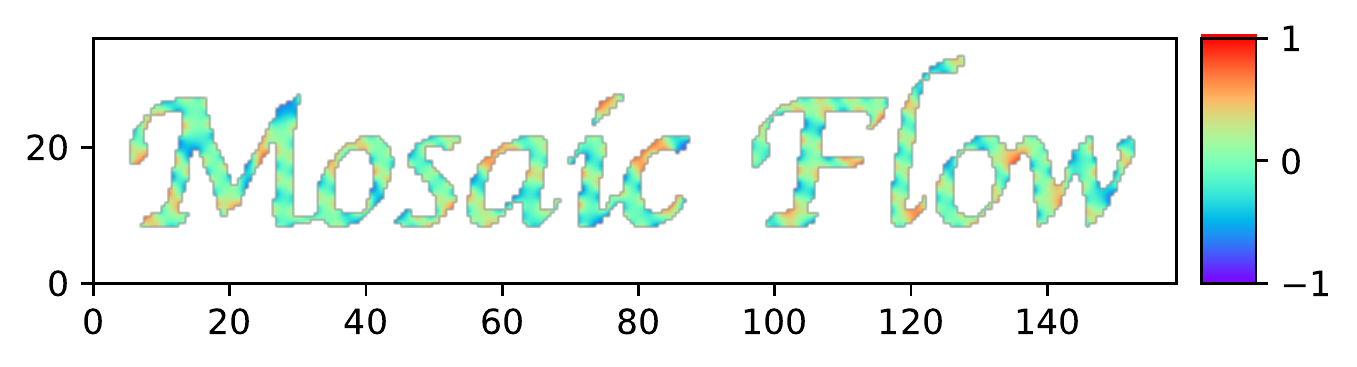}};

\node[anchor=south west,inner sep=0] at (1.75-\xsoldis,\y + 0.05cm) {\includegraphics[width=0.5\textwidth]{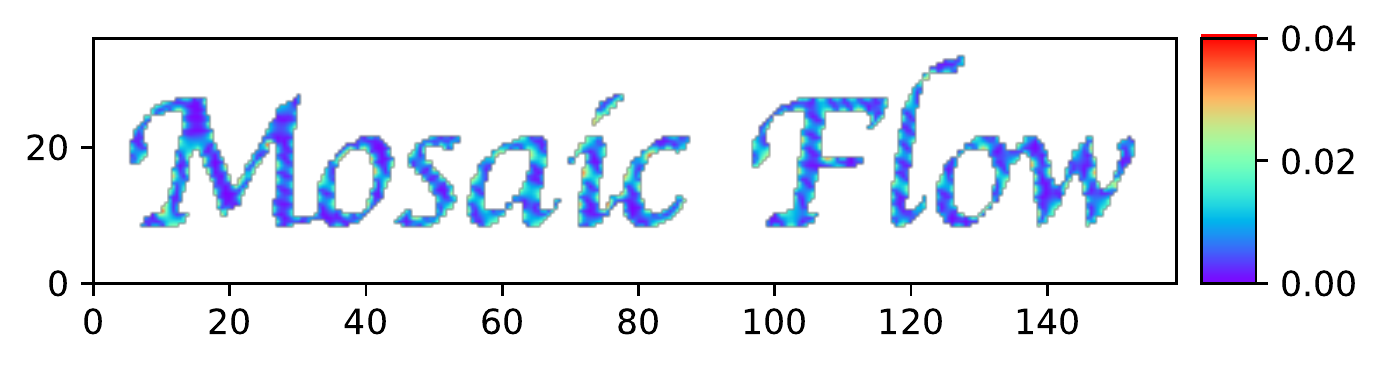}};

\begin{scope}
\clip[](0.1- \xtrim,5.8- \ydis) circle (0.85cm);
\node[anchor=south west,inner sep=0] at (-3.02 - \xtrim ,4.65) {\includegraphics[width=0.49\textwidth]{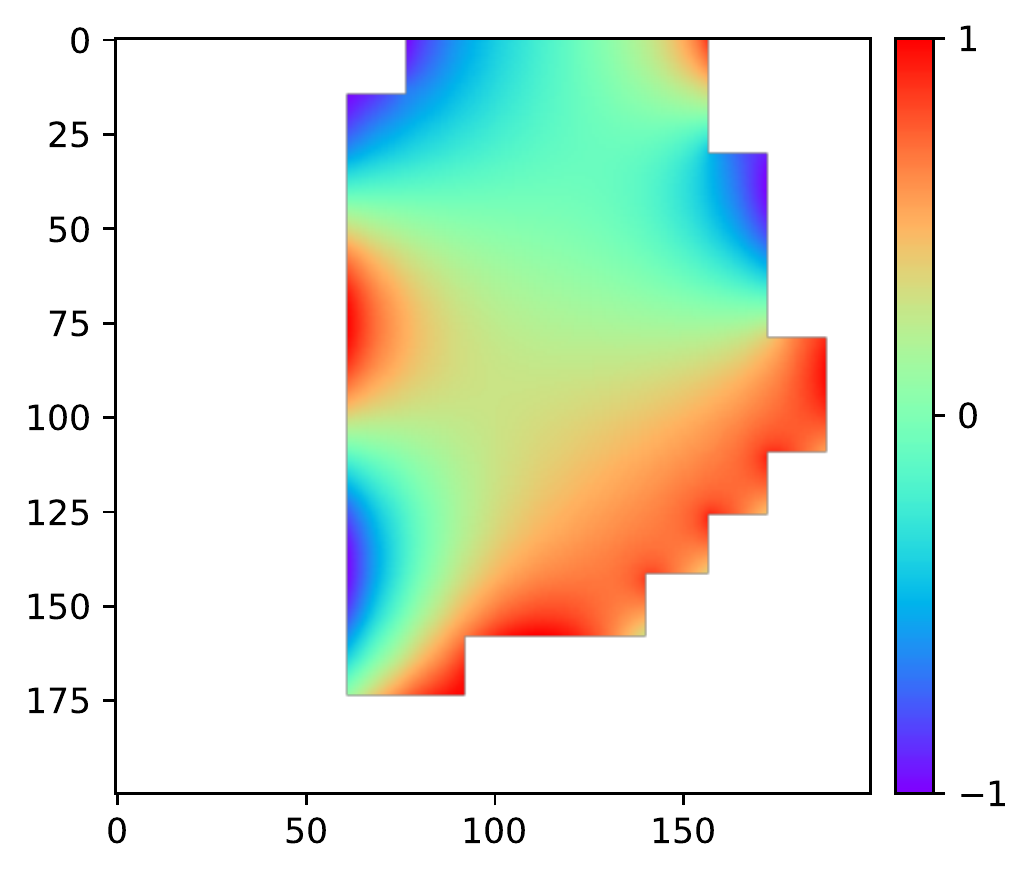}};

\draw[red,thick,dashed, ] (0.1- \xtrim,5.8- \ydis) circle (0.85cm);
\draw [line width=0.8mm] (-0.27- \xtrim,5.7 - \ydis -\ydisp) rectangle (0.0 + 0.58 - \xtrim,6.55 - \ydis -\ydisp - 0.05);
\end{scope}

\node at (-2.7,3.7 + \yunits) {\small Ground Truth via AMG Solver};

\node at (5.7,7.67) {\small Solution via \mfpShort };

\node at (5.7,3.7 + \yunits) {\small Absolute Error = $\mid$ AMG Solver - \mfpShort $\mid$ };

\def \xdispl{0.4}
\def \xmar{0.12}
\def \ymar{0.06}
\draw [rounded corners=0.3, line width=0.15mm] (-6.44 + \xdispl,5.5) rectangle (0.98,7.49);

\node at (-2 - \xtrim,7.1) {\small $\Omega$ (1x1)};
\draw[red,thick,dashed, ] (2.7,6.7 - \ydis + \LCD) circle (0.2cm);

\draw[thick, dashed, red] (0.5- \xtrim,5.05 - \ydis ) -- (2.8,6.5 - \ydis+ \LCD);
\draw[thick, dashed, red] (+ 0.3 - \xtrim,6.65 - \ydis ) -- (2.8,6.88 - \ydis + \LCD);

\draw[-> , line width=0.5mm, dashed, black] (-1.3 - \xtrim,7) -- (-0.5 - \xtrim,6.5);
\node[scale=1, anchor=west] at (-6.5 + \xdispl + \xmar,7.25 - \ymar) {\footnotesize \textbf{Training Stage}};
\node[scale=1,  anchor=west] at (-6.4 + \xdispl + \xmar,6.6- \ymar) {\footnotesize { $\dfrac{\partial^2 u}{\partial x^2}$} $+$ { $\dfrac{\partial^2 u}{\partial y^2}$}  $= 0 $ in $ \Omega$};

\node[scale=1, anchor=west] at (-6.4 + \xdispl + \xmar,5.9- \ymar) {\footnotesize Subject to};
\node[scale=1, anchor=west] at (-4.8 + \xdispl + \xmar,5.9- \ymar) {\footnotesize $u(x,y) = g(x,y) $ on $ \partial \Omega$};

\end{tikzpicture}

%% file: related.tex
\vspace{-1mm}
\section{Related Work}\label{sec:related}
Existing approaches for building DNNs that solve PDEs employ a wide range of training mechanisms and network architectures. These choices largely depend on the application and the characteristics of the PDEs. The architectures are typically built with fully connected (FC), convolutional, or recurrent \cite{RN847, RN772, RN770} layers. In many works a combination of these layers, with or without residual connections \cite{RN1060}, are employed as well. 
For instance, networks similar to Resnet \cite{RN1060} and UNet \cite{RN1451}, whose building blocks are convolutional neural networks (CNNs), are generally the backbone of super-resolution frameworks that aim to reconstruct a high-resolution solution from a coarse solution \cite{jiang2020meshfreeflownet, Linebarger1899}. 
Such networks have also been trained in the Fourier space to learn a PDE operator using the modal space \cite{li2020neural, RN1250}. 
CNNs, which may also include FC layers, are extensively used as surrogates for predicting steady laminar flows around 2D and 3D objects \cite{RN1452}, estimating 2D flows around specific airfoils \cite{Bhatnagar2019, Obiols-Sales2020}, flow visualization \cite{Wandel2020}, and many other applications \cite{RN1067, RN1065, RN1063, dong2019adaptive}.

Solving PDEs via CNN-based networks is data-efficient (due to parameter sharing and learning spatially invariant kernels \cite{RN1163, RN982, RN978, RN981}), but it requires developing a mechanism for interpolating the solution (and its gradients) on non-grid points \cite{RN1422}. 
One strategy to avoid such interpolation errors is to only employ FC layers. 
This strategy has been used for inverse estimation of PDE coefficients \cite{Raissi2019}, inverse estimation of flow fields given observations on a passive scalar \cite{RN1069}, predicting the Reynolds stress in Reynolds-averaged NS simulations \cite{RN1162, RN1173}, and solving Poisson equation and eigenvalue problems \cite{RN1190}. 
Networks based on FC layers have also been reformulated for variational learning (to improve predictions on the boundaries) \cite{RN1191, RN1188}, surrogating fractional PDEs (where AD cannot calculate residual errors) \cite{RN1420}, emulating long time-dependent PDEs (where long temporal features render the training very difficult) \cite{RN1169}, and learning operators that map functions to functions \cite{RN1381, RN1382, Lu2019}. 

Regardless of the architecture, existing works lack \emph{transferability}, i.e., they build models that are largely useless if the domain and BCs associated with the PDE system are modified. This important issue has been rarely addressed.
DNNs that learn a PDE operator \cite{RN1421, Lu2019}, can in theory handle varying BCs but they are not domain-agnostic and their applications thus far have been limited to simple PDEs over small, fixed domains.
In \cite{Ozbay2019}, CNNs are used to solve the $2D$ Poisson equation on rectangular domains with different aspect ratios. 
In \cite{Maulik2019}, the DNN is trained to solve the NS equations over backward-facing steps with varying heights. 
In \cite{Shahane2020}, the DL model infers the lift and drag of various elliptic objects with different aspect ratios. 
In all these works, the geometry type (e.g., rectangle or ellipse) is fixed and only varies within a very limited range with one parameter.
A few studies \cite{Tompson2017,dong2019adaptive,Bhatnagar2019, Wandel2020, Obiols-Sales2020} have proposed more general techniques to embed geometry information into the training stage of a DNN such that it can predict the quantities of interests in unseen domains. 
However, the overall domain size is still fixed, grid data are required, and the variability of the objects is limited to specific applications.

Since the training costs of a PDE-solving DNN are generally high, parallel and distributed computations have been explored in their training \cite{Jagtap2020}. However, the overall training costs are still high given the iterative and combinatorial process of tuning the architecture, optimization parameters, and loss components. To justify these high training costs, a DNN is expected to be transferable across multiple applications.

%% file: DL_model.tex
\section{Transferable Learning of PDE\lowercase{s}}\label{sec:dl_model}
Our goal is to \revise{approximate} the solution of the following homogeneous BVP in an \textit{arbitrary} domain $\Omega$ for a wide range of $g(\pmb{x})$:
\begin{equation}
    \begin{aligned}
        H(u(\pmb{x})) &= 0, \quad \pmb{x} \in \Omega, \\
        u(\pmb{x}) &= g(\pmb{x}), \quad \pmb{x} \in \partial\Omega,
        \label{eqn:bvp}
    \end{aligned}
\end{equation}
where $H(\cdot)$ denotes a partial differential operator composed of spatial derivatives. Henceforth, we drop the dependence of $u$ to $\pmb{x}$ for notational convenience. We highlight that while Equation \ref{eqn:bvp} is scalar, our approach is directly applicable to PDE systems such as the NS equations, see Section \ref{sec:ns}.

Building a \textit{single} DNN that solves the BVP of Equation \ref{eqn:bvp} in \textit{any} domain and under \textit{any} BC is an extremely difficult (if not infeasible) learning task because considering arbitrarily large variations in domains and BCs in training drastically amplifies the challenges described in Section \ref{intro}. In our framework, we solve this difficult task by decomposing it into two simpler (but highly coupled) tasks that correspond to ($1$) training a GFNet that solves the BVP in a genome (i.e., a small square) given any $g(\pmb{x})$, and ($2$) building \mfpShort that \revise{approximates} the solution in \revise{any unseen large domain that can be spanned with genomes} by stitching the genome-wise predictions made by the pre-trained GFNet. The advantage of our approach is that predicting the solution of the BVP \revise{in an unseen domain} only relies on task $2$, i.e., there is no need to train or re-train a DNN anymore.

Our task decomposition strategy is motivated by curriculum learning \cite{RN1157} which was originally developed for animal training \cite{RN1158} where difficult tasks are divided into simpler tasks which are easier to learn and collectively make up the original task. In Section \ref{sec:bvp_prelim} we demonstrate that this approach is also applicable to solving BVPs. Then, we elaborate on how to build a GFNet and an \mfpShort in Sections \ref{sec:nn_bvp} and \ref{sec:overlap_genome}, respectively. 

\subsection{BVP Decomposition: Why Mosaic Works?}\label{sec:bvp_prelim}

\begin{figure*}[!b]
    \centering
    \subfloat[Interchangeability between a BVP and its genomes' BVPs\label{fig:genome_bvp}]
    {\input{./fig/sub_domain_bvp.tex}}\\
    \subfloat[Direct solution via AMG\label{fig:iter_sol}]
    {\raisebox{4mm}{\includegraphics[width=5cm]{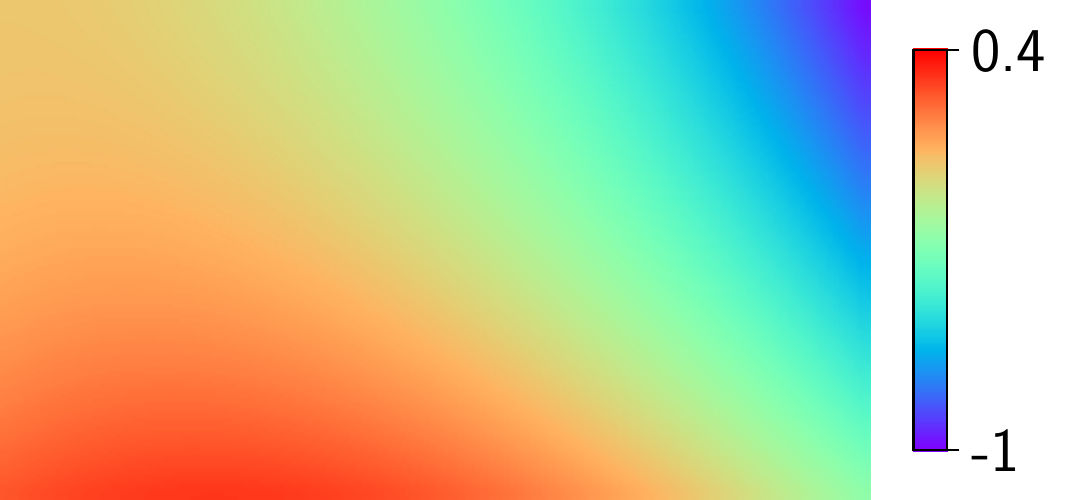}}}\hfil
    \subfloat[Domain decomposition\label{fig:iter_2genome}] 
    {\input{./fig/iter_2genome}}\\
    \subfloat[Convergence history in log-scale\label{fig:iter_hist}]
    {\input{./fig/iter_history}}
    \caption{Decomposition of a BVP: The Laplace equation is solved in $\Omega=[0,2]\times [0,1]$ with three methods: a direct and two iterative approaches. The iterative approach that leverages an auxiliary genome converges to the direct solution $12\times$ faster.
    }
    \label{fig:solve_genome_bvp}
\end{figure*}

The BVP in Equation \ref{eqn:bvp} is well-posed if the solution, $u$, is unique. For such a BVP, $u$ on any arbitrary sub-domain inside $\Omega$ also forms a well-posed BVP. 
Figure \ref{fig:genome_bvp} illustrates this relation for the Laplace equation where the domain $\Omega$ is split into two genomes, namely, $\Omega_1$ and $\Omega_2$. 
The boundaries of these genomes are denoted by $\partial\Omega_1$ and $\partial\Omega_2$ which share the border $\partial\Omega_1\cap\partial\Omega_2$ in the middle of $\Omega$ (marked blue in the figure). 
Once the Laplace equation is solved in $\Omega$, $u$ will be available on $\partial\Omega_1\cap\partial\Omega_2$. 
We can now solve two BVPs: one in $\Omega_1$ and the other in $\Omega_2$ to get the solutions $u_1$ and $u_2$, respectively. 
The above well-posedness relation indicates $u_1=u|_{\Omega_1}$ and $u_2=u|_{\Omega_2}$.

Conversely, if we can find $u_1$ and $u_2$ while ensuring that the PDE is satisfied on the shared border, i.e., $\nabla^2 u=0$ on  $\partial\Omega_1\cap\partial\Omega_2$, then we will have $u$ since $u=u_1 \cup u_2$. 
Indeed, to find $u_1$ and $u_2$ we need to solve two BVPs (one in $\Omega_1$ and the other in $\Omega_2$) both of which require $u$ on $\partial\Omega_1\cap\partial\Omega_2$. 
Since $u$ is unknown, we propose an iterative strategy to find it on the shared border: we first randomly assign values to $u$ on $\partial\Omega_1\cap\partial\Omega_2$ and solve two BVPs in $\Omega_1$ and $\Omega_2$ to find $u_1$ and $u_2$, respectively. 
Then, we use $u_1$ and $u_2$ to update our initial guess and repeat this process until a convergence criterion is met (e.g., until values on $\partial\Omega_1\cap\partial\Omega_2$ negligibly change across consecutive iterations).
\revise{
As illustrated below, we can dramatically accelerate the convergence speed of this simple iterative approach via the \emph{continuous Schwarz alternating algorithm} (aka Schwarz method) \cite{schwarz1972gesammelte} which greatly increases the pace of information propagation from $\partial\Omega$ to $\Omega$ using an \emph{auxiliary} genome that overlaps with $\Omega_1$ and $\Omega_2$ (see Figure \ref{fig:iter_2genome}).

We illustrate the Schwarz method by solving the BVP in Equation \ref{eqn:ex_laplace} in a domain of size $2\times1$ (i.e., $\Omega=[0,2]\times [0,1]$) subject to the boundary condition $g(\pmb{x}) =x^2 - y^2 + xy$. 
As illustrated in Figure \ref{fig:iter_sol}, we first use algebraic multi-grid (AMG) \cite{OlSc2018} to directly find the ground truth, i.e., $u$ in $\Omega$.
Then, we use AMG within both iterative approaches and show that while both the simple iterative approach and Schwarz method converge to the ground truth, the latter is significantly faster.
}


Since $\Omega$ is of size $2\times1$, we decompose the domain into two \textit{basic} genomes of size $1\times1$ that share a single border, see Figure \ref{fig:iter_2genome}. In the first iterative strategy, we initialize the shared border with zero values for $u$, solve for $u_1$ and $u_2$ using AMG, and exchange data across the border. This iterative process converges in 300 iterations when the mean absolute error (MAE) compared to the direct solution drops below 1e-14.

\revise{
In the Schwarz method, after finding $u_1$ and $u_2$, we update $u$ on the shared border by solving the Laplace equation in an auxiliary genome whose BCs on its two vertical borders are inside $\Omega_1$ and $\Omega_2$, see Figure \ref{fig:iter_2genome}.
}
This approach introduces a stronger and more effective coupling across the basic genomes which, in turn, accelerates the rate of information propagation from $\partial\Omega$ to the shared border. The higher rate of information propagation increases the convergence rate by almost $12$ times compared to the first iterative strategy, see Figure \ref{fig:iter_hist}. 

\revise{
For the general elliptic PDE, 
\begin{equation}
\begin{aligned}
    H(u(\pmb{x})) &=\nabla\cdot(\alpha\nabla u) - f= 0, \quad \pmb{x} \in \Omega, \\
    u(\pmb{x}) &= g(\pmb{x}), \quad \pmb{x} \in \partial\Omega,
\end{aligned}
\label{eqn:elliptic}
\end{equation}
where $\alpha(\bx)$ is a variable coefficient in space and $f(\bx)$ denotes the source term, the iterative solution by Schwarz method converges to the PDE's exact solution \cite{xu1992iterative, toselli2004domain}.
To illustrate this, we formulate the approximate solution after solving BVP in the $i^{th}$ genome ($\Omega_i$) as follow \cite{mathew2008domain}
\begin{equation}
    u^{k+\frac{i}{N_g}}(\bx) = u^{k+\frac{i-1}{N_g}}(\bx)  
    + P_i(u^*(\bx) - u^{k+\frac{i-1}{N_g}}(\bx)),
    \quad\bx\in\Omega \\
\label{eqn:iter_genome}
\end{equation}
where $N_g$ denotes the total number of basic and auxiliary genomes, $u^*$ is the exact solution for the given BC, and $u^{k+\frac{i}{N_g}}$ is the solution after solving BVP in the $i^{th}$ genome at the $k^{th}$ iteration. 
We define one iteration as updating all genomes exactly once.
In Equation \ref{eqn:iter_genome}, $P_i$ is a projection operator mapping the error $u^* - u^{k+(i-1)/N_g}$ to a function that has zero values outside $\Omega_i$.
We refer readers to \cite{mathew2008domain} for more details about $P_i$'s proof of existence and properties.
Applying Equation \ref{eqn:iter_genome} to all the genomes in one iteration yields the \emph{error propagation map} \cite{mathew2008domain},
\begin{equation}
|u^* - u^{k+1}| = |\prod_i^{N_g}(I - P_i)|\;|(u^*-u^{k})|,
\label{eqn:schwarz_err}
\end{equation}
where the coefficient $E = |\prod_i^{N_g}(I - P_i)|$ prescribes how fast the error decays. 
It can be proved that $E<1$ for general elliptic PDEs and this is independent of the numerical method (e.g., FE, FD, etc.) used for solving the genome-wise BVPs \cite{xu1992iterative,toselli2004domain}.

Our framework is akin to the Schwarz method, except for the following two aspects. 
First, instead of using a numerical method we always use a single pre-trained GFNet to predict the solution in either the basic genomes or the auxiliary genomes. 
Following Equation \ref{eqn:schwarz_err}, if GFNet's generalization error approaches 0 (see Section \ref{sec:comparison} for the definition of different error types), the prediction of our framework converges to the elliptic BVP's exact solution.
Second, there is no mechanism in the original Schwarz method to define the proper arrangement of the overlapping sub-domains.
In MF predictor, we distinguish between basic and auxiliary genomes and propose general guidance for placing genomes and ordering their updates (see Section \ref{sec:overlap_genome}).
}

\subsection{Genomic Flow Network (GFNet)}\label{sec:nn_bvp}
The iterative nature of \mfpShort requires a GFNet to be able to estimate the solution anywhere inside a genome for a wide range of BCs. Hence, the inputs of GFNet are:
\vspace{-1mm}
\begin{equation*}
    \text{GFNet inputs:}\quad
    \underbrace{g(\pmb{x}^{bc}_1),\;g(\pmb{x}^{bc}_2),\;\cdot\cdot\cdot,\; g(\pmb{x}^{bc}_{N_{bc}})}_{\displaystyle \pmb{g} \text{ of size } N_{var}\times N_{bc}}, \;\;
    \raisebox{-0.7mm}{$\underbrace{\;\pmb{x}_{}\;}_{\displaystyle N_{dim}}$}
    \label{eqn:nn_input}
\end{equation*}

\noindent where $N_{bc}$, $N_{var}$ and $N_{dim}$ denote the number of points on the genome boundary, variables, and spatial dimensions, respectively. We consider $2D$ problems in this work so $N_{dim} = 2$. In scalar PDEs such as the Laplace equation, $N_{var} = 1$ while $N_{var} = 3$ for PDEs such as the incompressible NS equations studied in Section \ref{sec:ns}.

In the above, $\pmb{g}$ represents a discretized version of $\bg$ at the genome's boundaries and $\pmb{x}$ indicates the coordinates of the point where GFNet predicts the solution, i.e., $u(\pmb{x})$ which has $N_{var}$ components. The loss function of GFNet is adopted from Equation \ref{eqn:ex_pinn_loss} where the size of the training data (e.g., $N_1$ and $N_2$) and coefficients (e.g., $\alpha$ and $\beta$) are chosen based on the BVP. For instance, our studies indicate that while residual loss improves the accuracy of GFNet when learning the Laplace equation, it reduces the accuracy in the case of the NS equations, see Section \ref{sec:eval} for the details and our reasoning. Additionally, when residual errors improve the accuracy of GFNet, we introduce a mechanism to adaptively (instead of randomly) choose the location of collocation points in the genome. 

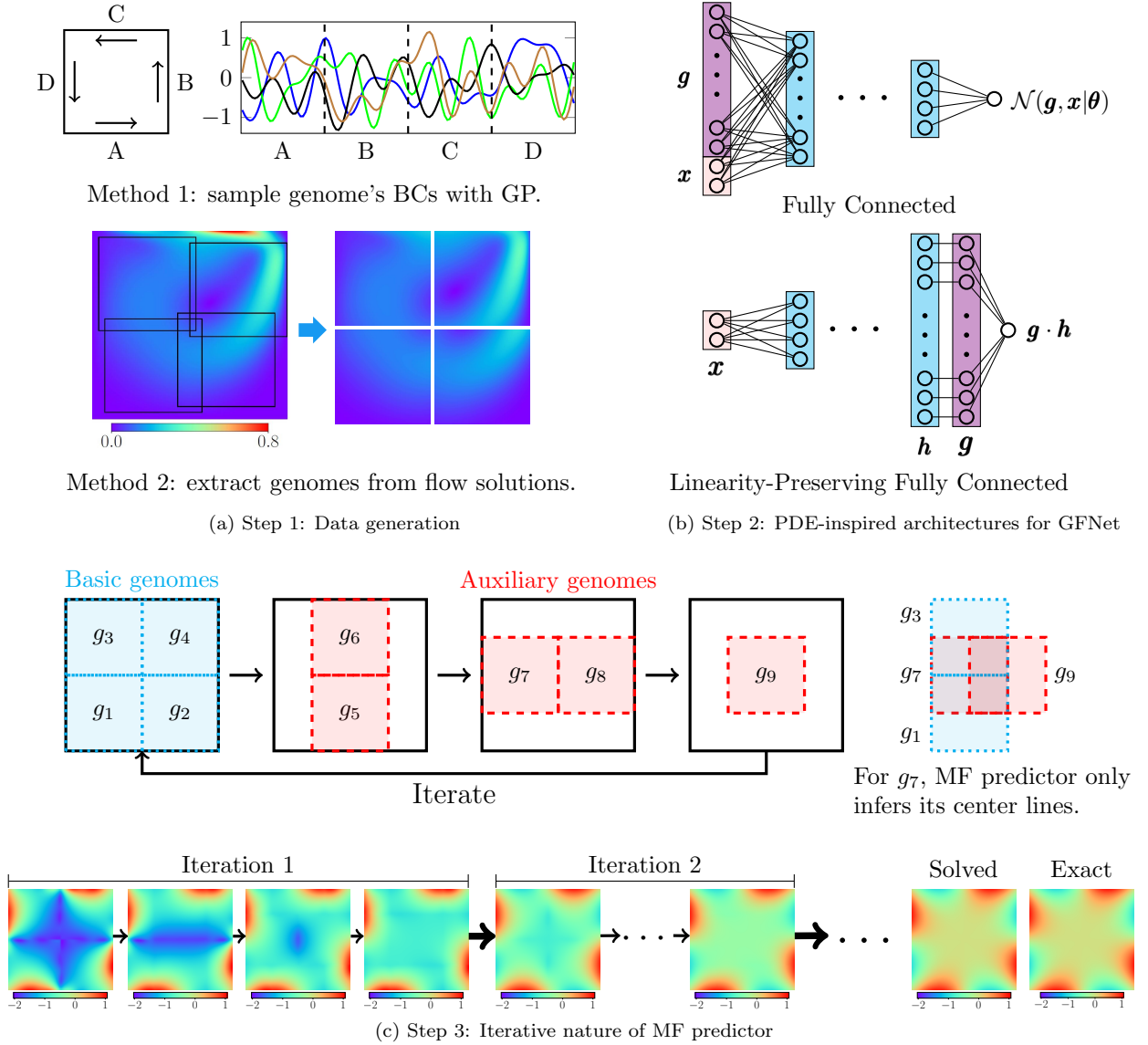
\begin{figure*}[!b]
    \centering
    \subfloat[Step 1: Data generation\label{fig:step1}]
    {\input{./fig/genome_bc_gp.tex}}\hfil
    \subfloat[Step 2: PDE-inspired architectures for GFNet\label{fig:step2}]
    {\input{./fig/gfnet_arch.tex}}\\
    \subfloat[Step 3: Iterative nature of \mfpShort\label{fig:step3}]
    {\input{./fig/mf_predict_genome.tex}}
    \caption{Primary steps of our framework for transferable PDE learning: Steps $1$ and $2$ are done only once and their cost is fully amortized in step $3$ which can be applied to unseen domains.
    \subc{a} Training data are generated by either solving a BVP whose BCs are sampled from a GP, or by sweeping the solution of the BVP in a large domain with a genome (the plot in method 2 corresponds to the velocity magnitude in a lid-driven cavity simulation).
    \subc{b} A GFNet takes the discretized boundary function $\pmb{g}$ and $\pmb{x}$ as inputs and outputs the solution at $\pmb{x}$. We use LPFC and FC architectures for linear and non-linear PDEs, respectively.
    \subc{c} Mosaic flow (MF) predictor estimates the solution in an unseen domain by systematically coupling the predictions of a GFNet on basic and auxiliary genomes. Note that \baseg can overlap with each other and \axg can be placed anywhere inside the domain (we have avoided these scenarios to improve clarity). 
    }
    \label{fig:workflow}
\end{figure*}

We design the architecture of GFNet based on the PDE. In linear PDEs such as the Laplace equation, $u(\pmb{x})$ is essentially a linear combination of the discretized boundary function values, $\pmb{g}$ (note that the solution can still be non-linear if $\bg$ is a linear function).
To preserve the linearity, we propose a novel \emph{linearity-preserving fully connected} (LPFC) architecture shown in Figure \ref{fig:step2}. 
In LPFC, we pass the coordinates of the input point $\pmb{x}$ through multiple fully connected hidden layers and set the last hidden layer's outputs, $\pmb{h}$, to have the same size as $\pmb{g}$. The output of GFNet is a linear combination of $\pmb{g}$ with $\pmb{h}$, i.e., $\pmb{g}\cdot\pmb{h}$. With this configuration, $\pmb{h}$ leverages the spatial correlations in $\Omega$ and $\pmb{g}\cdot\pmb{h}$ enforces it to conform to the linearity of the PDE.
For non-linear PDEs such as the NS equations, we adopt a \emph{fully connected} (FC) architecture where $\pmb{g}$ and $\pmb{x}$ are concatenated into one input vector and passed through all the FC hidden layers, see Figure \ref{fig:step2}.
A detailed evaluation of LPFC and FC architectures is presented in Section \ref{sec:eval}.

Since a GFNet takes $\pmb{g}$ as an input, a wide range of well-posed (and ideally negligibly correlated) BCs must be used in training.
For linear PDEs such as the Laplace equation, the BVP has a unique solution as long as the boundary function $\bg$ is smooth, i.e., its derivatives in $H(\cdot)$ exist. Hence, we use GPs to generate a smooth function in $1D$ and then wrap that function around the perimeter of the genome, see Figure \ref{fig:step1}. GPs are stochastic processes that place a prior distribution on functions. By judiciously selecting the hyperparameters of their kernel, we can generate a wide range of smooth functions that are minimally correlated.
For highly nonlinear PDEs such as the NS equations, the BCs must satisfy mass, momentum, and energy conservation laws which makes it highly challenging to create well-posed BCs via a GP.
Hence, we first solve the PDEs with realistic BCs on a large domain and then extract solutions for small genomes embedded in the domain. This strategy is demonstrated in Figure \ref{fig:step1} where the flow in a lid-driven cavity is swept by a genome while recording the flow on the boundary as well as inside the genome (the cavity's domain spans $[0,1]\times[0,1]$ while the genome has a size of $0.5\times0.5$). As argued in Section \ref{sec:bvp_prelim}, this procedure generates valid data since the flow on each genome's boundary constitutes the BCs for the BVP on that genome.

\revise{Note that the genomes in training and prediction have the same shape because GFNet does not learn the effect of the genome's shape. We can address this limitation by including the boundary's geometrical information as input but this is beyond the scope of this work.}

\subsection{Mosaic Flow Predictor}\label{sec:overlap_genome}
As detailed in Section \ref{sec:bvp_prelim} we can (1) convert a BVP in a large domain to multiple BVPs in genomes that cover that domain and then solve these BVPs iteratively, and (2) employ auxiliary genomes to significantly accelerate the convergence. Based on the above two attributes, \mfpShort decomposes an unseen domain into two sets of genomes, see Figure \ref{fig:step3}.
The first set consists of \baseg that cover the entire domain and can overlap with each other. The BCs on the border of \baseg are either known (if the borders lie on the large domain's boundary) or must be inferred via MF predictor.
The second set consists of \axg that overlap with the \baseg to accelerate the propagation of boundary information into the domain, i.e., to speed up the iterative process of estimating the unknown BCs of the basic genomes.
We distribute these two sets of genomes based on the following general guidelines.
First, we arrange the basic genomes to cover the entire domain with as little overlap as possible ($g1$ through $g4$ in Figure \ref{fig:step3}).
Then, we place auxiliary genomes on the vertical and horizontal shared borders of the basic genomes ($g5$ through \revise{$g8$} in Figure \ref{fig:step3}).
Finally, we place an additional auxiliary genome on every corner/region where four basic genomes connect/overlap ($g9$ in Figure \ref{fig:step3}). We have developed this spatial genome distribution to balance the pace of information transfer (which directly affects convergence speed) and implementation efforts. As we demonstrate in Section \ref{sec:ns_eval}, for highly nonlinear PDEs such as the NS equations, more auxiliary genomes are required to facilitate information propagation. In such cases, following the above procedure, we place the additional auxiliary genomes on the shared borders of other auxiliary genomes. We will explore more advanced techniques in our future works.

\mfpShort finds the solution in the genomes set by set using a pre-trained GFNet and iterates until convergence. 
\revise{In one iteration, \mfpShort updates all the genomes once in the order that the genomes are grouped and placed.
That is, it first predicts for all the basic genomes, then infers auxiliary genomes covering the basic genomes' horizontal and vertical borders, and at last updates the genomes overlapping the basic genomes' corners.
The rationale behind this order is that, updating genomes with more boundary information (adjacent or close to boundary) before the ones with less information can maximize the pace at which the information is propagated to the domain's interior.
For instance, the basic genomes are inferred first as they cover the entire domain boundary while the auxiliary genomes overlapping the basic genomes' corners are inferred at last because they are typically immersed inside the domain without direct contact to the boundary.
}
The basic genomes' borders are initialized either with the domain's BC if they lie on the domain's boundary or with 0 if they lie between \baseg.
The predictions in \baseg provide the BCs for \axg and then the predictions in \axg update the unknown BCs of the basic genomes. 
\revise{ The iterations stop when the change in the inferred BCs of \baseg is smaller than the user-defined tolerance $\epsilon$.
}

Figure \ref{fig:step3} illustrates the spatial distribution of basic and auxiliary genomes and the iterative nature of MF predictor for solving the Laplace equation in $\Omega = [0,2]\times[0,2]$ with genomes of size $1\times1$.
In this figure, there are 4 basic genomes ($g1$-$g4$) that, without any overlap, cover $\Omega$. 
The BCs are only known on two sides of the \baseg and hence the \mfpShort aims to predict the unknown BCs on the other two sides. 
There are also 5 \axg ($g5$-$g9$) that accelerate the pace of inferring the unknown BCs of the basic genomes. 
\revise{ In one iteration, the genomes are updated in the following order: $g1$-$g4$ $\rightarrow$ $g5$, $g6$ $\rightarrow$ $g7$, $g8$ $\rightarrow$ $g9$.
(within this order, non-overlapping genomes can be updated concurrently, e.g., $g1$ through $g4$ or $g5$ and $g6$.
We plan to exploit this parallelism in our future work.)
}
In this example, \mfpShort converges to the exact solution in $18$ iterations.

In contrast to conventional numerical methods and PINN, \mfpShort does \textit{not} solve for all the grid/collocation points in the entire domain. Instead, it only infers the solution on the borders of basic and auxiliary genomes. For instance, for genome $g_7$ in Figure \ref{fig:step3}, MF predictor uses the pre-trained GFNet to predict the solution on its vertical and horizontal center lines. These predictions are then used for inferring the solutions in genomes $g_9$ (in the current iteration) as well as $g_1$ and $g_3$ (in the next iteration).

%% file: fig/sub_domain_bvp.tex
\begin{tikzpicture}
\def\a{2};
\def\b{2};
\def\c{6};

\draw[thick] (2*\a+\c, 0) rectangle +(2*\a, \b);
\draw[very thick, blue] (3*\a+\c, 0) -- +(0, \b);
\node at (2.5*\a+\c, 0.5*\b) {$u_1 = u|_{\Omega_1}$};
\node at (3.5*\a+\c, 0.5*\b) {$u_2 = u|_{\Omega_2}$};
\node[below] at (\a, 0) {Original BVP on $\Omega = \Omega_1\cup\Omega_2$};

\draw[thick] (0, 0) rectangle +(2*\a, \b);
\node at (\a, 0.5*\b) {
$\begin{aligned} 
  \nabla^2 u &= 0,\quad\;\;\pmb{x}\in\Omega \\
  u &= g(\pmb{x}), \:\pmb{x}\in\partial\Omega 
\end{aligned}$
};
\node[below] at (3*\a+\c, 0) {Two well-posed BVPs in $\Omega_1$ and $\Omega_2$};

\draw [line width=2, ->, >=stealth, pink] (2*\a+0.1*\c, 0.7*\b) -- node[above, black] {
  \parbox[t]{3.4cm}
  {Extract genomes' BC from $g(\pmb{x})$ and $u(\pmb{x})$.}
  } ++(0.8*\c, 0);

\draw [line width=2, ->, >=stealth, pink] (2*\a+0.9*\c, 0.3*\b) -- node[below, black] {\parbox[t]{3.6cm}{
  $u = \begin{cases} 
    u_1, &\pmb{x} \in \Omega_1 \\ 
    u_2, &\pmb{x} \in \Omega_2 
  \end{cases}$, \\
  $\nabla^2 u = 0$ on $\partial\Omega_1\cap\partial\Omega_2$
  }} ++(-0.8*\c, 0);

\end{tikzpicture}

%% file: fig/iter_2genome.tex
\begin{tikzpicture}
\def\h{2.2};
\def\hh{0.2};
\draw[line width = 0.5mm ] (0,0) rectangle (\h,\h);
\draw[line width = 0.5mm] (\h, 0) rectangle (2*\h, \h);

\node[cyan] at (0.5*\h-0.07*\h, -0.12*\h) {\basegone 0};
\node[cyan] at (1.5*\h + 0.1*\h, -0.12*\h) {\basegone 1};
\node[red] at (\h, 1.12*\h) {\axgone};

\draw[line width = 0.5mm ,cyan,  dotted, fill=cyan,  opacity=.1,] (0,0) rectangle (\h,\h);
\draw[line width = 0.5mm , dotted, cyan,] (0,0) rectangle (\h,\h);
\draw[line width = 0.5mm , dotted, cyan,fill=cyan,  opacity=.1,] (\h, 0) rectangle (2*\h, \h);
\draw[line width = 0.5mm , dotted, cyan,] (\h, 0) rectangle (2*\h, \h);

\draw[thick, dashed, red, fill=red,  opacity=.15] (0.5*\h, 0) rectangle (1.5*\h, \h);
\draw[thick, dashed, red,] (0.5*\h, 0) rectangle (1.5*\h, \h);

\end{tikzpicture}

%% file: fig/iter_history.tex
\begin{tikzpicture}
\begin{axis}[
  ylabel = MAE ($10^{n}$),
  every axis y label/.style = {at={(ticklabel cs:0.5)}, rotate=90, anchor=near ticklabel},
  every axis x label/.style = {at={(ticklabel cs:0.5)}, anchor=near ticklabel},
  legend pos = north east,
  legend cell align = {left},
  legend style = {draw=none},
  width = 8cm,
  height = 4cm,
  xmin = 1,
  xmax = 400,
  ymin = 1e-14,
  ymax = 1.0,
  ymode = log,
  xtick pos = left,
  ytick pos = left,
  xtick = {1, 100, 200, 300, 400},
  xticklabels = {1,100, 200, 300, Iterations},
  ytick = {1.0, 1e-5, 1e-10, 1e-14},
  yticklabels = {0, -5, -10, -14},
]
\addplot[cyan, line width=1.2] table[x=i, y=mean]{./fig/non_overlap.log};
\addplot[red,  line width=1.2] table[x=i, y=mean] {./fig/overlap.log};
\legend{w/o aux genome , w/ \ \  aux genome};
\end{axis}
\end{tikzpicture}

%% file: fig/genome_bc_gp.tex
\begin{tikzpicture}
\def\h{1.5};
\draw[thick] (0,0) rectangle (\h,\h);
\node[below] at (0.5*\h, 0)  {A};
\node[right] at (\h, 0.5*\h) {B};
\node[above] at (0.5*\h, \h) {C};
\node[left]  at (0, 0.5*\h)  {D};
\draw[thick, ->] (0.3*\h, 0.1*\h) -- (0.7*\h, 0.1*\h);
\draw[thick, ->] (0.9*\h, 0.3*\h) -- (0.9*\h, 0.7*\h);
\draw[thick, ->] (0.7*\h, 0.9*\h) -- (0.3*\h, 0.9*\h);
\draw[thick, ->] (0.1*\h, 0.7*\h) -- (0.1*\h, 0.3*\h);

\begin{scope}[shift={(1.7*\h, 0)}]
\def\a{1.4};
\begin{axis}[
xmin = 0.0,
xmax = 4,
ymin = -\a,
ymax =  \a,
xtick = {0.5, 1.5, 2.5, 3.5},
xticklabels={A, B, C, D},
xtick style={draw=none},
ytick = {-1, 0, 1},
width = 6.4cm,
height = 3.2cm,
]
\addplot[blue,  line width=0.8] table[x=xx, y=s1] {./fig/samples.txt};
\addplot[green, line width=0.8] table[x=xx, y=s2] {./fig/samples.txt};
\addplot[black, line width=0.8] table[x=xx, y=s3] {./fig/samples.txt};
\addplot[brown, line width=0.8] table[x=xx, y=s4] {./fig/samples.txt};
\draw[thick, dashed] (axis cs: 1,-\a) -- (axis cs: 1,\a);
\draw[thick, dashed] (axis cs: 2,-\a) -- (axis cs: 2,\a);
\draw[thick, dashed] (axis cs: 3,-\a) -- (axis cs: 3,\a);
\end{axis}
\end{scope}
\node at (5*\h, 0) {};
\node[above right] at (0, -3.3*\h) {\includegraphics[width=7cm]{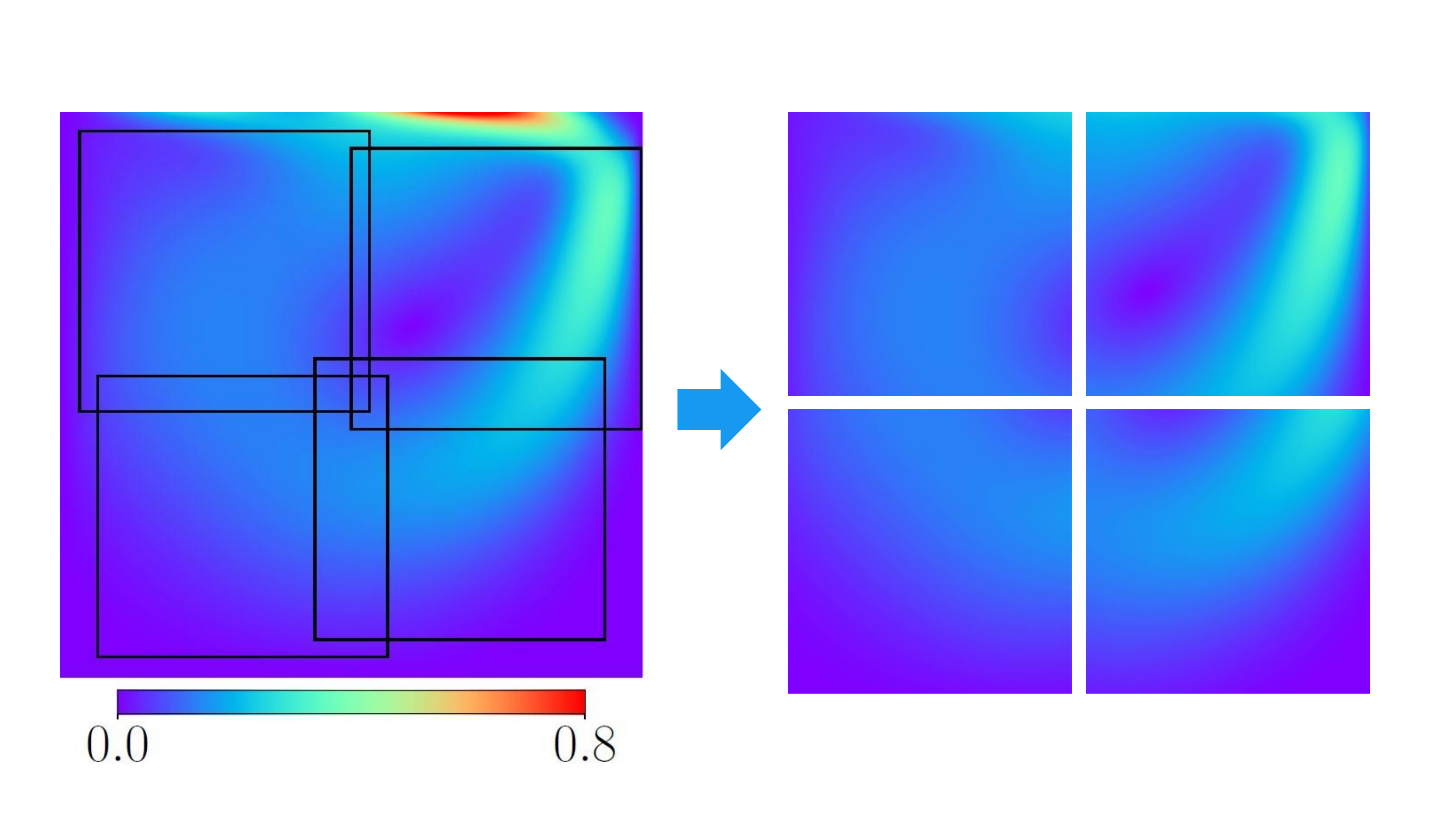}};
\node at (2.8*\h, -0.6*\h) {
	Method 1: \parbox[t]{6cm}{sample genome's BCs with GP.}
};
\node at (2.6*\h, -3.4*\h) {
	Method 2: \parbox[t]{6cm}{extract genomes from flow solutions.}
};

\end{tikzpicture}

%% file: fig/gfnet_arch.tex
\begin{tikzpicture}[
	every neuron/.style={
	circle,
	draw,
	thick,
	minimum size=2mm,
	inner sep=0,
},
    position/.style args={#1:#2 from #3}{
	at=(#3.#1), anchor=#1+180, shift=(#1:#2)
},
>=stealth,
]
\def\w{0.4}
\def\h{2.8}
\def\g{1.2}

\begin{scope}
\draw[fill=pink,   fill opacity=0.4] (0, 0) rectangle (\w, 0.2*\h);
\draw[fill=violet, fill opacity=0.4] (0, 0.2*\h) rectangle (\w, \h);
\draw[fill=cyan,   fill opacity=0.4] (\g, 0.15*\h) rectangle (\g+\w,0.85*\h);
\draw[fill=cyan,   fill opacity=0.4] (2.5*\g, 0.3*\h) rectangle (2.5*\g+\w,0.7*\h);
\node[every neuron] (o) at (3.5*\g, 0.5*\h) {};
\node[] at (4.3*\g, 0.48*\h) {$\nn(\pmb{g},\pmb{x}|\hp)$};

\node[every neuron] (i-1) at (0.5*\w, 0.05*\h) {};
\node[every neuron] (i-2) at (0.5*\w, 0.15*\h) {};
\node[every neuron] (i-3) at (0.5*\w, 0.25*\h) {};
\node[every neuron] (i-4) at (0.5*\w, 0.35*\h) {};
\node[every neuron] (i-5) at (0.5*\w, 0.85*\h) {};
\node[every neuron] (i-6) at (0.5*\w, 0.95*\h) {};
\node at (0.5*\w, 0.7*\h) {\scalebox{2}{$\vdots$}};

\node[every neuron] (h-1-1) at (0.5*\w+\g, 0.2*\h) {};
\node[every neuron] (h-1-2) at (0.5*\w+\g, 0.3*\h) {};
\node[every neuron] (h-1-3) at (0.5*\w+\g, 0.7*\h) {};
\node[every neuron] (h-1-4) at (0.5*\w+\g, 0.8*\h) {};
\node at (0.5*\w+\g, 0.58*\h) {\scalebox{2}{$\vdots$}};

\foreach \m in {1, 2, 3, 4, 5, 6}
	\foreach \n in {1, 2, 3, 4}
		\draw (i-\m) -- (h-1-\n);

\node at (1.8*\g+0.5*\w, 0.5*\h) {\scalebox{2}{$\cdots$}};
\node [every neuron] (h-2-1) at (2.5*\g+0.5*\w, 0.65*\h) {};
\node [every neuron] (h-2-2) at (2.5*\g+0.5*\w, 0.55*\h) {};
\node [every neuron] (h-2-3) at (2.5*\g+0.5*\w, 0.45*\h) {};
\node [every neuron] (h-2-4) at (2.5*\g+0.5*\w, 0.35*\h) {};

\foreach \m in {1, 2, 3, 4}
		\draw (h-2-\m) -- (o);

\node[right] at (-1.2*\w, 0.6*\h) {$\pmb{g}$};
\node[right] at (-1.2*\w, 0.1*\h) {$\pmb{x}$};
\end{scope}

\begin{scope}[shift={(0, -1.2*\h)}]

\draw[fill=pink,   fill opacity=0.4] (0, 0.4*\h) rectangle (\w, 0.6*\h);
\node at (0.5*\w, 0.3*\h) {\large $\pmb{x}$};
\node[every neuron] (i-1) at (0.5*\w, 0.55*\h) {};
\node[every neuron] (i-2) at (0.5*\w, 0.45*\h) {};

\draw[fill=cyan,   fill opacity=0.4] (\g, 0.3*\h) rectangle (\g+\w,0.7*\h);
\node[every neuron] (h-1-1) at (0.5*\w+\g, 0.35*\h) {};
\node[every neuron] (h-1-2) at (0.5*\w+\g, 0.45*\h) {};
\node[every neuron] (h-1-3) at (0.5*\w+\g, 0.55*\h) {};
\node[every neuron] (h-1-4) at (0.5*\w+\g, 0.65*\h) {};

\foreach \m in {1, 2}
\foreach \n in {1, 2, 3, 4}
\draw (i-\m) -- (h-1-\n);

\node at (1.7*\g+0.5*\w, 0.5*\h) {\scalebox{2}{$\cdots$}};

\draw[fill=cyan,   fill opacity=0.4] (2.5*\g, 0) rectangle (2.5*\g+\w,\h);
\node[every neuron] (h-2-1) at (2.5*\g+0.5*\w, 0.05*\h) {};
\node[every neuron] (h-2-2) at (2.5*\g+0.5*\w, 0.15*\h) {};
\node[every neuron] (h-2-3) at (2.5*\g+0.5*\w, 0.25*\h) {};
\node[every neuron] (h-2-4) at (2.5*\g+0.5*\w, 0.75*\h) {};
\node[every neuron] (h-2-5) at (2.5*\g+0.5*\w, 0.85*\h) {};
\node[every neuron] (h-2-6) at (2.5*\g+0.5*\w, 0.95*\h) {};
\node at (2.5*\g+0.5*\w, 0.55*\h) {\scalebox{2}{$\vdots$}};
\node at (2.5*\g+0.5*\w, -0.1*\h) {$\pmb{h}$}; 

\draw[fill=violet, fill opacity=0.4] (3*\g, 0) rectangle (3*\g+\w, \h);
\node[every neuron] (g-1) at (3*\g+0.5*\w, 0.05*\h) {};
\node[every neuron] (g-2) at (3*\g+0.5*\w, 0.15*\h) {};
\node[every neuron] (g-3) at (3*\g+0.5*\w, 0.25*\h) {};
\node[every neuron] (g-4) at (3*\g+0.5*\w, 0.75*\h) {};
\node[every neuron] (g-5) at (3*\g+0.5*\w, 0.85*\h) {};
\node[every neuron] (g-6) at (3*\g+0.5*\w, 0.95*\h) {};
\node at (3*\g+0.5*\w, 0.55*\h) {\scalebox{2}{$\vdots$}};
\node at (3*\g+0.5*\w, -0.1*\h) {\large $\pmb{g}$};

\foreach \m in {1, 2, 3, 4, 5, 6}
\draw (h-2-\m) -- (g-\m);	

\node[every neuron] (o) at (3.5*\g+0.5*\w, 0.5*\h) {};
\node at (4*\g+0.5*\w,0.5*\h) {$\pmb{g}\cdot\pmb{h}$};
\node at (4.5*\g+0.5*\w,0.5*\h) {};

\foreach \m in {1, 2, 3, 4, 5, 6}
\draw (g-\m) -- (o);
\end{scope}

\node at (2*\g, -0.06*\h) {Fully Connected};
\node at (2*\g, -1.5*\h) {Linearity-Preserving Fully Connected};

\end{tikzpicture}

%% file: fig/mf_predict_genome.tex
\begin{tikzpicture}

\def\a{2.2};
\def\aa{0.8}; 
\def\b{0.8};  


\draw[very thick] (\aa +0, 0) rectangle +(\a, \a);

\draw[very thick, cyan, dotted, fill = cyan, opacity= 0.1 ] (\aa +0, 0) rectangle +(\a/2, \a/2);
\draw[very thick, cyan, dotted ] (\aa +0, 0) rectangle +(\a/2, \a/2);
\draw[very thick, cyan, dotted, fill = cyan, opacity= 0.1 ] (\aa +\a/2, \a/2) rectangle +(\a/2, \a/2);
\draw[very thick, cyan, dotted ] (\aa +\a/2, \a/2) rectangle +(\a/2, \a/2);
\draw[very thick, cyan, dotted, fill = cyan, opacity= 0.1 ] (\aa +\a/2, 0) rectangle +(\a/2, \a/2);
\draw[very thick, cyan, dotted ] (\aa +\a/2, 0) rectangle +(\a/2, \a/2);
\draw[very thick, cyan, dotted , fill = cyan, opacity= 0.1 ] (\aa +0, \a/2) rectangle +(\a/2, \a/2);
\draw[very thick, cyan, dotted ] (\aa +0, \a/2) rectangle +(\a/2, \a/2);

\node at (\aa +0.25*\a, 0.25*\a) {$g_1$};
\node at (\aa +0.75*\a, 0.25*\a) {$g_2$};
\node at (\aa +0.25*\a, 0.75*\a) {$g_3$};
\node at (\aa +0.75*\a, 0.75*\a) {$g_4$};


\draw [very thick, ->] (\aa +\a+0.2*\b, 0.5*\a) -- (\aa +\a+0.8*\b, 0.5*\a);

\draw[very thick] (\aa +\a+\b, 0) rectangle +(\a, \a);
\draw[very thick, dashed, red, fill = red, opacity = 0.1] (\aa +1.25*\a+\b, 0.5*\a) rectangle +(0.5*\a, 0.5*\a);
\draw[very thick, dashed, red] (\aa +1.25*\a+\b, 0.5*\a) rectangle +(0.5*\a, 0.5*\a);
\node at (\aa +1.5*\a+\b, 0.25*\a) {$g_5$};
\draw[very thick, dashed, red, fill = red, opacity = 0.1] (\aa +1.25*\a+\b, 0) rectangle +(0.5*\a, 0.5*\a);
\draw[very thick, dashed, red] (\aa +1.25*\a+\b, 0) rectangle +(0.5*\a, 0.5*\a);
\node at (\aa +1.5*\a+\b, 0.75*\a) {$g_6$};

\draw [very thick, ->] (\aa +2*\a+1.2*\b, 0.5*\a) -- (\aa +2*\a+1.8*\b, 0.5*\a);

\draw[very thick] (\aa +2*\a+2*\b, 0) rectangle +(\a, \a);
\draw[very thick, dashed, red, fill =red, opacity = 0.1] (\aa +2*\a+2*\b, 0.25*\a) rectangle +(0.5*\a, 0.5*\a);
\draw[very thick, dashed, red] (\aa +2*\a+2*\b, 0.25*\a) rectangle +(0.5*\a, 0.5*\a);
\node at ((\aa +2.25*\a+2*\b, 0.5*\a) {$g_7$};
\draw[very thick, dashed, red, fill =red, opacity = 0.1] (\aa +2.5*\a+2*\b, 0.25*\a) rectangle +(0.5*\a, 0.5*\a);
\draw[very thick, dashed, red] (\aa +2.5*\a+2*\b, 0.25*\a) rectangle +(0.5*\a, 0.5*\a);
\node at ((\aa +2.75*\a+2*\b, 0.5*\a) {$g_8$};

\draw [very thick, ->] (\aa +3*\a+2.2*\b, 0.5*\a) -- (\aa +3*\a+2.8*\b, 0.5*\a);

\draw[very thick] (\aa +3*\a+3*\b, 0) rectangle +(\a, \a);
\draw[very thick, dashed, red, fill = red, opacity = 0.1] (\aa +3.25*\a+3*\b, 0.25*\a) rectangle +(0.5*\a, 0.5*\a);
\draw[very thick, dashed, red] (\aa +3.25*\a+3*\b, 0.25*\a) rectangle +(0.5*\a, 0.5*\a);
\node at (\aa +3.5*\a+3*\b, 0.5*\a) {$g_9$};

\draw[very thick, ->] 
(\aa +3.5*\a+3*\b,0) -- (\aa +3.5*\a+3*\b,-0.15*\a) 
--node[below] {\large Iterate} (\aa +0.5*\a, -0.15*\a) -- (\aa +0.5*\a, 0);

\node[cyan] at (\aa +0.5*\a, 1.12*\a) {Basic genomes};
\node[red] at (\aa +2.5*\a+2*\b, 1.12*\a) {Auxiliary genomes};

\def\c{0.6*\b}; 
\draw[very thick, dashed, red, fill=red, fill opacity=0.1] 
(\aa+4*\a+4*\b+\c, 0.25*\a) rectangle +(0.5*\a,0.5*\a);
\node[left] at (\aa+4*\a+4*\b+\c, 0.5*\a) {$g_7$};
\draw[very thick, dotted, cyan, fill=cyan, fill opacity=0.1] 
(\aa+4*\a+4*\b+\c, 0) rectangle +(0.5*\a,0.5*\a);
\node[left] at (\aa+4*\a+4*\b+\c, 0.1*\a) {$g_1$};
\draw[very thick, dotted, cyan, fill=cyan, fill opacity=0.1] 
(\aa+4*\a+4*\b+\c, 0.5*\a) rectangle +(0.5*\a,0.5*\a);
\node[left] at (\aa+4*\a+4*\b+\c, 0.9*\a) {$g_3$};
\draw[very thick, dashed, red, fill=red, fill opacity=0.1] 
(\aa+4.25*\a+4*\b+\c, 0.25*\a) rectangle +(0.5*\a,0.5*\a);
\node[right] at (\aa+4.75*\a+4*\b+\c, 0.5*\a) {$g_9$};
\node[below] at (\aa+4.4*\a+4*\b+\c, -0.05*\a) {
\parbox[t]{4cm}{For $g_7$, MF predictor only infers its center lines.}};


\def\a{2.7};
\def\c{-1.4*\a}; 
\def\d{1.7cm}; 

\def\f{0.2cm}; 
\def\g{1.0cm}; 
\def\i{0.1cm}; 
\def\j{0.1cm}; 
\def\h{1.2cm}; 
\def\e{-0.05*\a}; 
\def\w{1.5cm}; 
\def\ce{0.9} 
\node[anchor=south west,inner sep=0] (f0) at ( 0.1 + \e, \c) {
  \includegraphics[width=\w]{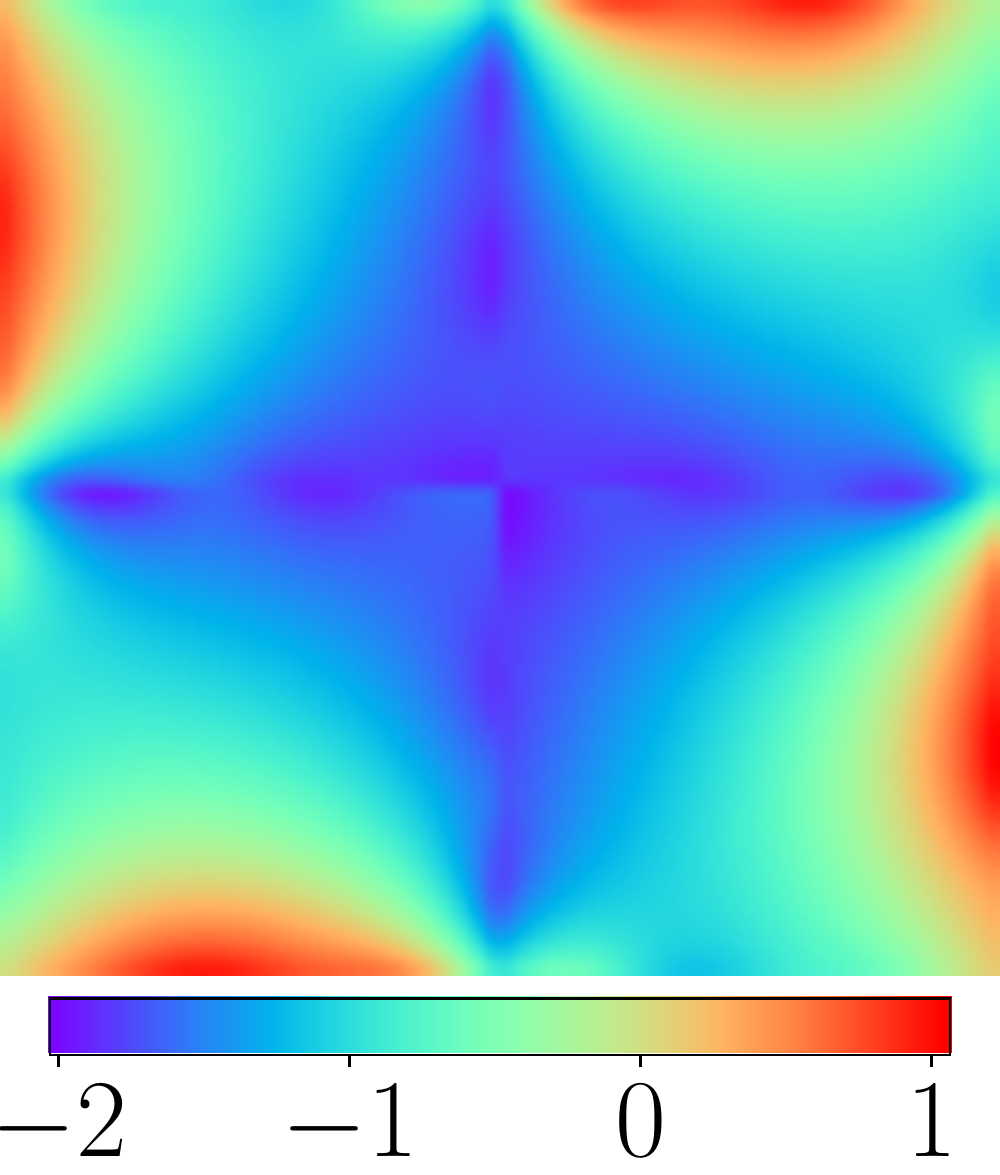}
};
\node[anchor=south west,inner sep=0] (f1) at (\e+\d, \c) {
  \includegraphics[width=\w]{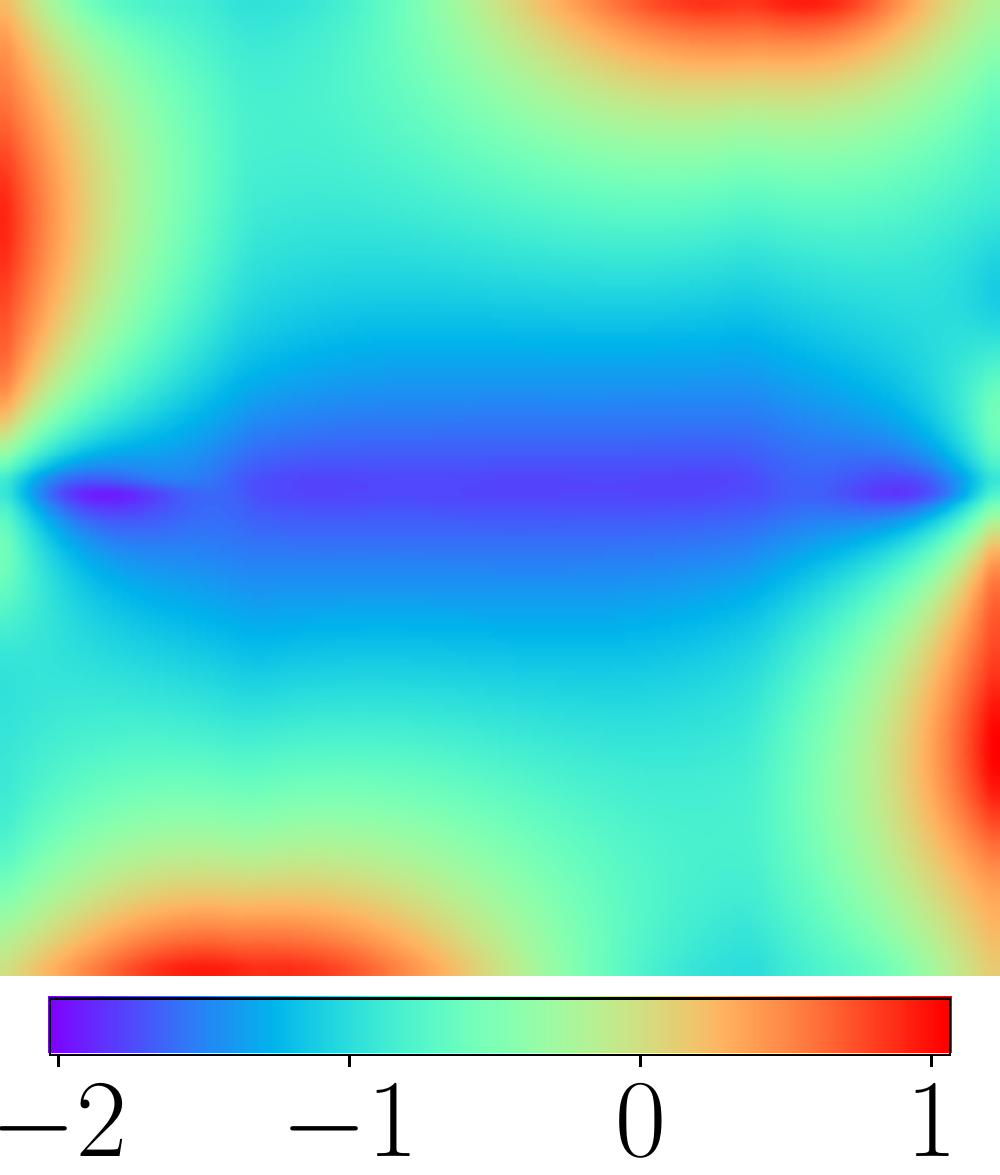}
};
\node[anchor=south west,inner sep=0] (f2) at (\e+2*\d, \c) {
  \includegraphics[width=\w]{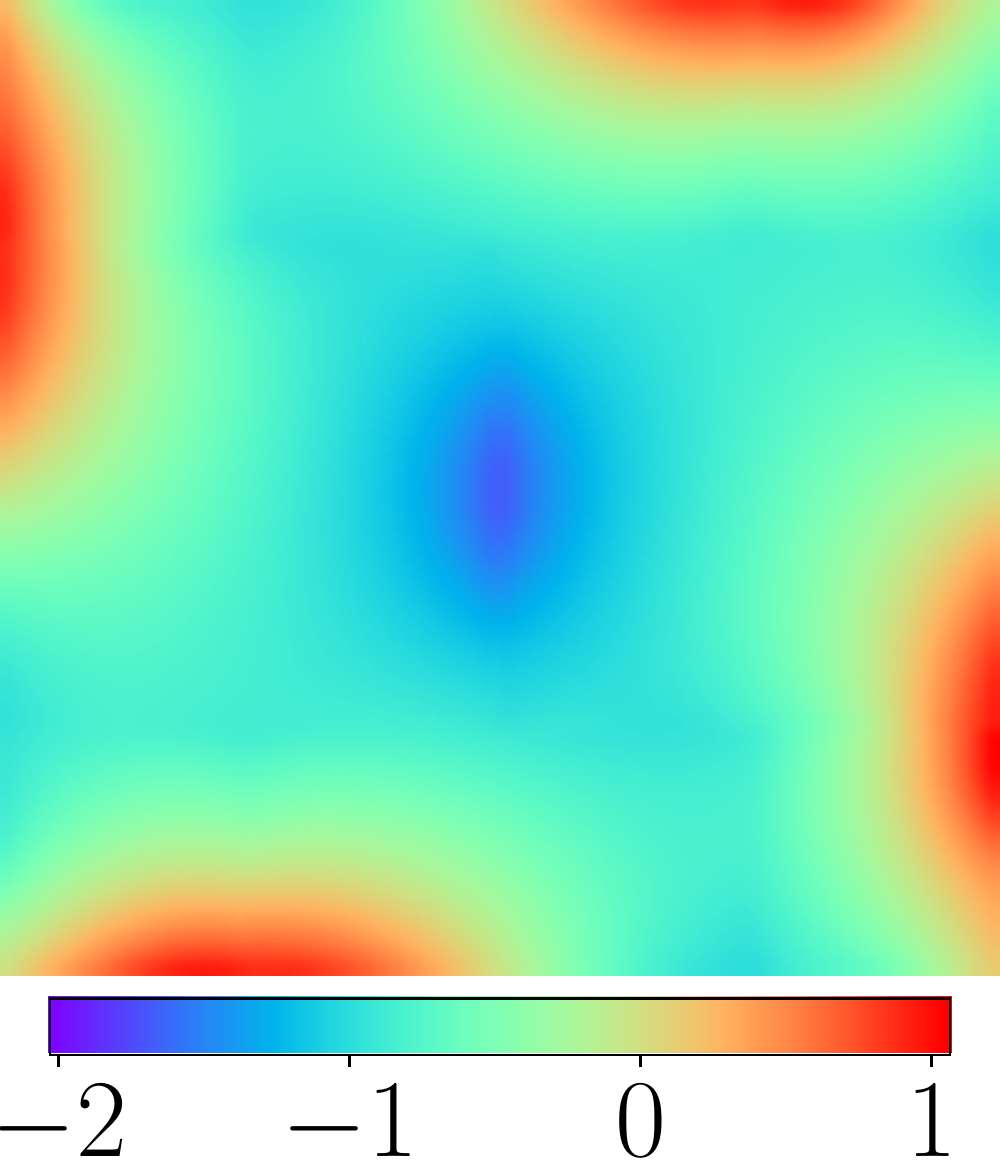}
};
\node[anchor=south west,inner sep=0] (f3) at (\e+3*\d, \c) {
  \includegraphics[width=\w]{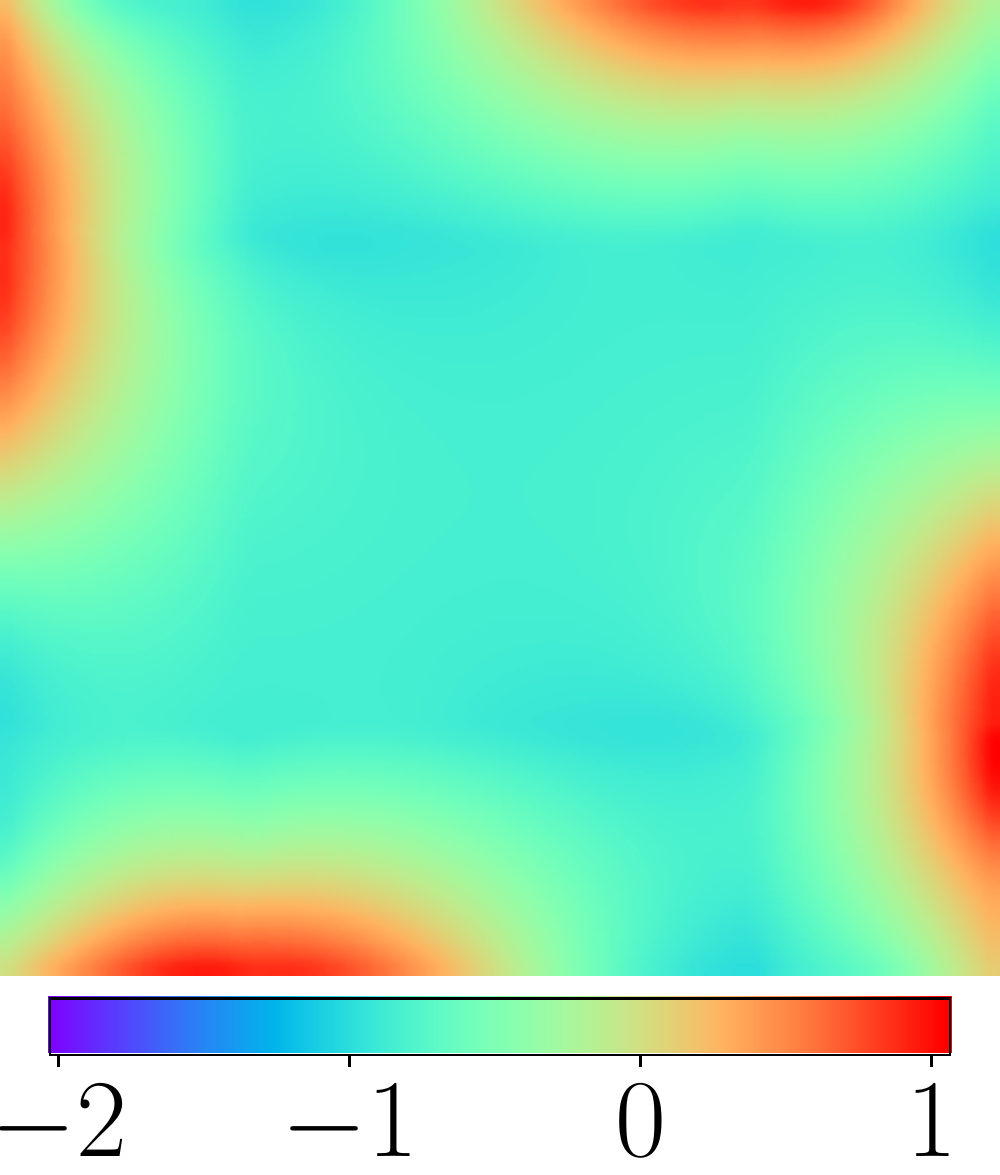}
};
\node[anchor=south west,inner sep=0] (f4) at (\e+4*\d + \f, \c) {
  \includegraphics[width=\w]{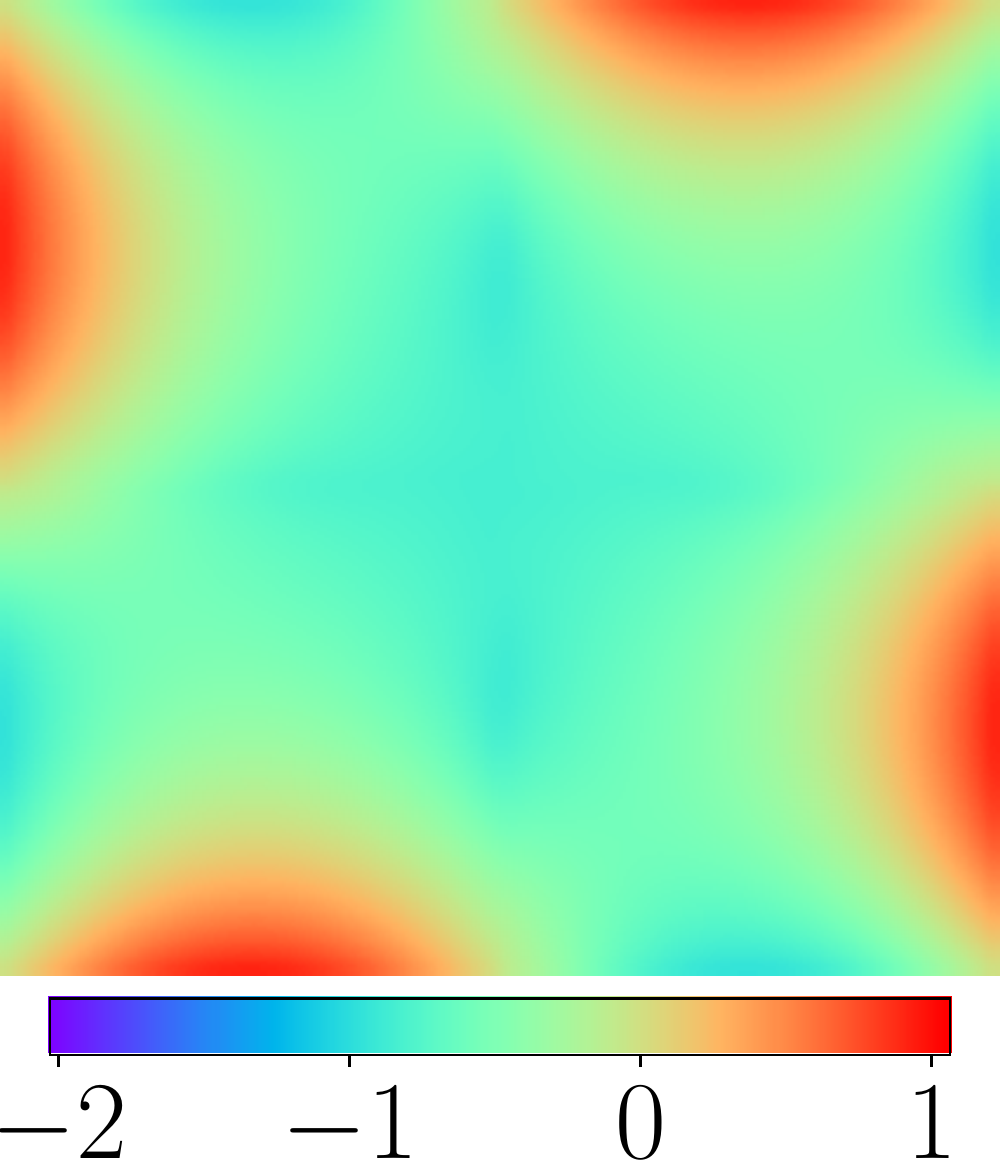}
};
\node[anchor= west,inner sep=0] (f5) at (\e+5*\d + \f + \i, \c + \ce)
{\raisebox{3mm}{\scalebox{1.6}{$\cdots$}}};

\node[anchor=south west,inner sep=0] (f6) at (\e+5*\d + \g + \f + \i, \c) {
  \includegraphics[width=\w]{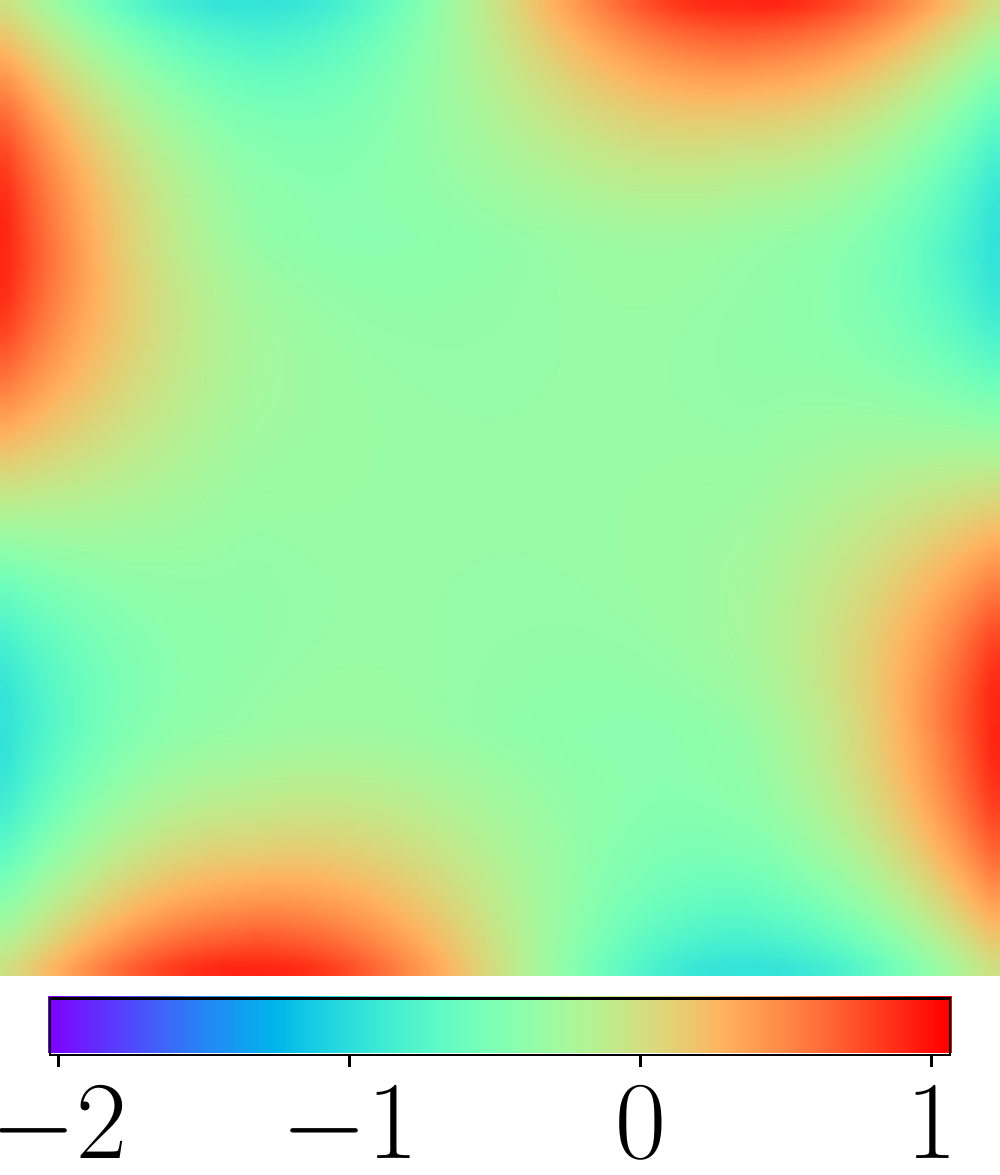}
};
\node[anchor= west,inner sep=0] (f7) at (\e+6*\d + 2*\f + \g + \i + \j, \c + \ce)
{$\;$\raisebox{2.2mm}{\scalebox{2}{$\cdots$}}};

\node[anchor=south west,inner sep=0] (f8) at (\e+6*\d+ 2*\f + \g +\h + \i + \j, \c) {
  \includegraphics[width=\w]{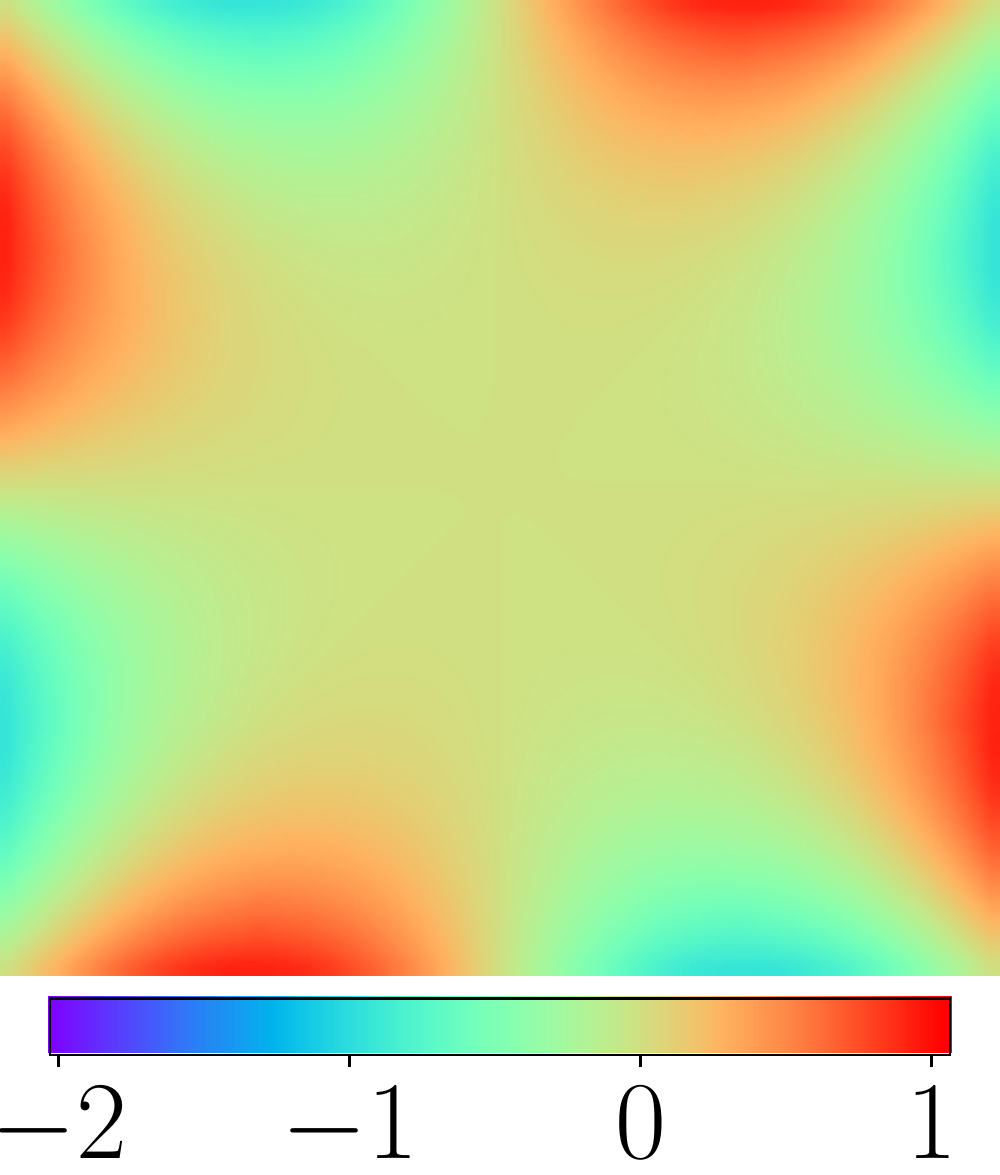}
};
\node[anchor=south west,inner sep=0] (f9) at (\e+7*\d + 2*\f +  \g + \h + \i + \j, \c) {
  \includegraphics[width=\w]{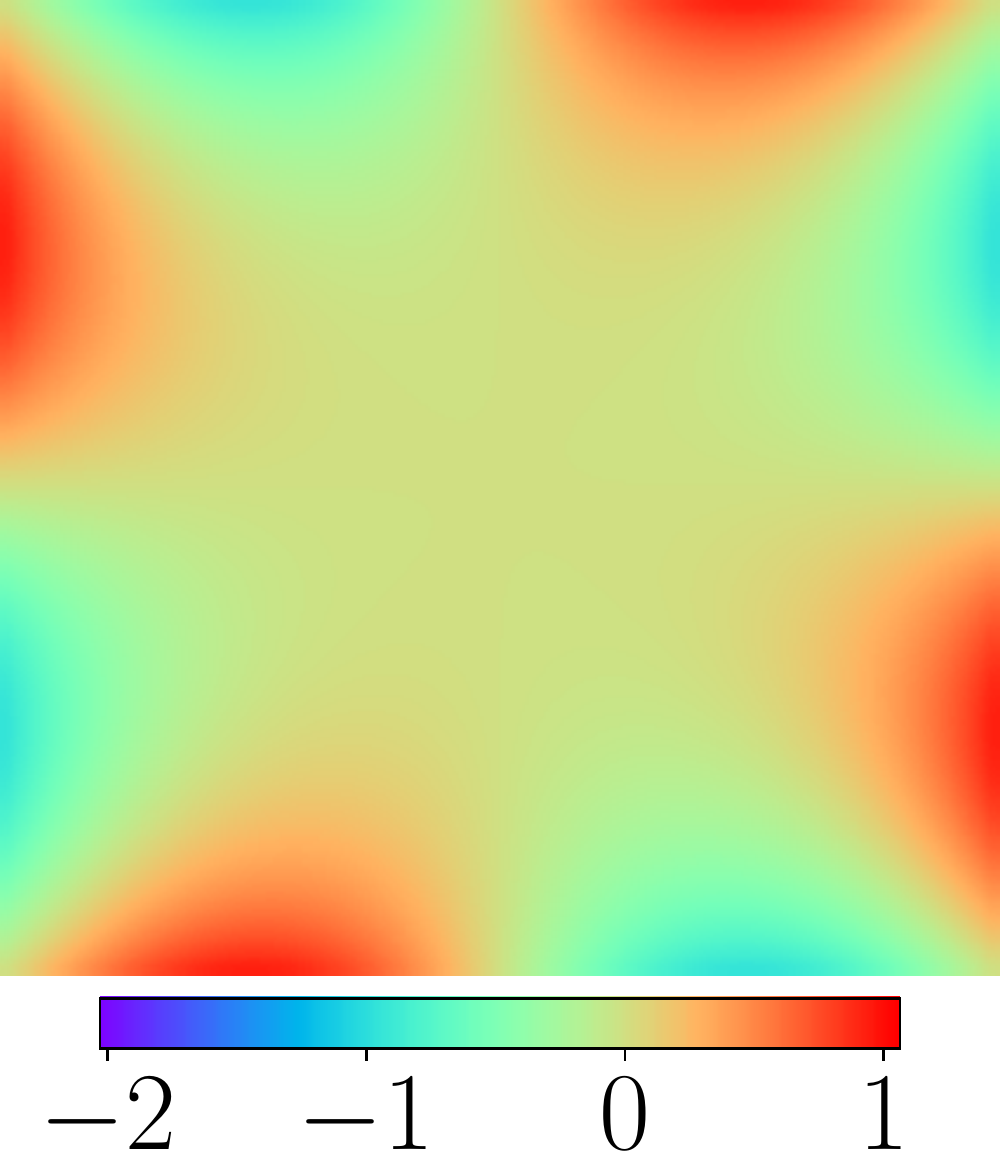}
};

\draw[very thick, ->, transform canvas={yshift=2mm}] (f0) -- (f1);
\draw[very thick, ->, transform canvas={yshift=2mm}] (f1) -- (f2);
\draw[very thick, ->, transform canvas={yshift=2mm}] (f2) -- (f3);
\draw[line width = 1mm, ->, transform canvas={yshift=2mm}] (f3) -- (f4);
\draw[very thick, ->, transform canvas={yshift=2mm}] (f4) -- (f5);
\draw[very thick, ->, transform canvas={yshift=2mm}] (f5) -- (f6);
\draw[line width = 1mm, ->, transform canvas={yshift=2mm}]
(f6) -- (f7);

\draw [|-|,  transform canvas={yshift=1cm}] 
(f0.west) -- node[above] {Iteration 1} (f3.east);
\draw [|-|,  transform canvas={yshift=1cm}] 
(f4.west) -- node[above] {Iteration 2} (f6.east);
\node[black] at (\e+6*\d+ 2*\f + \g + \h + \i + \j + \w/2, \c + 2.3*\ce) { Solved };
\node[black] at (\e+7*\d+ 2*\f + \g + \h + \i + \j + \w/2, \c + 2.3*\ce) { Exact };

\end{tikzpicture}

%% file: eval.tex
\section{Results and Discussions}\label{sec:eval}
We evaluate GFNet and \mfpShort by solving the linear Laplace equation and the non-linear incompressible NS equations in Sections \ref{sec:laplace} and \ref{sec:ns}, respectively. In each section, we first present the details on data generation, GFNet architecture, accuracy, and performance (the latter refers to computational costs).
Then, we compare GFNet and \mfpShort against the state-of-the-art PINNs. 
\revise{In Section \ref{sec:Error_Analysis} we do an in-depth error analysis of our framework and test its accuracy when the Fourier neural operator (FNO) \cite{li2020neural} and deep operator networks (DeepONet) \cite{Lu2019} are used as GFNets and in Section \ref{sec:exact_bc} we demonstrate the advantage of enforcing the network architecture to satisfy the applied BCs.} 

We use Tensorflow \cite{tensorflow2015-whitepaper} and ADAM \cite{Kingma2015} optimizer. 
We implement GFNet with TensorFlow 2 (2.2.0) whereas the PINN related sources \cite{Raissi2019, Wang2020a} used for comparison are built with TensorFlow 1 (1.8.0). \revise{In Section \ref{sec:Error_Analysis}, we implement FNO with Pytorch 1.9.0 and DeepONet with Tensorflow 2 (2.4.0). }
For our models, we initialize the learning rate, $\eta$, by 5e-4 and reduce it by $20\%$ when the validation loss decreases by less than $0.01\%$ across $200$ consecutive epochs. We use a $9:1$ ratio for splitting the data into training and validation samples and terminate the training when $\eta=$ 1e-7. We measure accuracy via mean absolute error (MAE) and mean absolute residual (MAR) metrics calculated as:
\begin{align*}
    \text{MAE:} \, \sum^{N_g}_{i=1}|\nn(\pmb{g},\pmb{x}_i|\hp) - u(\pmb{x}_i)|/N_g \hspace{1cm}\;
    \text{MAR:} \, \sum^{N_g}_{i=1}| H\left(\nn(\pmb{g},\pmb{x}_i|\hp)\right)|/N_g.
\end{align*}
All the simulations are done on an NVIDIA Quadro RTX 8000 GPU.

\begin{figure*}[!b]
	\centering
	\subfloat[Boundary and data points\label{fig:laplace_d}]{\includegraphics[scale=0.5]{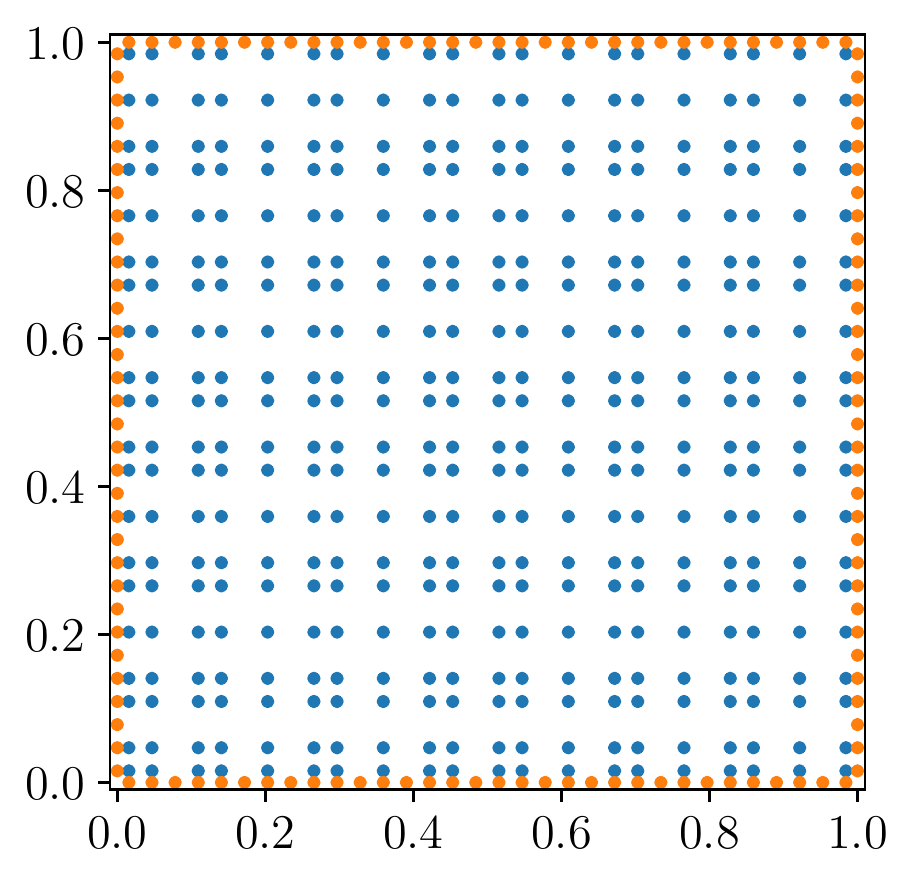}}\hfil
	\subfloat[GFNet's gradient magnitude, $|\nabla\mathcal{N}|$\label{fig:laplace_g}]
	{\includegraphics[scale=0.5]{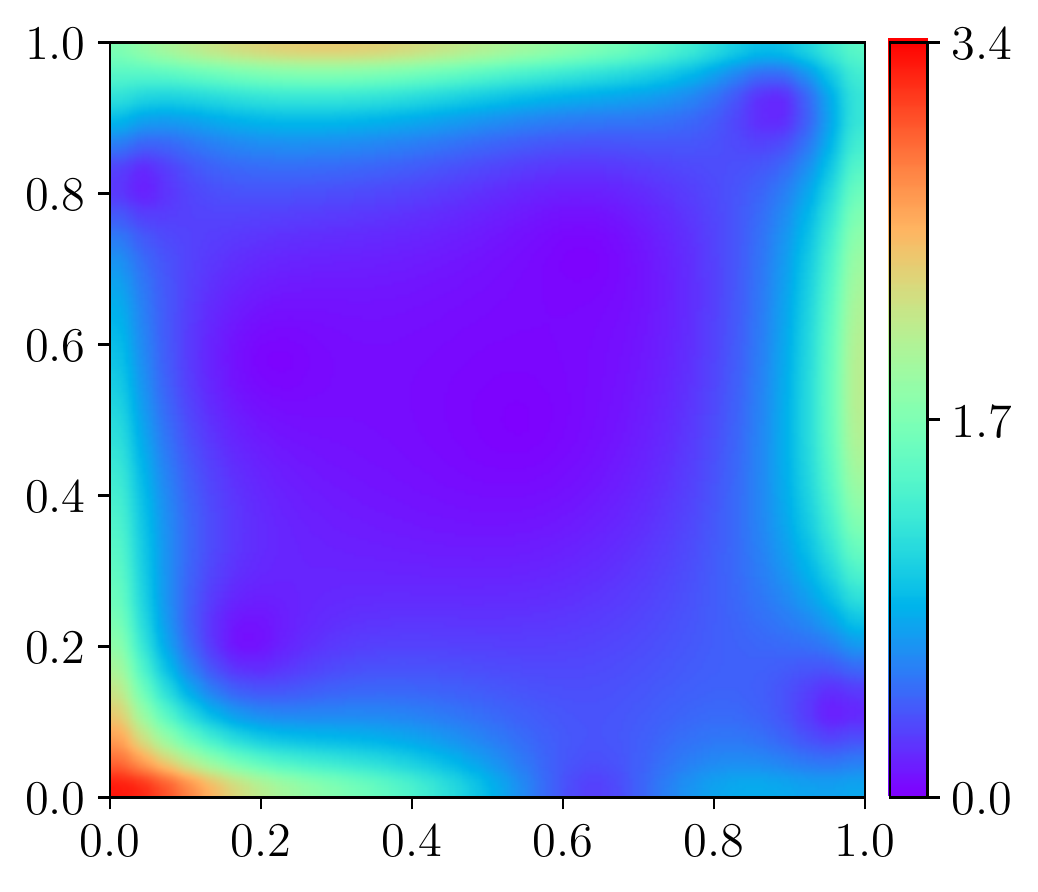}}\hfil
	\subfloat[Collocation points\label{fig:laplace_col}]
	{\includegraphics[scale=0.5]{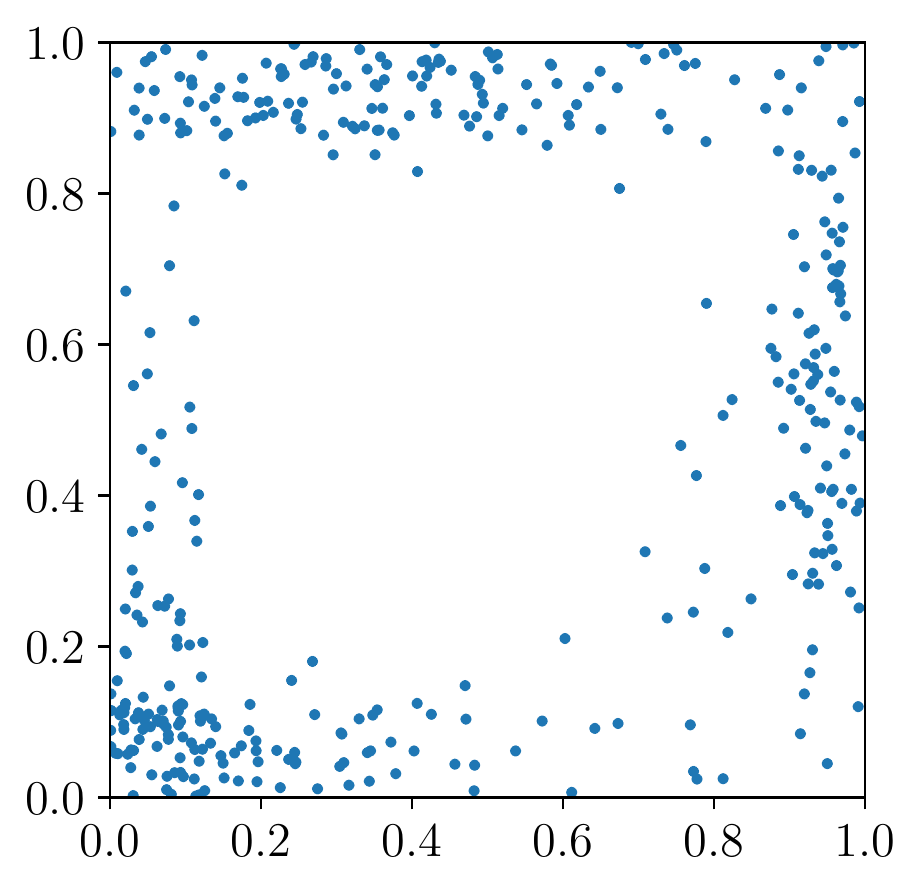}}
	\caption{Distribution of training data in $\Omega$: 128 boundary points, 400 data points, and 400 collocation points. The locations of boundary and data points are fixed during training while those of the collocation points adaptively change based on $|\nabla\mathcal{N}|$.}
	\label{fig:laplace_points}
\end{figure*}

\subsection{Laplace Equation}\label{sec:laplace}
Laplace equation is \revise{an elliptic} BVP that ubiquitously appears in fluid dynamics and heat transfer \cite{RN1440}, see Equation \ref{eqn:ex_laplace}. In this section, we learn the $2D$ Laplacian operator via a GFNet in a unit square domain, i.e., the genome covers $[0,1]\times[0, 1]$. Then, we use \mfpShort to solve the Laplace equation in domains that are up to $1200$ times larger. We compare the accuracy and performance of GFNet and \mfpShort against PINN \cite{Raissi2019} and XPINN \cite{Jagtap2020}.
\subsubsection{\textbf{Training Data.}}~
As described in Section \ref{sec:nn_bvp}, we use GPs to generate BCs. In particular, we use Sobol sequence \cite{RN114} to sample the hyperparameters of the infinitely differentiable Gaussian kernel of a $1D$ GP. Then, we draw a sample function (i.e., a $1D$ curve) from each GP and wrap it around the genome, see Figure \ref{fig:workflow}a. Finally, for each BC, we numerically solve the Laplace equation via PyAMG \cite{OlSc2018} which relies on FD to provide the solution on a uniform grid that consists of $32\times 32$ cells and 128 boundary points, see Figure \ref{fig:laplace_d}. 

\subsubsection{\textbf{Architecture and Training of GFNet.}}~
As defined in Section \ref{sec:nn_bvp}, GFNet's inputs include $\pmb{x}$ and the discretized boundary value function $\pmb{g}$ which in this case has $N_{bc}=128$ points. Since Laplace equation has one scalar variable (i.e., $N_{var}=1$), GFNet has a total of $N_{bc}\times N_{var} + N_{dim} = 130$ inputs. The output of GFNet is a single value $\nn(\pmb{g},\pmb{x}|\hp)$ that approximates $u(\pmb{x})$.

We choose $\tanh$ as the activation function and evaluate both FC and LPFC architectures introduced in Section \ref{sec:nn_bvp}. The loss function of GFNet is defined in Equation \ref{eqn:ex_pinn_loss} and uses $\beta=0$ and $\alpha=$1e-3. \revise{When minimizing this loss, three data types are used in each epoch: boundary points, data points , and collocation points.}
While the location of boundary and data points are determined by PyAMG (and hence fixed), we adaptively choose the location of collocation points during training based on the spatial gradients that GFNet estimates inside the genome. These gradients are estimated via AD and their magnitude, $|\nabla\mathcal{N}|$, controls the spatial distribution of the collocation points by increasing the point density in regions where $|\nabla\mathcal{N}|$ is large. Figure \ref{fig:laplace_g} demonstrates the spatial distribution of these three data types in one randomly selected epoch during training. In this figure, boundary and data points are on a grid (since PyAGM generates them based on FD) while collocation points concentrate near the genome's border where the solution tends to change sharply.

\subsubsection{\textbf{Accuracy for Unseen BCs.}}~
We train the FC and LPFC architectures using $2000$ samples where each sample consists of $128$ boundary, $100$ data, and $400$ collocation points. The FC architecture consists of $8$ hidden layers of sizes $128^2\otimes96^2\otimes64^2\otimes32^2$ which make a cone as shown in Figure \ref{fig:step2}. The notation $k^i\otimes l^j$ denotes $i$ FC layers of size $k$ followed by $j$ FC layers of size $l$. We invert the order of these hidden layers for LPFC so that both architectures have the same number of neurons. The resulting GFNets are tested on $400$ unseen samples generated by PyAMG where the MAE and MAR are evaluated at all the $32\times32$ grid cells.

\begin{table*}[!b]
	\centering
	\caption{Accuracy of GFNet on learning the Laplacian operator: LPFC architecture learns the Laplacian operator more accurately than FC. The training column includes the number of layers, samples, data/collocation points, and the mean absolute error (MAE). The test column reports the MAE and the mean absolute residual (MAR) which are calculated by evaluating GFNet's predictions against PyAMG \cite{OlSc2018}.}
	\input{./tab/laplace_train}
	\label{tab:laplace_train}
\end{table*}

The first two rows in Table \ref{tab:laplace_train} compare the accuracy of the two architectures. We observe that LPFC learns the Laplacian operator more accurately than FC by one order of magnitude. This superior accuracy is because FC applies a non-linear activation function ($\tanh$) to $\pmb{g}$ which conflicts with the linearity of the Laplace equation. On the contrary, LPFC applies the non-linear activation function only to $\pmb{x}$ and retains the linearity of the Laplacian operator. Therefore, we use the LPFC architecture in the remainder of this section.

We further improve GFNet's accuracy by increasing its depth and the size of training data. 
(To reduce training costs, we do not use all the $32\times32$ points where PyAMG provides the solution.)
In particular, we use 14 layers of size $32^2\otimes64^2\otimes96^5\otimes128^5$ and $8000$ samples to reduce the MAE and MAR of LPFC by $1.3$ and $1.8$ times, respectively.
Moreover, LPFC consistently outperforms FC in every case and exhibits 2 times higher accuracy with 8000 samples.
Further increasing the number of samples or hidden layers did not noticeably affect accuracy. Therefore, we select LPFC architecture and use the $14$-layer GFNet trained with $18000$ samples as the final model in MF predictor.
We also test single and double-precision computations but no significant difference in accuracy is observed for either GFNet (see Table \ref{tab:laplace_train}) or \mfpShort (see Figure \ref{fig:PINNvsGenomes}). Since the former is $17\%$ faster, we choose the model trained with single-precision. 

\subsubsection{\textbf{Accuracy on Unseen Domains Subject to Unseen BCs.}}\label{sec:laplace_accuracy_unseen}
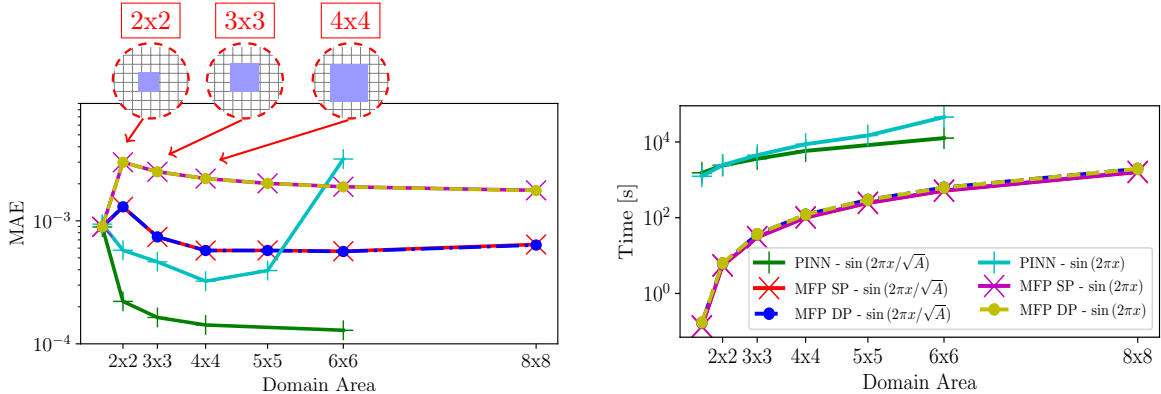
\begin{figure}[!t]
	\centering
	\input{./fig/Laplace_MFNet_vs_PINN}\hfil
	\caption[Inference with overlapped genomes vs PINN]{Accuracy and performance of \mfpShort and PINN for the Laplace equation with two different BCs: For the boundary condition $g_1(s) = \sin(2\pi s / \sqrt{A})$, PINN achieves better accuracy than MF predictor. However, for the more complex BC $g_2(s) = \sin(2\pi s)$, the accuracy of PINN drops, especially in larger domains. Note that while  PINN is re-trained for every new BC, \mfpShort has never seen these BCs. In terms of computational costs (sum of training and inference times), PINN is several orders of magnitude slower than MF predictor (we were unable to train PINN on the $8\times8$ domain due to very high costs). In addition, \mfpShort with single (SP) and double-precision (DP) yields near identical results.}
	\label{fig:PINNvsGenomes}
\end{figure}

~
We first evaluate \mfpShort on square domains of area $A > 1$ with two boundary functions $g_1(s) = \sin(2\pi s/\sqrt{A})$ and $g_2(s) = \sin(2\pi s)$ where $s$ parameterizes the boundary.
We gradually increase $A$ and execute \mfpShort in both single and double-precision.
The accuracy of \mfpShort is measured by the MAE between the converged solution and the ground truth generated with PyAMG. Figure \ref{fig:PINNvsGenomes} summarizes the results.

At $A=1$, the error is solely due to GFNet (i.e., inferring the solution for an unseen BC). For both $g_1(s)$ and $g_2(s)$, the MAE spikes at $A=2\times 2$ as it starts to include additional error accumulated during the iterative process in MF predictor. However, as $A$ increases, the influence of BC on the inner region of the domain reduces, and therefore \mfpShort can more accurately predict the solution. As a result, MAE decays and plateaus once $A>4\times4$.

The converged MAEs for $g_1(s)$ and $g_2(s)$ are different. For $g_1(s)$, the boundary function oscillates less across the domain boundary as $A$ increases which simplifies the genome-wise BVP compared to $A=1$. Therefore, at $A=8\times 8$, even with the additional accumulated error due to MF predictor, the MAE is less than that at $A=1$. However, for $g_2(s)$ the oscillation frequency does not depend on $A$ and hence the MAE at $A=8\times8$ exceeds that at $A=1$ and is also $4$ times larger than the corresponding MAE for $g_1(s)$.

We also demonstrate the transferability and scalability of \mfpShort by solving the Laplace equation in a domain that resembles ``Mosaic Flow" and is $1222.5\times$ larger than the training domain (i.e., a genome), see Figure \ref{fig:sclogo}. The prescribed BC is $g(\pmb{x}) = \sin (2\pi (x/6 + y/5))$ and $2020$ genomes (basic and auxiliary) are used in \mfpShort which converges in $9821$ seconds with an MAE of 2.61e-3.

\revise{In this study, we only consider square genomes for GFNet. 
Accordingly, MF predictor can accurately assemble GFNet's inference for domains spanned by genomes, i.e., with rectilinear boundaries.
However, for non-rectilinear domains such as the above calligraphy shape, we need to adopt a \emph{mosaic} representation of the domain by decomposing the curved boundary to small squares, which results in the zigzag-shaped boundary in Figure \ref{fig:sclogo}.
We will investigate how to accurately consider non-square genomes in our future work.
}

\subsubsection{\textbf{Comparison with PINN and XPINN.}}~
We compare the accuracy and computational costs of GFNet and \mfpShort against a state-of-the-art PINN which has the feed-forward architecture of $60\otimes60\otimes60\otimes60$. To maximize the accuracy of this PINN, we train it with a two-stage optimization process \cite{Jin2021} that starts with $40000$ iterations using ADAM \cite{Kingma2015} and is followed by a second-order optimization method based on L-BFGS \cite{liu1989limited}. We use $20000\times A$ collocation points and $32\times4\times\sqrt{A}$ boundary points to consider the effect of $A$. Contrary to PINN, L-BFGS is not used in fine-tuning GFNet.

The evaluation results are illustrated in Figure \ref{fig:PINNvsGenomes}. 
PINN and GFNet achieve a similar accuracy for a single genome.
For the simpler BC, $g_1(s) = \sin(2\pi x/\sqrt{A})$, PINN achieves smaller MAEs compared to \mfpShort for $A>1$. 
This higher accuracy is because $(1)$ the parameters of PINN are fine-tuned with L-BFGS while GFNet is trained via Adam (GFNet cannot be trained via L-BFGS since it has too many parameters), and $(2)$ PINN is specifically trained on this BC (and cannot be used for any other BC) while \mfpShort assembles the predictions from a GFNet that has never seen this BC.
For the more complex BC, $g_2(s) = \sin(2\pi x)$, PINN fails to scale up to large $A$; indicating that its architecture is domain-specific and must be optimized anew.
Unlike PINN, \mfpShort scales up quite robustly to large domains with unseen BCs without the need to re-train or fine-tune its GFNet. Remarkably, \mfpShort is between 1-3 orders-of-magnitude faster than PINN depending on the domain size, which underscores its potential for scalable inference, especially on large and complex domains (for instance, we were unable to train a PINN on an $8\times8$ domain due to the extremely high training costs). 

Finally, we test \mfpShort against XPINN \cite{Jagtap2020} which is an extension of PINN that aims to address PINN's scalability issues. XPINN divides a domain into some sub-domains and trains a PINN on each one while ensuring continuity across the sub-domains. In our comparison, we choose a square domain of size $4\times4$ subject to the BC: $g_1(s) = \sin(2\pi x/\sqrt{A})$ and divide it into $4$ square sub-domains of equal sizes. The $4$ PINNs used in XPINN have an architecture of $20\otimes20\otimes20\otimes20$ and are trained using Adam for $5000$ epochs with $16000$ collocation points and $128$ boundary points. Table \ref{tab:XPINNvsGenomes} enumerates the accuracy of XPINN and \mfpShort and shows that our approach is two orders of magnitudes more accurate.

\begin{table}[!t]
	\centering
	\vspace{3mm}
	\caption{XPINN vs. MF predictor: Both methods \revise{approximate} the Laplace operator in a square domain of size $4\times4$.}
	\input{./tab/XPINN_res}
	\label{tab:XPINNvsGenomes}
\end{table}

\subsection{Navier-Stokes Equations}\label{sec:ns}
The $2D$ \revise{steady} incompressible NS equations are:
\begin{equation}
    \begin{aligned}
        &H_1: \partial_x u + \partial_y v = 0, \\
        &H_2: u\partial_x u + v\partial_y u + \partial_x p/\rho - \nu\nabla^2 u = 0, \\
        &H_3: u\partial_x v + v\partial_y v + \partial_y p/\rho - \nu\nabla^2 v = 0,
        \label{eqn:ns}
    \end{aligned}
\end{equation}
where $u(\pmb{x}), v(\pmb{x}),$ and $p(\pmb{x})$ denote velocity components and pressure, respectively. We solve Equation \ref{eqn:ns} in a lid-driven cavity problem where the top lid moves with an arbitrary horizontal velocity, i.e., $u=g(x), v=0$, and the remaining boundaries are treated as static viscous walls where $u = v = 0$. 
We fix $\rho=1.0$ and $\nu=0.002$ and aim to predict $u(\pmb{x}), v(\pmb{x}),$ and $p(\pmb{x})$ in unseen domains that have unseen BCs on the top lid and are larger than a genome.
Our imposed lid velocity profiles can be highly nonlinear functions and collectively generate Reynolds numbers in the $(0, 500)$ range. 
The corresponding flow fields are also more complex than cases where the top lid moves with a uniform velocity. 

\revise{The convergence theory in Section \ref{sec:bvp_prelim} does not strictly apply to the NS equations since the non-linear terms in $H_2$ and $H_3$ do not conform to the formulation of the elliptic systems in Equation \ref{eqn:elliptic}.
Nonetheless, the incompressible NS equations can still be approximately viewed as elliptic because the information propagates in the same way as it does in elliptic systems where any fluctuation at one point affects the entire domain.
Several studies have successfully employed Schwarz method with other numerical methods in solving the NS equations and present accurate numerical results \cite{fischer1997overlapping, brakkee2000schwarz, fischer2005hybrid, blayo2016towards}.
In the following sections, we will demonstrate that MF predcitor with GFNet can also learn the NS equations quite accurately and infer flow features that have not been seen in training.
}

\subsubsection{\textbf{Training Data.}}~
We sample the velocity profile of the top lid using GPs and solve the NS equations with OpenFOAM \cite{jasak2007openfoam} on a uniform grid of $129\times 129$ vertices. Then, for each solution field obtained via OpenFOAM, we sweep it with a genome of size $0.5\times 0.5$ and record the values on the boundary and inside the genome, see Figure \ref{fig:step1}. This procedure results in highly correlated training samples that adversely affect GFNet's training. Hence, we ($1$) discard a sample if the Euclidean distance of its BC to another sample's BC, i.e., $|\pmb{g}_i-\pmb{g}_j|$, is small, and ($2$) include cavities of size $1\times 2$ and $2\times 1$. Following this procedure, $2634$ samples are generated.

\subsubsection{\textbf{Architecture and Training of GFNet.}}\label{sec:trainig_NS}~ 
For each sample, \revise{we uniformly extract $128$ boundary points from the grid vertices located on the boundary.} GFNet's inputs include the velocities $(u, v)$ at boundary points and $\pmb{x}$, i.e., a total of $N_{var}\times N_{bc} + N_{dim} = 128\times 2 + 2 = 258$ inputs. Following \cite{Jin2021}, we exclude the BC on $p$ from the inputs and instead solve it as an internal variable. \revise{The GFNet has three outputs $\nn(\pmb{g}, \pmb{x}|\hp) = (\hat{u}, \hat{v}, \hat{p})$ which approximate the solution $(u, v, p)$ at $\pmb{x}$.} The loss function of GFNet is adopted from Equation \ref{eqn:ex_pinn_loss} as follows:
\begin{equation}
    \begin{split}
    L(\hp) &=\frac{1}{N_0}\sum^{N_0}\left(\gamma_1(\hat{u} - u)^2 + \gamma_2(\hat{v} - v)^2 + \gamma_3(\hat{p} - p)^2 \right)\\
    &+ \frac{1}{N_1}\sum^{N_1}\left (\alpha_1H_1^2 + \alpha_2H_2^2 + \alpha_3H_3^2\right ) + \beta |\hp|^2.
    \label{eqn:ns_loss}
    \end{split}
\end{equation}
where the first summation minimizes the error on predictions, the second summation penalizes the residuals, and the last term represents the Tikhonov regularization.
$\pmb{\gamma}=[\gamma_{1}, \gamma_{2}, \gamma_{3}]$, $\pmb{\alpha}=[\alpha_{1}, \alpha_{2}, \alpha_{3}]$, and $\beta$ balance the contributions to the overall loss. 
\revise{When minimizing the PDE residuals, we cannot adapt the collocation points based on one variable's spatial gradients in the same way we do for learning the Laplace equation (see Section 4.1.2). This is because the point distribution based on one variable's gradient may not align with the distributions based on the other two variables' gradients. 
We could use the norm of the gradient vector but that will bias the training to favor learning one variable better in expense of losing accuracy for the other variables. 
To avoid this issue, we increase the number of collocation points by 3 times compared to solving the Laplace equations (1200 over 400, see Tables \ref{tab:laplace_train} and \ref{tab:ns}) and randomly initialize their positions with a uniform spatial distribution. 
These collocation points are fixed during training.}
We use the FC architecture due to the non-linear nature of NS equations and select the layer sizes as $256^3\otimes128^3\otimes96^3\otimes64^3\otimes32^2\otimes3$. All except the last hidden layer use $\tanh$ as the activation function.
\begin{table*}[!b]
    \centering
    \caption{GFNet's accuracy in learning the NS equations: 
    The first and second values in column one indicate the number of data and collocation points, respectively. $\gamma_{1}$, $\gamma_3$, $\alpha$, and $\beta$ are the coefficients in Equation \ref{eqn:ns_loss}. The three validation error and loss values correspond to $u(\pmb{x}), v(\pmb{x}),$ and $p(\pmb{x})$ which denote the velocity components and pressure, respectively. 
    }
    \input{./tab/ns_train.tex}
    \label{tab:ns}
\end{table*}
The depth and size of our GFNet are considerably larger than the networks in related works \cite{Raissi2019, Wang2020a,Jin2021,Cai2021} because we ($1$) use BCs (which are high-dimensional) as inputs, and ($2$) find that a deep network with Tikhonov regularization generalizes better than shallow networks that exclude regularization. To find the optimal values of $\pmb{\gamma}$, $\pmb{\alpha}$, and $\beta$, we set $\gamma_1=\gamma_2$ and $\alpha_1=\alpha_2=\alpha_3=\alpha$ since $u$ and $v$ are equally important and the three equations are highly coupled, i.e., no reason to penalize one equation's residual more than the others. We alter $\beta$ between 1e-13 and 1e-9 and for each value of $\beta$, test a few combinations of $\gamma_1$, $\gamma_3$ and $\alpha$. The most accurate results obtained are presented in Table \ref{tab:ns}.

\begin{figure*}[!b]
    \centering
    \subfloat[Ground truth\label{fig:cavity_openfoam}]{\includegraphics[width = 0.25\textwidth]{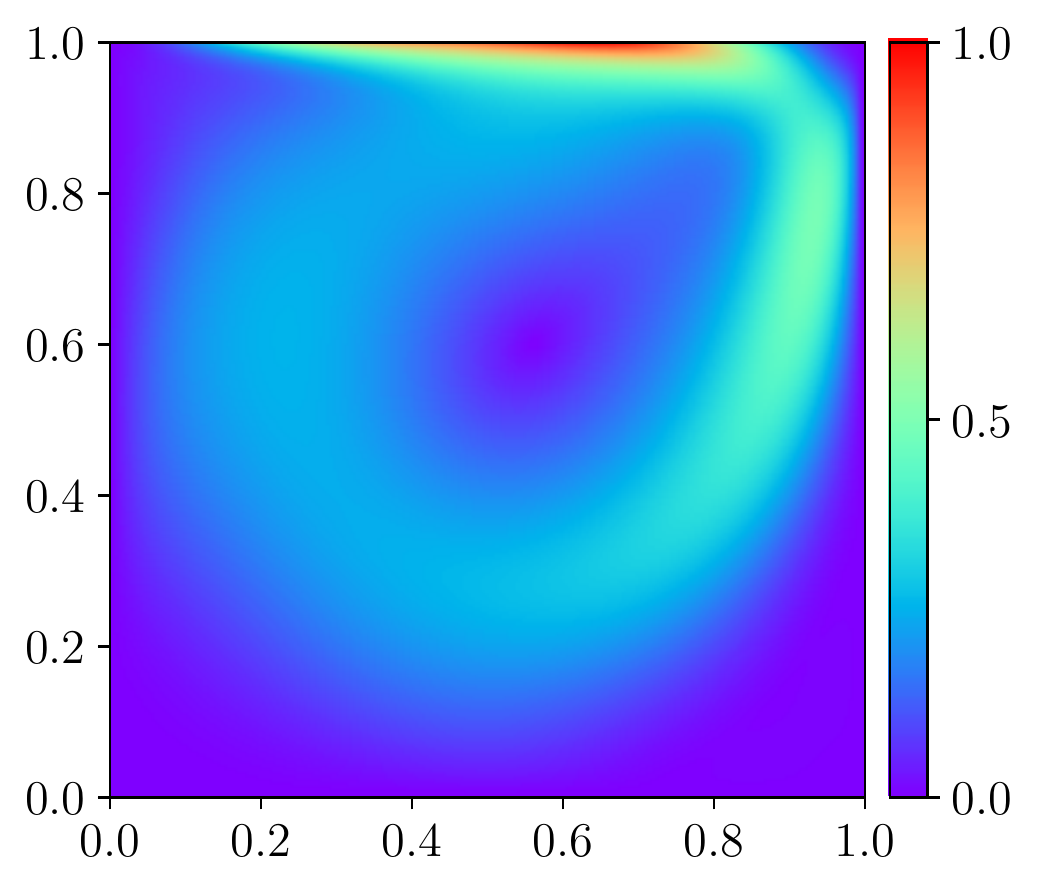}}\hfil
    \subfloat[\mfpShort, 9 genomes\label{fig:cavity_9g}]
    {\includegraphics[width = 0.25\textwidth]{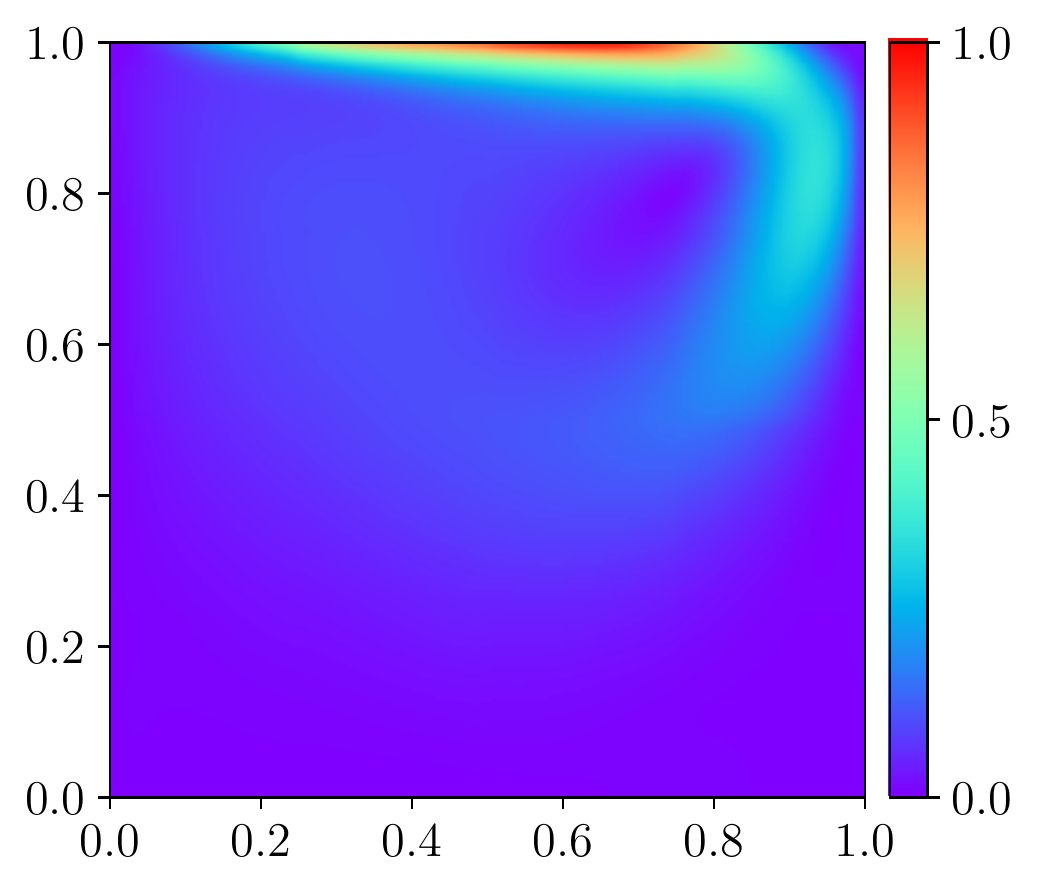}}\hfil
    \subfloat[\mfpShort, 19 genomes\label{fig:cavity_19g}]
    {\includegraphics[width = 0.25\textwidth]{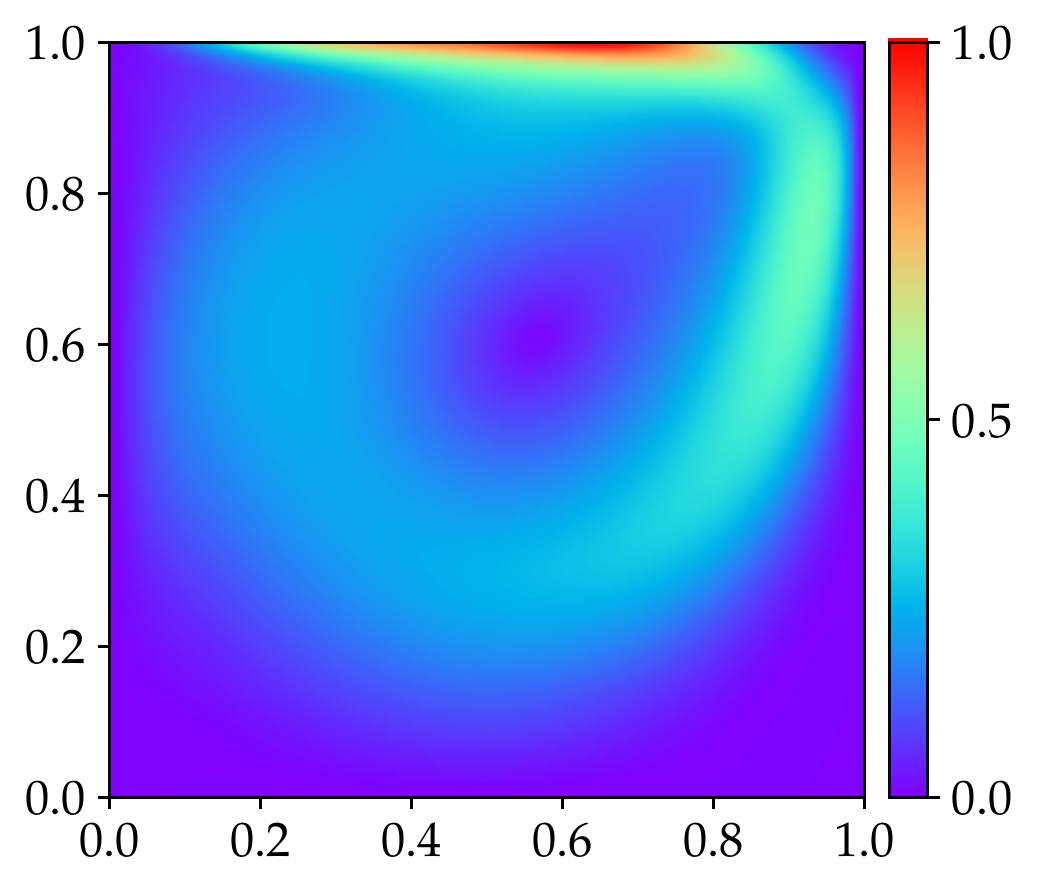}}\hfil
    \subfloat[PINN\label{fig:cavity_pinn}]
    {\includegraphics[width = 0.25\textwidth]{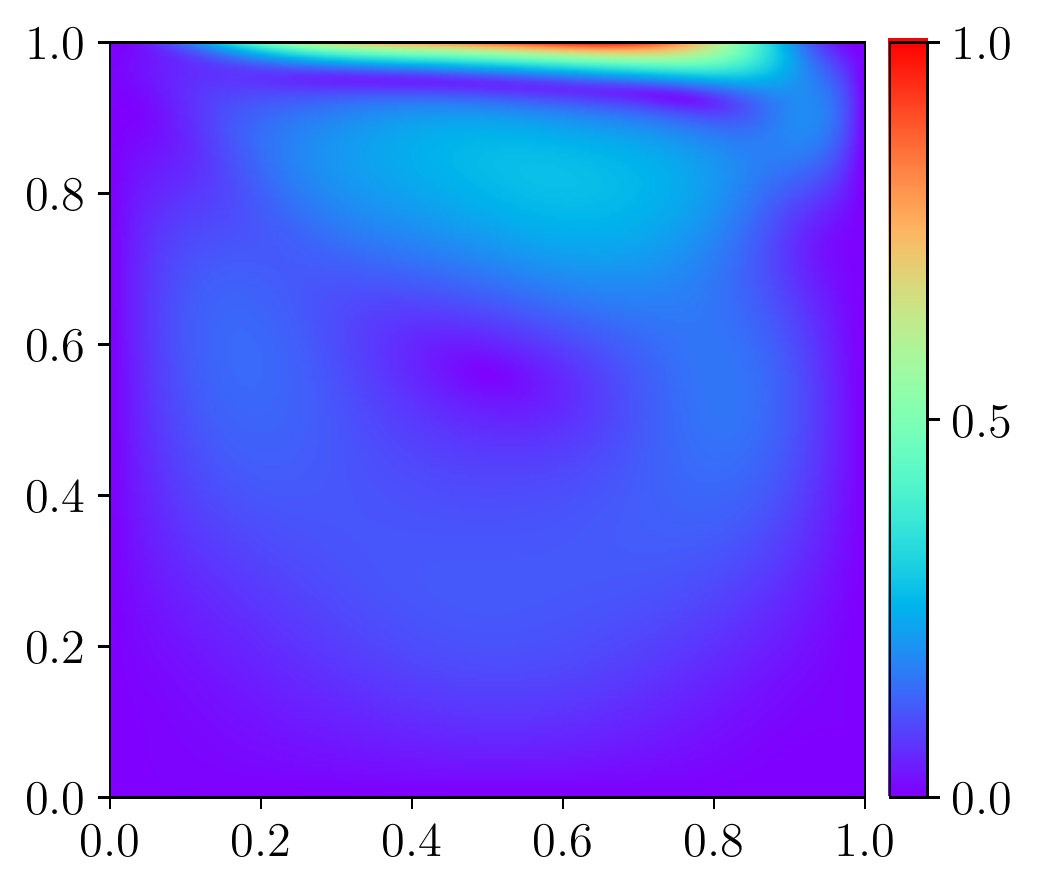}}\\
    \subfloat[The arrangement of 10 additional auxiliary genomes\label{fig:19g}]
    {\input{./fig/19g.tex}}
    \caption{Velocity magnitudes $\sqrt{u^2+v^2}$ for flow in a square lid-driven cavity with unseen BC: 
    \subc{a} The ground truth is simulated by OpenFOAM \cite{jasak2007openfoam}. 
    \subc{b} \mfpShort with 9 genomes fails to accurately predict the velocities. 
    \subc{c} \mfpShort with 19 genomes accurately predicts the unseen conditions and achieves an MAEs of 5.68e-3 and 4.96e-3 for velocities $(u, v)$, respectively.
    \subc{d} PINN fails to resolve the flow and achieves MAEs 1.52e-1 and 1.39e-1 for velocities $(u, v)$.
    \subc{e} In addition to the 9 genomes ($g_{1-9}$) arranged as shown in Figure \ref{fig:step3}, we add 10 more auxiliary genomes ($g_{10-19}$) to improve MF predictor's accuracy in resolving the unseen flow.
    }
    \label{fig:Single_Cavity}
\end{figure*}

For data-only GFNet, i.e., $\alpha=0$, reducing pressure's contribution to gradient descent improves the overall accuracy. 
We use $\gamma_3=0.5$ since further decreasing it to $\gamma_3=0.2$ negligibly improves the predictions for velocities while increasing the error on pressure by $27\%$.
Next, we increase the number of samples and \revise{minimize PDE residuals in GFNet's training.}
Using collocation points and $\alpha=\text{1e-3}$ reduces the accuracy by more than $50\%$ and further increasing $\alpha$ exacerbates the inaccuracies. 
The reasons that \revise{minimizing PDE residuals} reduces accuracy are twofold.
Firstly, as pointed out in \cite{Wang2020a, jagtap2020adaptive, wang2020ntk}, training physics-informed models is equivalent to solving a stiff system of ordinary differential equations which requires more advanced training techniques such as adaptive weight or adaptive activation functions. 
Leveraging these techniques can potentially increase the accuracy of our GFNet and will be pursued in our future works. 
Secondly, Equation \ref{eqn:ns_loss} uses AD to compute the derivatives analytically and regularizes the differential form of the NS equations while OpenFOAM solves the integral form of the NS equation. 
Note that the velocity and pressure from OpenFOAM are accurate but their derivatives differ from the ones obtained by analytical differentiation and dissatisfy the PDE residual in Equation \ref{eqn:ns_loss}.
For instance, OpenFOAM reports an average residual of 9e-10 for $H_1$ using the velocities shown in Figure \ref{fig:cavity_openfoam}. However, the same velocities result in an average residual of 3e-4 if FD is used to calculate the derivatives in Equation \ref{eqn:ns}.
As a result, the data loss and PDE residual contradict each other during gradient descent and decrease the overall accuracy of GFNet.

For the lid-driven cavity problem, we are interested in inferring the velocity accurately across unseen BCs and unseen domains. Therefore, we choose the model with the lowest MAE in velocity which is obtained with $\gamma_3=0.5$ and $\alpha=0$.

\begin{figure*}[!t]
	\centering
	\subfloat[$\sqrt{u^2+v^2}$ ground truth\label{fig:step_vel_openfoam}]
	{\input{./fig/step_vel_openfoam.tex}}\hfil
	\subfloat[$\sqrt{u^2+v^2}$ by \mfpShort\label{fig:step_vel_mfp}]
	{\input{./fig/step_vel_mfp.tex}}\hfil 
	\subfloat[Error in $\sqrt{u^2+v^2}$\label{fig:step_vel_err}]
	{\input{./fig/step_vel_err.tex}}\\ 
	\subfloat[Streamlines ground truth\label{fig:step_streamline_openfoam}] 
	{\input{./fig/step_streamline_openfoam.tex}}\hfil
	\subfloat[Streamlines by \mfpShort\label{fig:step_streamline_mfp}] 
	{\input{./fig/step_streamline_mfp.tex}}\\
	\caption{Velocity magnitude $\sqrt{u^2+v^2}$ and streamlines for the flow in a step-shaped lid-driven cavity: Compared to the ground truth simulated by OpenFOAM \cite{jasak2007openfoam}, \mfpShort achieves MAEs of 1.35e-2 and 1.24e-2 for velocity components $u$ and $v$, respectively. The streamline plots (colored by velocity magnitude) reveal that the step induces a complex flow pattern that consists of three inter-related vortices and \mfpShort successfully captures all these vortex structures while the GFNet has never seen such scenarios during training.}
	\label{fig:MFNet_cavityL}
\end{figure*}
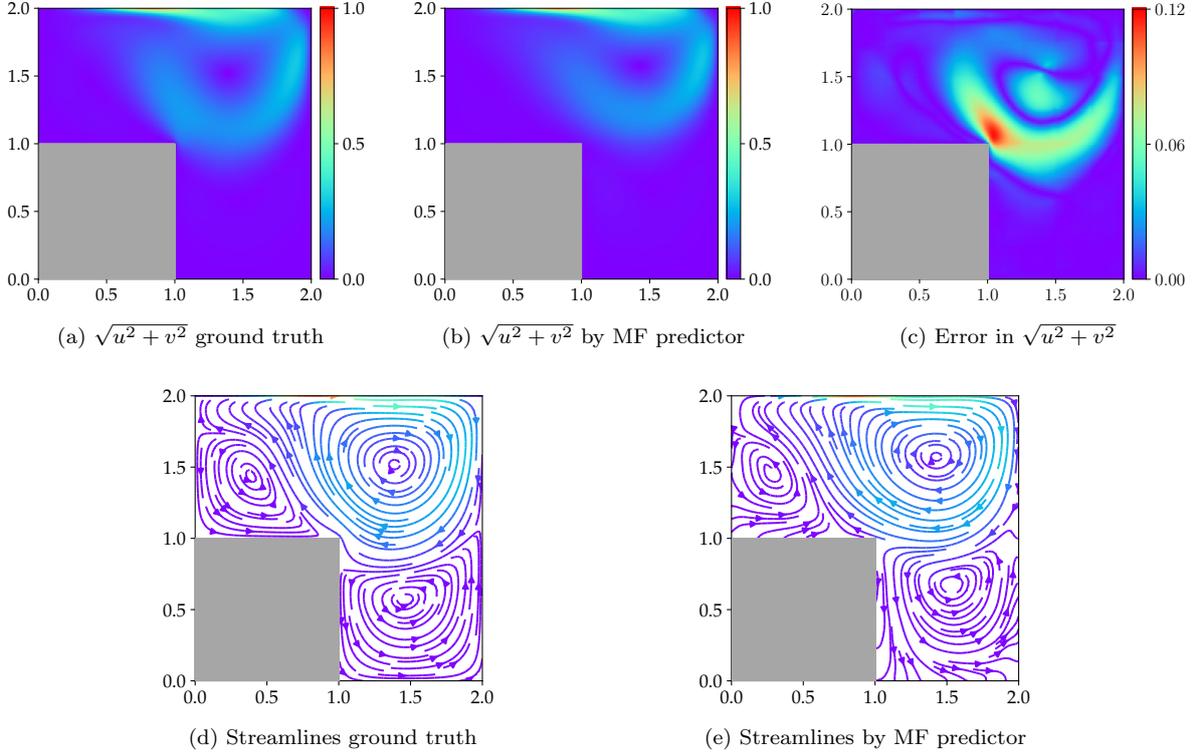
\subsubsection{\textbf{Accuracy for Unseen Domains subject to Unseen BCs and Comparison with PINN.}}~\label{sec:ns_eval}
We evaluate the accuracy of \mfpShort in a domain of size $[0, 1]\times[0, 1]$ which is four times larger than the genome used in training the GFNet. The unseen BC is again generated via a GP and determines the velocity profile of the top lid. We first use \mfpShort with $9$ genomes ($4$ basic and $5$ auxiliary) arranged as shown in Figure \ref{fig:step3} which is also used throughout Section \ref{sec:laplace}. However, as illustrated in Figure \ref{fig:cavity_9g}, using only $9$ genomes does not provide sufficient accuracy and the MAE for velocity components $u$ and $v$ are 7.15e-2 and 6.82e-2, respectively. This is due to the highly non-linear nature of the NS equations which complicates the propagation of the boundary information. To address this issue, we include more \axg (that overlap with the original $9$ genomes) to facilitate the information propagation, see Figure \ref{fig:19g}. As shown in Figure \ref{fig:cavity_19g}, using $19$ genomes significantly improves the accuracy and reduces the MAE of velocity components $u$ and $v$ to 5.68e-3 and 4.96e-3, respectively.

To compare accuracy and performance with the state-of-the-art, we construct the PINN designed in \cite{Wang2020a} to solve the lid-driven cavity problem with $u=1, v=0$ on the top lid (we were able to reproduce the accurate results reported in \cite{Wang2020a}). We re-train the model for $40000$ epochs with Adam using the architecture $50\otimes50\otimes50\otimes50\otimes50$, $4\times129$ boundary points, and $10000$ collocation points. As shown in Figure \ref{fig:cavity_pinn}, PINN is unable to resolve the flow field and the MAE for velocities $(u,v)$ are as large as 1.52e-1 and 1.39e-1. This indicates that the architecture of PINN only applies to the specific BC used in \cite{Wang2020a}, i.e., changing the BC requires not only re-training the PINN but also re-designing its architecture. Regarding computational costs, \mfpShort with $19$ genomes converges to an accurate solution in $117$ seconds while the PINN's training takes $8846$ seconds. 

We further evaluate \mfpShort by predicting the flow in a step-shaped lid-driven cavity that has an unseen velocity profile set to the top lid, see Figure \ref{fig:MFNet_cavityL}. 
This unseen cavity is $12$ times larger than our genomes and is decomposed into $12$ basic and $21$ auxiliary genomes. Figure \ref{fig:MFNet_cavityL} compares the results from \mfpShort and the ground truth simulated by OpenFOAM \cite{jasak2007openfoam} using the same grid resolution as in data generation.
\mfpShort converges in $313$ seconds which is almost $3$ times more than the cost for the square cavity in \ref{fig:cavity_pinn}. This increase in inference time is consistent with the 3:1 area ratio between the two domains.

The MAE for velocities $(u,v)$ are 1.35e-2 and 1.24e-2, respectively, which are larger than the MAE for the square cavity by 1 order of magnitude. 
As shown in Figure \ref{fig:step_vel_err} and \ref{fig:step_streamline_openfoam}, the increased error primarily comes from complex flow patterns that have never been seen by GFNet during its training: the flow over the step corner and the large vortex interleaved with two neighboring vortices on its left and bottom. 
Nonetheless, \mfpShort still captures all three vortex structures in the cavity as shown by the streamlines in Figure \ref{fig:step_streamline_mfp}.
We also observe some unphysical flows in the near-wall regions where the streamlines should be parallel with the wall.
\revise{This error is caused by the inaccuracies of GFNet in predicting the flow on the boundaries, which will be further discussed in Sections \ref{sec:comparison} and \ref{sec:exact_bc}
}

\revise{
\subsection{Detailed Error Analysis and Comparison with DeepONet and Fourier Neural Operator (FNO)}\label{sec:Error_Analysis}
\newcommand{\uopt}{\ensuremath{\pmb{\hat{u}}(\bx)}}
\newcommand{\uex}{\ensuremath{\pmb{u}^*(\bx)}}
\newcommand{\gex}{\ensuremath{\pmb{g}^*}}
\label{sec:comparison}
We attribute the error of our framework to two sources: the \emph{genomic error} which is introduced by GFNet's inference for a single genome with an unseen BC, and the \emph{assembly error} which is accumulated during the iterative procedure in MF predictor.
The genomic error can be further divided into three parts \cite{bottou2007tradeoffs, bottou2010large, jin2020quantifying},
\begin{equation}
\begin{split}
|\nn(\gex, \bx|\hp) - \uex| 
&= |\nn(\gex, \bx|\hp) - \uopt + \uopt - \uex| \\
&\le |\pmb{\delta}(\pmb{\bar{g}}, \bx)| + |\pmb{\delta}(\gex, \bx) - \pmb{\delta}(\pmb{\bar{g}}, \bx)| + |\uopt - \uex|,
\end{split}
\label{eqn:genome_error}
\end{equation}
where $\uex$ is the exact solution to a BVP with an unseen BC ($\pmb{g}^*$), $\uopt$ represents the closest approximation to $\uex$ that a given neural network can achieve, $\pmb{\bar{g}}$ is a training sample that resembles $\pmb{\hat{g}}$, and $\pmb{\delta}(\pmb{g},\bx) = \nn(\pmb{g},\bx|\hp) - \uopt$ measures the difference between $\uopt$ and the approximation achieved by training the neural network.
In Equation \ref{eqn:genome_error}, the first part represents the \emph{optimization} error introduced during training \cite{bottou2007tradeoffs, bottou2010large} while the second part quantifies the neural network's \emph{generalization} error \cite{jin2020quantifying} in predicting unseen inputs.
The last part defines the \emph{approximation} error \cite{bottou2007tradeoffs, bottou2010large} which appears when the target function $\uex$ is not covered by the neural network's functional space.
According to the universal approximation theorem \cite{Hornik1991, Debao1993}, the approximation error effectively diminishes by increasing the size of the neural network.

In practice, it is hard to compute the above errors rigorously since $\uopt$ is typically unknown. 
Nonetheless, we can define approximate metrics to reflect the error's magnitudes, see Figure \ref{fig:error}.
Given the large size of our neural networks which have up to $3\!\! \times\!\!10^6$ parameters, we assume that the approximation error is small compared to the optimization and generalization errors which are estimated by, respectively, GFNet's training MAE and the subtraction of training MAE from test MAE.
We compute the test MAE of GFNet by (1) decomposing an unseen domain into basic genomes, (2) extracting the basic genomes' BCs directly from the ground truth, and (3) inferring the solution in all the basic genomes using the extracted BCs, and (4) evaluating the MAE against the ground truth across all the genomes.
Note that this test MAE is evaluated without using MF predictor and it solely reveals the genomic error in predicting flows with unseen BCs.
At last, we approximate the assembly error by subtracting the GFNet's test MAE from the final MAE achieved by MF predictor.

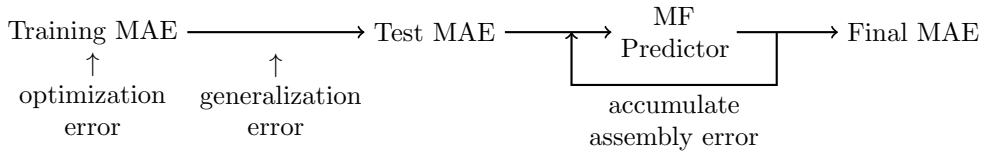
\begin{figure}[!hb]
\centering
\input{./fig/error_analysis}
\caption{\revise{Error decomposition:
         Our framework has two primary error sources. The genomic error is introduced in GFNet's prediction for unseen BCs while the assembly error is accumulated from the iterative procedure in MF predictor.
         The genomic error is divide into optimization error, generalization error, and approximation error \cite{bottou2007tradeoffs, bottou2010large, jin2020quantifying}. 
         The optimization error is estimated by the training MAE.
         The generalization error is measured by subtracting test MAE and training MAE.
         Given the large size of our neural networks, we assume the approximation error is negligible.
         The assembly error is evaluated by subtracting GFNet's test MAE from the final MAE achieved by MF predictor.
         }}
\label{fig:error}
\end{figure}

In what follows we present a detailed error analysis of our framework in solving the Laplace and the NS equations with unseen domains and BCs. 
We note that \textit{the transferability feature of framework is independent of the specific architecture used in GFNet}. That is, other state-of-the-art DNN architectures such as DeepONet \cite{Lu2019} and the Fourier neural operator (FNO) \cite{li2020neural} can also be used to learn the underlying PDE system in the genomic domain and, in turn, used in MF predictor. With this important note in mind, in this section we compare our GFNet's performance against DeepONet and FNO at the genomic as well as large scales to obtain a deeper understanding of the error sources.

\subsubsection{Laplace equation}

We train and compare LPFC against DeepONet and FNO where in both cases we test different architectures and optimization hyper-parameters and select the optimum one. The optimized DeepONet is an unstacked DeepONet with a branch network consisting of $6$ fully connected hidden layers of sizes $100^6$ and a trunk network consisting of $6$ fully connected hidden layers of sizes $100^6$, all using swish activation functions (see \cite{Lu2019} for more details on the architecture). The optimized FNO has an architecture with width $64$ and $18$ modes (see \cite{RN1257} for more details). The training procedure used in these two operator learners includes a total of $10000$ Laplace solutions from which $8000$ are used in training while $2000$ employed in validation. For each sample, we include 900 data points (placed on a regular grid), 124 boundary condition points, and no collocation points. The reason that collocation points are not used in training any of the models compared in this subsection is that FNO cannot be trained using AD because it predicts on a pre-selected grid which is fixed during training. Hence, the output of FNO is not single-valued and the AD-based residual of a single point is not properly defined. Thus, for fair comparison, we retrain the LPFC architecture with the same setup explained above. The accuracy of these three models are also evaluated against the best LPFC architecture of section \ref{sec:laplace_accuracy_unseen} (which uses AD-based residuals in training) to investigate the effect of collocation points on the performance.

The training results are summarized in Table \ref{tab:Operator_Train} and indicate that in this experiment FNO achieve the best MAE results in both training and validation. We note that while our best LPFC architecture presents the worst validation error, it outperforms the other models in terms of the PDE residual. This result is a direct consequence of training it with collocation points but less labeled data points. Additionally, we attribute the better performance of LPFC over DeepONet to two different reasons: $(1)$ LPFC can be considered as a simplified version of DeepONet and therefore is easier to train, and $(2)$ LPFC preserves the linearity of the boundary conditions which is a key feature of the Laplace equation.

\begin{table}[!h]
	\centering
	\vspace{3mm}
	
	\caption{\revise{Accuracy of different operator learners: The residual is computed with both FD (with a step of $1/256$) and AD. AD-based residual cannot be calculated for FNO due to its architecture and training are set up. Only LPFC best uses collocation points in its training.}}
\revise{

\input{./tab/Operator_Train.tex}}
	\label{tab:Operator_Train}
\end{table}

Next, we evaluate the performance of these models when used in \mfpShort. We perform this test on our largest domain that represents the spelling of "Mosaic Flow", see Figure \ref{fig:sclogo}. The results are summarized in Table \ref{tab:MFPredictor_Operator} and indicate interesting features about the error sources. We observe that the training and validation errors shown in Table \ref{tab:Operator_Train} slightly correlate with the generalization error, i.e., the model with lower training/validation error has smaller generalization error compared to other models. However, this trend is not observed in the assembly error, i.e., a lower generalization error does not result in a smaller assembly error. A closer look at Tables \ref{tab:Operator_Train} and \ref{tab:MFPredictor_Operator} indicates that the best assembly error is obtained for the model that achieves smallest residual error, i.e., the LPFC architecture which employs collocation points in its training. In fact, the largest assembly error is obtained by the FNO architecture whose FD-based residual diverges as the FD step size decreases. From these observations we conclude that when \mfpShort is used in large domains the PDE residuals at the genome scale greatly contribute to the final MAE and hence the trained GFNet must achieve not only a small MAE in training/validation, but also a small PDE residual.

\begin{table}[!ht]
	\centering
	\vspace{3mm}
	\caption{\revise{Effect of GFNet on the errors: The errors are defined in Section \ref{sec:Error_Analysis} (see also Figure \ref{fig:error}) and calculated for the "Mosaic Flow" logo in Figure \ref{fig:sclogo}.}}
	\revise{\input{./tab/MFPredictor_Operator.tex}}
	\label{tab:MFPredictor_Operator}
\end{table}

\subsubsection{Navier-Stokes}
We repeat the analyses of previous section for the NS equations. In particular, we build two GFNets using the DeepONet and FNO architectures using the training data introduced in Section \ref{sec:trainig_NS} (2634 samples with 961 data points per sample and no collocation points). We test all GFNets within MF predictor to estimate the flow in the step-shaped cavity. 

Rows one through three in Table \ref{tab:step_cavity_err} enumerate the breakdown of errors where FNO achieves the lowest optimization error while FC slightly outperforms the other two models in terms of the generalization error. Compared to the results in Table \ref{tab:MFPredictor_Operator}, the ratio between generalization and optimization errors has increased because $(1)$ compared to the Laplace equation, the NS equations are much more complex and have higher dimensionality, and $(2)$ the test samples used for calculating the generalization error are extracted from the step-shaped cavity which not only has unseen BCs but also contains very complex unseen flow features, i.e., the flow over the sharp corner and the interleaved vortex structure (see Figure \ref{fig:step_streamline_openfoam}). These features adversely affect the performance of all models in this experiment.

\begin{table}[!h]
    \centering
    \caption{\revise{
    Accuracy of different GFNets for learning the NS equations:
    We compare four different GFNet implementations that include the fully connected (FC) network, DeepONet, fourier neural operator (FNO), and FC with the enforcement of the input BC (FC$+$BC). The generalization and assembly error are evaluated on the step-shaped cavity. Each error includes two entries for the two velocity components which are $u$ and $v$, respectively.
    DeepONet and FNO achieve lower optimization errors but comparable/worse generalization and assembly errors. 
    Enforcing the exact BC not only improves GFNet's training and generalization errors but also reduces the assembly error in MF predictor.
    }}
    \revise{\input{./tab/ns_error}}
    \label{tab:step_cavity_err}
\end{table}

\begin{table}[!h]
    \centering
    \caption{\revise{
    Residual comparison of different models for learning the NS equations:
    We compare the PDE residuals of three GFNets where FC outperforms the other models by 1-2 orders of magnitudes. The calculations are performed on validation data from the training. 
    }}
    \revise{\input{./tab/ns_operator_residual}}
    \label{tab:ns_operator_residual}
\end{table}

In terms of the assembly and final errors, the GFNet based on FC accumulates the least error.
This observation aligns with the results reported in the previous section for the Laplace equation: FC achieves smaller residuals compared to DeepONet and FNO (see Table \ref{tab:ns_operator_residual}) and hence it provides smaller assembly error. 
While this observation motivates the use of collocation points in training a GFNet, for the reasons argued in Section \ref{sec:trainig_NS}, we do not use residuals in approximating the solution of the NS equations. We will research effective ways to incorporate PDE residuals in learning the NS equations in our future work.

\subsubsection{Assembly error and genome density}

In the previous section we showed that the final MAE of the \mfpShort depends on the generalization and assembly errors where the latter error source is directly affected by the PDE residuals of the GFNet. In section \ref{sec:ns_eval}, we demonstrated that the genome density also affects the assembly error where using more auxiliary genomes improved the predicted flow in the single cavity domain, see Figure \ref{fig:Single_Cavity}. Following this observation, we examine the effect of genome density on the assembly error in this section. 

To consistently evaluate assembly error against genome density, we evaluate the effect of placing extra layers of auxiliary genomes that overlap with the borders of the previously added auxiliary layers. Following the descriptions in Section \ref{sec:overlap_genome}, we divide each layer of these auxiliary genomes into three different categories based on the borders of the existing auxiliary genomes: we distinguish between genomes that overlap with $(i)$ the vertical borders (e.g., genomes 5 and 6 in Figure \ref{fig:step3} or genomes 10, 11, 12 and 13 in Figure \ref{fig:19g}), $(ii)$ the horizontal borders (e.g., genomes 7 and 8 in Figure \ref{fig:step3} or genomes 14, 15, 16 and 17 in Figure \ref{fig:19g}), and $(iii)$ the corners where 4 basic genomes reside (e.g., genome 9 in Figure \ref{fig:step3} or genomes 18 and 19 in Figure \ref{fig:19g}). As different combinations of these additional layers leads to different overall genome arrangements (which, in turn affects the iterative update procedure), we add each of these layers in several steps: adding only central overlapping genomes, adding only vertical or only horizontal genomes, adding both vertical and horizontal, and adding all genomes (i.e., vertical, horizontal and central) at once. 

We conduct this experiment on the single cavity domain for the NS equation using an FC architecture as a GFNet. The results are summarized in Figure \ref{fig:Genome_vs_Accuracy} and demonstrate two interesting trends. $(1)$ There is a local minimum in the MAE which appears as a result of balancing propagated errors and boundary information. That is, as we increase the number of genomes the both the BCs and errors propagate to the interior of the domain and after the local minimum the propagated errors dominate the transferred information. $(2)$ The extra horizontal overlapping genomes improve the accuracy of \mfpShort for the single cavity domain more than the other genomes. We attribute this observation to the fact that the flow is predominately affected by the applied lid-velocity at the top of the domain which means that the information is mostly propagated downwards.

\begin{figure}[!hb]
	\centering
	\includegraphics[width=.9\textwidth]{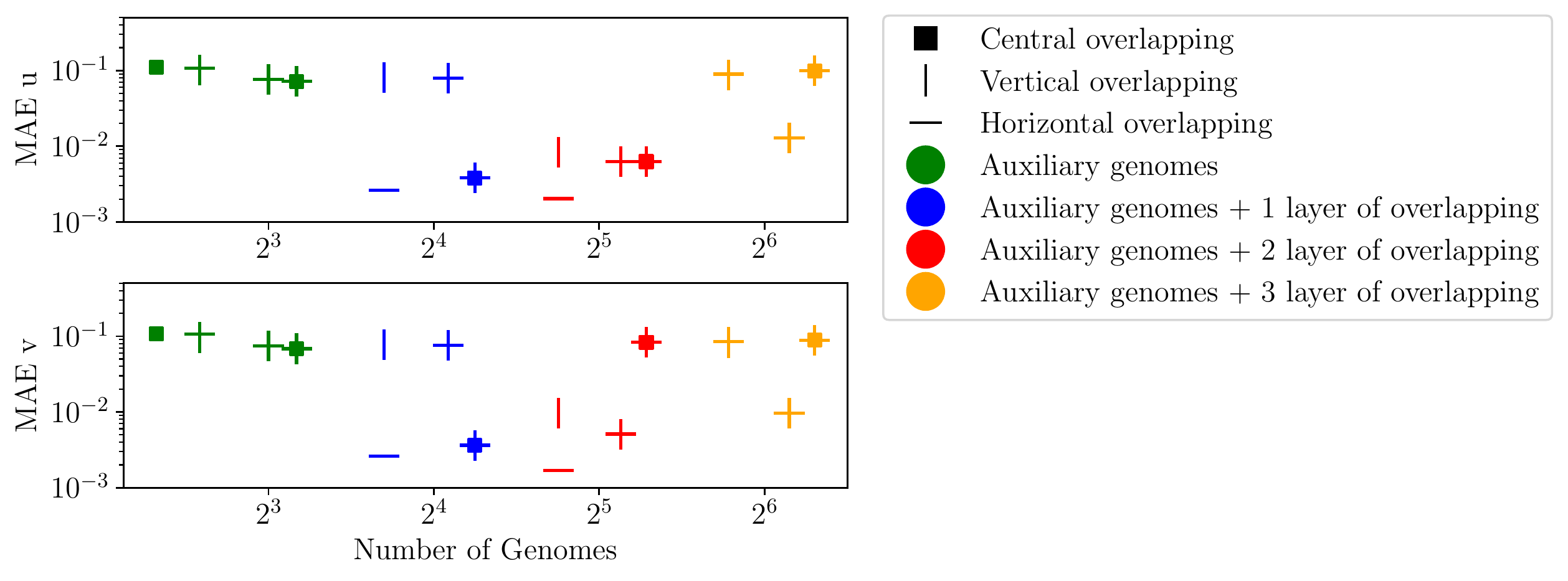}
	\caption{\revise{\mfpShort accuracy vs genome density for the single cavity domain: The accuracy of \mfpShort is evaluated under different arrangements and numbers of auxiliary genomes. Here, green indicates that only a single layer of auxiliary genomes are used. These genomes are placed over the borders of the basic genomes and can be centered on the vertical borders (e.g., genomes 5 and 6 in Figure \ref{fig:step3} ), horizontal borders (e.g., genomes 7 and 8 in Figure \ref{fig:step3}), corner of 4 genomes (e.g., genome 9 in Figure \ref{fig:step3}), or at any combination of these three configurations. The color blue shows that an extra layer of overlapping auxiliary genomes is placed at the vertical, horizontal, or central borders of the existing auxiliary genomes (e.g., Figure \ref{fig:19g}). The colors red and orange indicate, respectively, that two and three extra layers of overlapping auxiliary genomes are used.}}
	\label{fig:Genome_vs_Accuracy}
\end{figure}

Locating the abovementioned optimum has two primary challenges. First, the optimum configuration is problem dependent. For instance, adding extra layers for the step-shape cavity domain would add extra assembly errors and increase final MAEs regardless of the employed arrangement. Second, computational costs increase as more genomes are employed in MF Predictor. In our experiments, we have observed that the simple arrangement described in Section \ref{sec:overlap_genome} consistently outperforms other arrangements (either simpler or more complex) and therefore we recommend it as the default technique. In our future work we will research automatic and adaptive techniques to arrange the basic and auxiliary genomes.

\subsection{Enforcing Exact Boundary Conditions in GFNet}\label{sec:exact_bc}
For the step-shaped cavity, GFNet yields inaccurate predictions near the walls where the streamline should be parallel with the boundaries, see Figure \ref{fig:step_streamline_openfoam} and \ref{fig:step_streamline_mfp}.
This is because GFNet applies the BCs by penalizing the MSE at the boundary points (see Figure \ref{fig:laplace_points}), i.e., the input BCs are not strictly enforced. The BC for stationary walls is $u=v=0$ where even very small errors greatly affect the velocity direction and cause unphysical streamlines. To reduce this error, which will improve the accuracy of both GFNet and MF predictor, we design the architecture of GFNet to strictly reproduce the input BCs.

To design a GFNet that reproduces a BC exactly, we follow \cite{berg2018unified} and decompose the approximated solution into three parts:
\begin{equation}
    u(\bx) = G(\bx) + \phi(\bx)\cdot\nn(\bx|\hp),
    \label{eqn:exact_bc}
\end{equation}
where $G(\bx)$ is a smooth function defined on $\Omega$ that satisfies the BC, i.e., $\bG = \bg, \bx \in \partial\Omega$, $\phi(\bx)$ is a smooth function that equals to zero on $\partial\Omega$, and $\nn(\bx|\hp)$ denotes the neural network which can be implemented with any of the models explored in Section \ref{sec:comparison}.
$G(\bx)$ can be constructed by either extrapolating $\bg$ from $\partial\Omega$ to $\Omega$ \cite{sukumar2021exact} or training a neural network that interpolates the boundary points \cite{berg2018unified}.
To avoid masking the network, $\phi(\bx)$ is recommended to be nonzero inside $\Omega$.
The typical choice of $\phi(\bx)$ is the signed distance function that measures the distance from $\bx\in\Omega$ to $\partial\Omega$ \cite{berg2018unified, sukumar2021exact, sun2020surrogate}.

Note that Equation \ref{eqn:exact_bc} only solves the PDE with a specific BC on one domain.
We make the following adjustments to enforce the input BC in GFNet and retain its transferability across unseen BCs.
First, our model takes both $\pmb{g}$ and $\bx$ as inputs. 
We follow \cite{sukumar2021exact} to extrapolate $\pmb{g}$ by weighted averaging,
\begin{equation}
G(\pmb{g}, \bx) = \sum_{i=1}^{N_{bc}}
      \frac{|\bx - \bx^{bc}_i|^{-2} \pmb{g}_i}
           {\sum_{j=1}^{N_{bc}}|\bx - \bx_j^{bc}|^{-2}},
\end{equation}
where $|\bx - \bx_i^{bc}|$ represents the distance between $\bx\in\Omega$ and the $i^{th}$ boundary point and $\pmb{g}_i = g(\bx_i^{bc})$. 
In practice, we add a small constant $\epsilon=10^{-10}$ to the distance to avoid division by zero.
Second, given that GFNet only infers flows in square genomes, we simply choose $\phi(\bx) = x(l-x)y(l-y)$ where $l$ is the square genome's edge length.
Finally, we apply the neural network defined in Section \ref{sec:nn_bvp} , $\nn(\pmb{g},\bx|\hp)$, to Equation \ref{eqn:exact_bc} to achieve transferability across unseen BCs. 

We now approximate the solution of the NS equations with a FC-based GFNet that enforces the input BC as described above. We denote this GFNet as FC$+$BC and use the training data and the hyperparameters' tuning strategies explained in Section \ref{sec:ns}.
In our tests, the architecture $400^6\otimes3$ delivers the best accuracy, improving GFNet's training MAE of $u$ and $v$ by 62\% and 84\%, respectively, see Table \ref{tab:step_cavity_err}.
It also shows that FC+BC not only decreases the optimization and generalization errors but also reduces the assembly error by an order of magnitude.
To explain this improvement, we revisit the example in Figure \ref{fig:iter_2genome} where the two genomes $\Omega_1$ and $\Omega_2$ share the BC at border $\partial\Omega_1\cap\partial\Omega_2$. 
For FC-based GFNet, its predictions ($u_1$ and $u_2$ in Figure \ref{fig:iter_2genome}) based on BCs from $\partial\Omega_1$ and $\partial\Omega_2$ can differ at the shared border.
This discrepancy results in a discontinuous solution per iteration and contributes to the assembly error accumulated in MF predictor. 
Enforcing the input BC in GFNet eliminates the discrepancy at the shared border and therefore drastically reduces the assembly error.
Similar to the FC-based GFNet, the assembly error of \mfpShort with FC$+$BC mainly comes from the unseen flow features.
Nonetheless, imposing the exact BC drastically reduces the unphysical errors in streamlines, which become parallel with the boundaries in most of the near-wall regions, see Figure \ref{fig:step_streamline_mfpbc}.
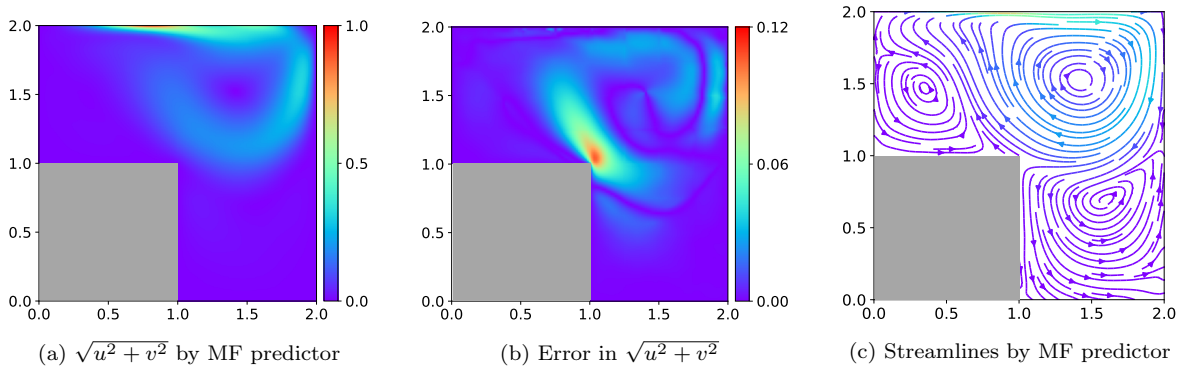
\begin{figure}[!t]
	\centering
	\subfloat[$\sqrt{u^2+v^2}$ by \mfpShort\label{fig:step_vel_mfpbc}]
	{\input{./fig/step_vel_mfpbc.tex}}\hfil 
	\subfloat[Error in $\sqrt{u^2+v^2}$\label{fig:step_vel_err_mfpbc}]
	{\input{./fig/step_vel_err_mfpbc.tex}}\hfil
	\subfloat[Streamlines by \mfpShort\label{fig:step_streamline_mfpbc}] 
	{\input{./fig/step_streamline_mfpbc.tex}}\\
	\caption{\revise{Velocity magnitude $\sqrt{u^2+v^2}$ and streamlines for the step-shaped lid-driven cavity using the GFNet with exact BC enforced. 
	Compared to the GFNet using FC, enforcing the exact BC improves MAEs to 8.07e-3 and 1.00e-2 for velocity components $u$ and $v$, respectively. 
	The error plot shows that the errors still primarily from the sharp corner and the interleaved vortex structures. 
	The streamline plot reveals that enforcing the exact BC significantly diminishes the unphysical errors in the near-wall regions, as the streamlines are largely parallel with the boundaries.}}
	\label{fig:step_cavity_mfpbc}
\end{figure}

For elliptic PDEs with non-Dirichlet BCs, We note that FC$+$BC can be easily extended to enforce Neumann/Robin BCs and when used in MF predictor the assembled inference converges to the PDE's exact solution.
We refer the readers to \ref{sec:robin_bc} and \cite{sukumar2021exact} for more details.
}

%% file: tab/laplace_train.tex
\begin{tabular}{|l|c|>{\centering\arraybackslash}p{1.2cm}>{\centering\arraybackslash}p{1.2cm}>{\centering\arraybackslash}p{1.2cm} >{\centering\arraybackslash}p{1.2cm}|>{\centering\arraybackslash}p{1.2cm}>{\centering\arraybackslash}p{1.2cm}|}
	\hline
	\multirow{2}{*}{Network} & \multirow{2}{*}{Precision} & \multicolumn{4}{c|}{Training}  &
	\multicolumn{2}{c|}{Test on unseen BCs} \\
	&             & Layers  & Samples & Points                 &
	MAE  & MAE       & MAR     \\ \hline
	FC   & Single & 8       & 2000    & 100\textbackslash{}400 & 1.33e-3  & 2.64e-3   & 1.72e-1 \\
	LPFC & Single & 8       & 2000    & 100\textbackslash{}400 & 3.88e-4  & 5.42e-4   & 2.33e-2 \\
	FC   & Single & 14      & 4000    & 400\textbackslash{}400 & 5.57e-4  & 1.30e-3   & 7.12e-2 \\
	LPFC & Single & 14      & 4000    & 400\textbackslash{}400 & 2.55e-4  & 4.33e-4   & 2.05e-2 \\
	FC   & Single & 14      & 8000    & 400\textbackslash{}400 & 4.06e-4  & 8.79e-3   & 4.17e-2 \\
	LPFC & Single & 14      & 8000    & 400\textbackslash{}400 & 2.37e-4  & 4.10e-4   & 1.30e-2 \\
	FC   & Double & 14      & 8000    & 400\textbackslash{}400 & 3.86e-4  & 8.12e-4   & 3.84e-2 \\
	LPFC & Double & 14      & 8000    & 400\textbackslash{}400 & 2.39e-4  & 4.07e-4   & 1.51e-2 \\
	LPFC & Single & 14      & 18000   & 400\textbackslash{}400 & 1.59e-4  & 4.10e-4   & 1.46e-2 \\
	LPFC & Double & 14      & 18000   & 400\textbackslash{}400 & 1.59e-4  & 4.10e-4   & 1.46e-2 \\\hline
\end{tabular} 

%% file: fig/Laplace_MFNet_vs_PINN.tex
\begin{tikzpicture}
\def\a{4cm} 
\node[anchor=south west,inner sep=0] at (-8cm,\a) {\includegraphics[width=0.455\textwidth]{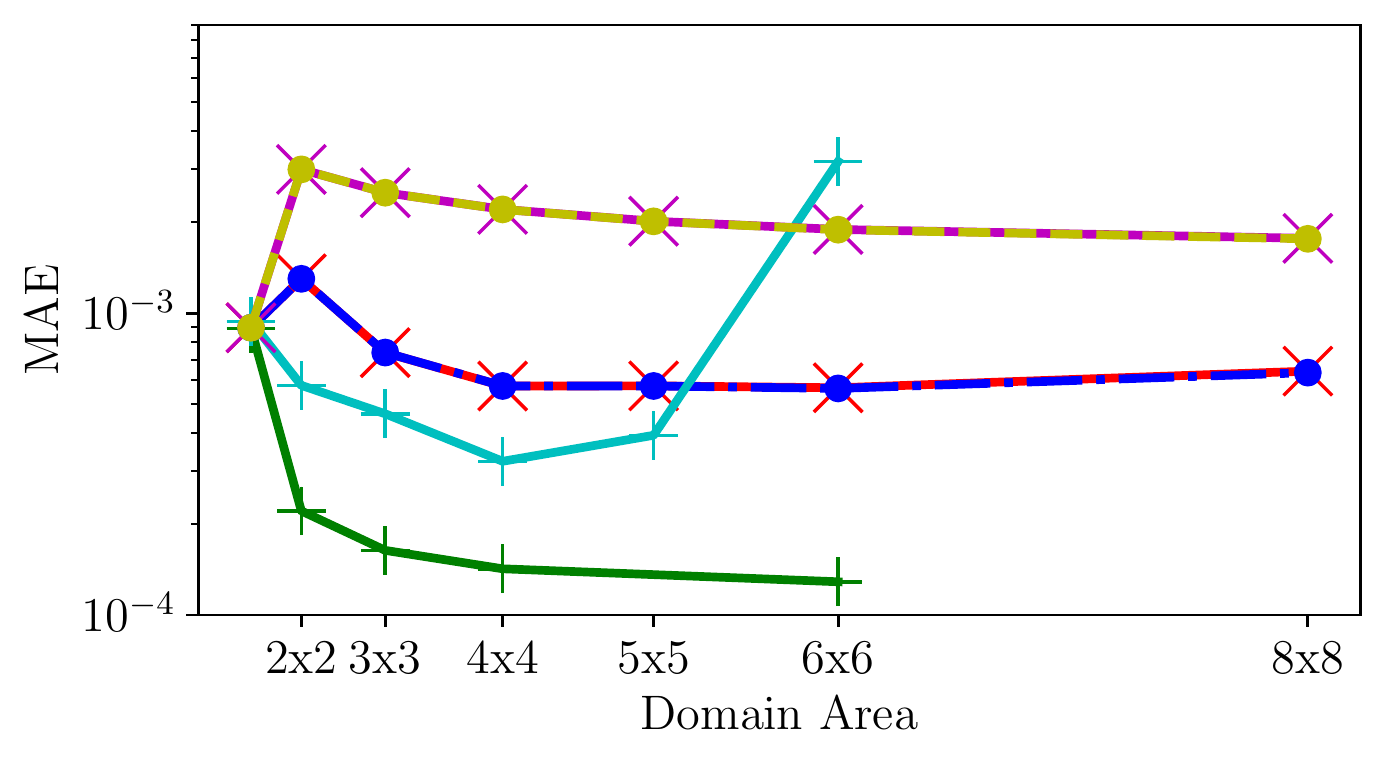}};
\node[anchor=south west,inner sep=0] at (0,\a) {\includegraphics[width=0.45\textwidth, height=0.25\textwidth]{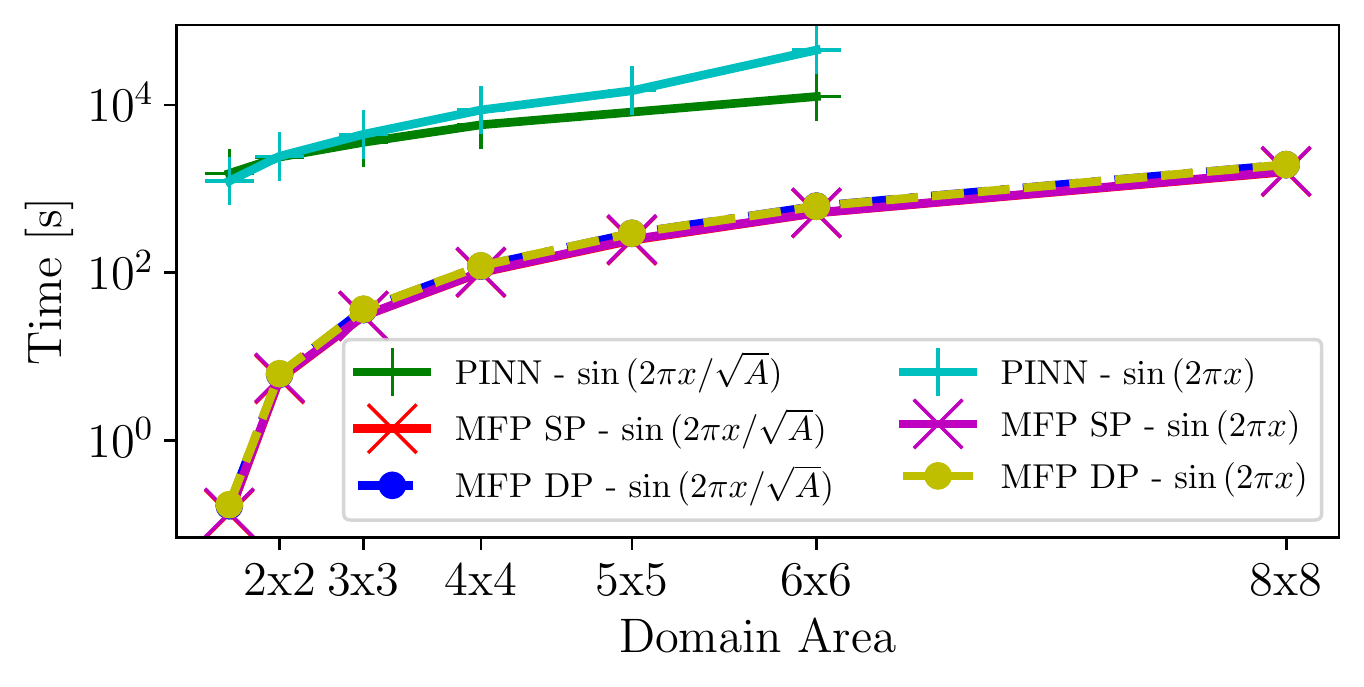}};

\def\h{4.2} 
\def\w{-7.7} 
\begin{scope}
\node[draw, red] at (1.7 + \w,4.9 + \h) {2x2};
\draw[red,thick,dashed, fill=white,] (1.7+ \w,4.1+ \h) circle (0.5cm);
\clip[](1.7+ \w ,4.1+ \h) circle (0.5cm);
\draw[step=0.12cm,gray,very thin] (0.97+ \w,3.5+ \h) grid (2.5+ \w,4.55+ \h);
\draw[red,thick,dashed, ] (1.7+ \w,4.1+ \h) circle (0.5cm);
\fill[blue!40!white] (1.54+ \w,3.95+ \h) rectangle (1.81+ \w,4.21+ \h);
\end{scope}

\begin{scope}
\node[draw, red] at (2.94+ \w,4.9+ \h) {3x3};
\draw[red,thick,dashed, fill=white,] (2.94+ \w,4.1+ \h) circle (0.5cm);
\clip[](2.94+ \w,4.1+ \h) circle (0.5cm);
\draw[step=0.12cm,gray,very thin] (1.97+ \w,3.5+ \h) grid (4.5+ \w,4.55+ \h);
\draw[red,thick,dashed, ] (2.94+ \w,4.1+ \h) circle (0.5cm);
\fill[blue!40!white] (2.75+ \w,3.95+ \h) rectangle (3.12+ \w,4.33+ \h);
\end{scope}

\begin{scope}
\node[draw, red] at (4.33+ \w,4.9+ \h) {4x4};
\draw[red,thick,dashed, fill=white,] (4.33+ \w,4.1+ \h) circle (0.5cm);
\clip[](4.33+ \w,4.1+ \h) circle (0.5cm);
\draw[step=0.12cm,gray,very thin] (3+ \w,3.5+ \h) grid (5+ \w,4.55+ \h);
\draw[red,thick,dashed, ] (4.33+ \w,4.1+ \h) circle (0.5cm);
\fill[blue!40!white] (4.07+ \w,3.82+ \h) rectangle (4.56+ \w,4.32+ \h);
\end{scope}

\draw[thick,->, red] (1.6+ \w,3.6+ \h) -- (1.38+ \w,3.28+ \h);
\draw[thick,->, red] (2.9+ \w,3.58+ \h) -- (1.93+ \w,3.12+ \h);
\draw[thick,->, red] (4.3+ \w,3.58+ \h) -- (2.6+ \w,3.05+ \h);

\end{tikzpicture}

%% file: tab/XPINN_res.tex
\begin{tabular}{|c|c|}
    \hline
    $\hspace{1 cm}$\textbf{Method}$\hspace{1 cm}$& $\hspace{1cm}$\textbf{MAE}$\hspace{1cm}$ \\ \hline
    MF predictor  & 5.82e-4\\\hline
    XPINN    & 2.12e-2\\\hline
\end{tabular} 

%% file: tab/ns_train.tex
\begin{tabular}{|c|cc|c|c|ccc|ccc|}
\hline
Points&\multicolumn{2}{|c|}{$\gamma_1, \gamma_3$} & $\alpha$ & $\beta$ & \multicolumn{3}{c|}{Validation MAE} & \multicolumn{3}{c|}{Validation PDE loss} \\ \hline
961/-   & 1.0  & 1.0  & 0    &  1e-11 &  1.02e-03 & 1.01e-03 & 6.49e-04   &   -      &   -      &    -     \\
961/-   & 1.0  & 0.5  & 0    &  1e-10 &  6.81e-04 & 6.43e-04 & 5.51e-04   &   -      &   -      &    -     \\
961/-   & 1.0  & 0.2  & 0    &  1e-10 &  6.42e-04 & 6.30e-04 & 7.05e-04   &   -      &   -      &    -     \\ 
576/1200 & 1.0  & 0.5  & 1e-3 &  0     &  1.48e-03 & 1.24e-03 & 8.79e-04   & 8.30e-04 & 2.58e-03 & 4.51e-04 \\
576/1200 & 1.0  & 0.5  & 1e-2 &  1e-10 &  1.88e-03 & 1.56e-03 & 9.84e-04   & 3.20e-04 & 4.57e-04 & 1.94e-04 \\ \hline
\end{tabular}

%% file: fig/19g.tex
\begin{tikzpicture}
\def\a{3}; 
\def\b{1.5}; 

\draw[thick] (0, 0) rectangle +(\a, \a);
\draw[very thick, dashed, red, fill=red, fill opacity=0.1]
(0.125*\a, 0) rectangle +(0.5*\a, 0.5*\a);
\node[right] at (0.125*\a, 0.25*\a) {$g_{10}$};
\draw[very thick, dashed, red, fill=red, fill opacity=0.1]
(0.375*\a, 0) rectangle +(0.5*\a, 0.5*\a);
\node[left] at (0.875*\a, 0.25*\a) {$g_{11}$};
\draw[very thick, dashed, red, fill=red, fill opacity=0.1]
(0.125*\a, 0.5*\a) rectangle +(0.5*\a, 0.5*\a);
\node[right] at (0.125*\a, 0.75*\a) {$g_{12}$};
\draw[very thick, dashed, red, fill=red, fill opacity=0.1]
(0.375*\a, 0.5*\a) rectangle +(0.5*\a, 0.5*\a);
\node[left] at (0.875*\a, 0.75*\a) {$g_{13}$};

\draw[thick] (\a+\b, 0) rectangle +(\a, \a);
\draw[very thick, dashed, red, fill=red, fill opacity=0.1]
(\a+\b, 0.125*\a) rectangle +(0.5*\a, 0.5*\a);
\node[above] at (1.25*\a+\b, 0.125*\a) {$g_{14}$};
\draw[very thick, dashed, red, fill=red, fill opacity=0.1]
(1.5*\a+\b, 0.125*\a) rectangle +(0.5*\a, 0.5*\a);
\node[above] at (1.75*\a+\b, 0.125*\a) {$g_{15}$};
\draw[very thick, dashed, red, fill=red, fill opacity=0.1]
(\a+\b, 0.375*\a) rectangle +(0.5*\a, 0.5*\a);
\node[below] at (1.25*\a+\b, 0.875*\a) {$g_{16}$};
\draw[very thick, dashed, red, fill=red, fill opacity=0.1]
(1.5*\a+\b, 0.375*\a) rectangle +(0.5*\a, 0.5*\a);
\node[below] at (1.75*\a+\b, 0.875*\a) {$g_{17}$};

\draw[thick] (2*\a+2*\b, 0) rectangle +(\a, \a);
\draw[very thick, dashed, red, fill=red, fill opacity=0.1]
(2.125*\a+2*\b, 0.125*\a) rectangle +(0.5*\a, 0.5*\a);
\node[above right] at (2.125*\a+2*\b, 0.125*\a) {$g_{18}$};
\draw[very thick, dashed, red, fill=red, fill opacity=0.1]
(2.375*\a+2*\b, 0.375*\a) rectangle +(0.5*\a, 0.5*\a);
\node[below left] at (2.875*\a+2*\b, 0.875*\a) {$g_{19}$};

\end{tikzpicture}

%% file: fig/step_vel_openfoam.tex
\begin{tikzpicture}
\node[anchor=south west,inner sep=0] at (0.0,0.0) {\includegraphics[width=0.3\textwidth]{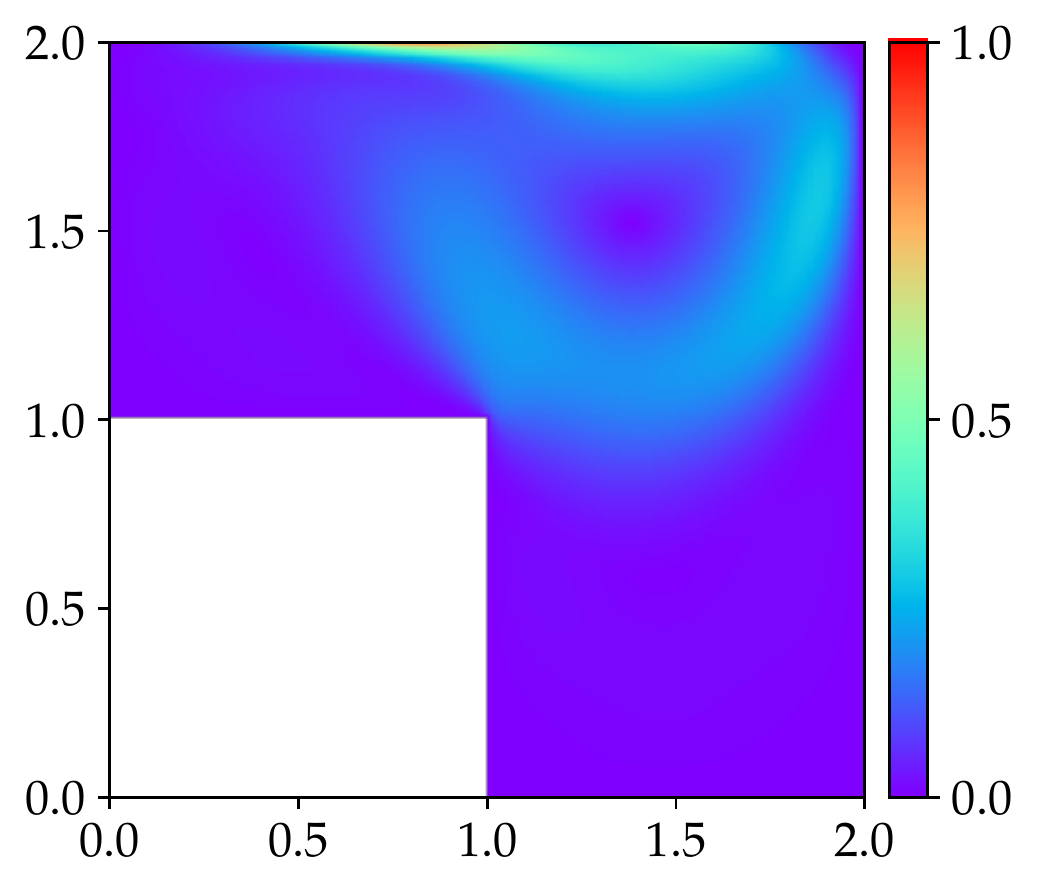}};
\path[fill=gray!70!] 
(0.54,0.46) rectangle +(1.79, 1.79);
\end{tikzpicture}

%% file: fig/step_vel_mfp.tex
\begin{tikzpicture}
\node[anchor=south west,inner sep=0] at (0.0,0.0) {\includegraphics[width=0.3\textwidth]{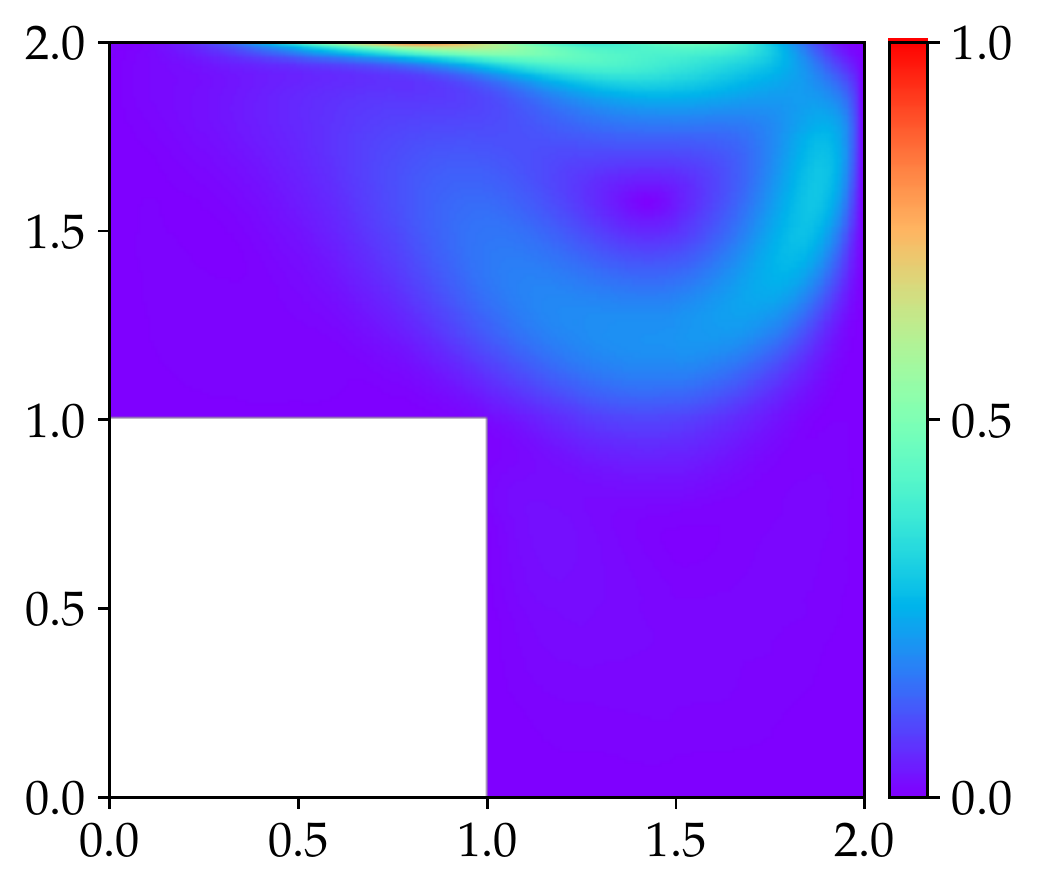}};
\path[fill=gray!70!] 
(0.54,0.46) rectangle +(1.79, 1.79);
\end{tikzpicture}

%% file: fig/step_vel_err.tex
\begin{tikzpicture}
\node[anchor=south west,inner sep=0] at (0.0,0.0) {\includegraphics[width=0.306\textwidth]{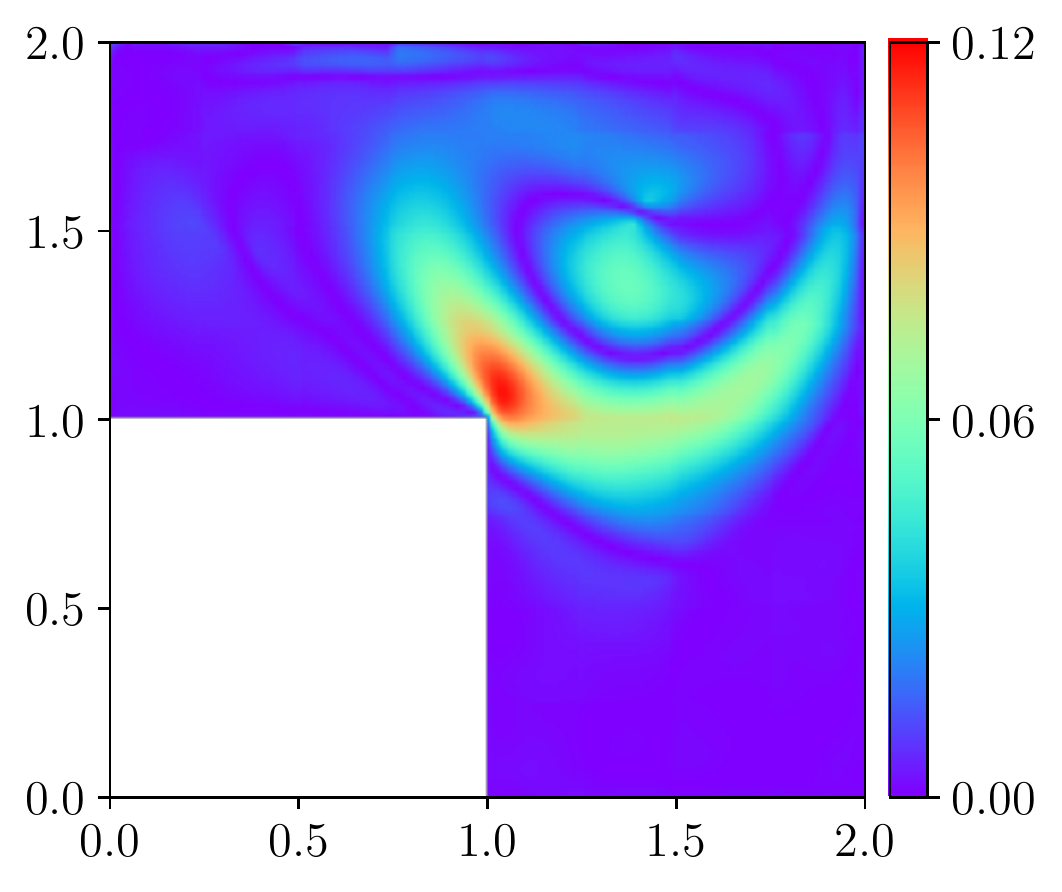}};
\path[fill=gray!70!] 
(0.54,0.46) rectangle +(1.79, 1.79);
\end{tikzpicture}

%% file: fig/step_streamline_openfoam.tex
\begin{tikzpicture}
\node[anchor=south west,inner sep=0] at (0.0,0.0) {\includegraphics[width=0.28\textwidth]{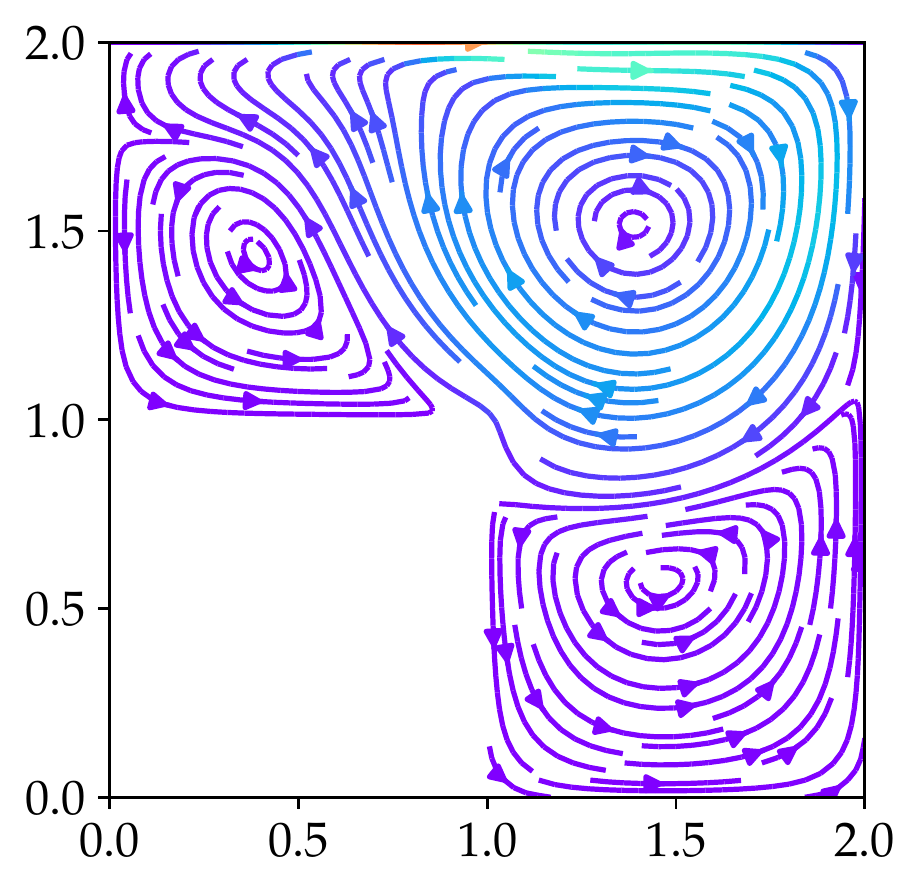}};
\path[fill=gray!70!] 
(0.56,0.48) rectangle +(1.9, 1.9);
\end{tikzpicture}

%% file: fig/step_streamline_mfp.tex
\begin{tikzpicture}
\node[anchor=south west,inner sep=0] at (0.0,0.0) {\includegraphics[width=0.28\textwidth]{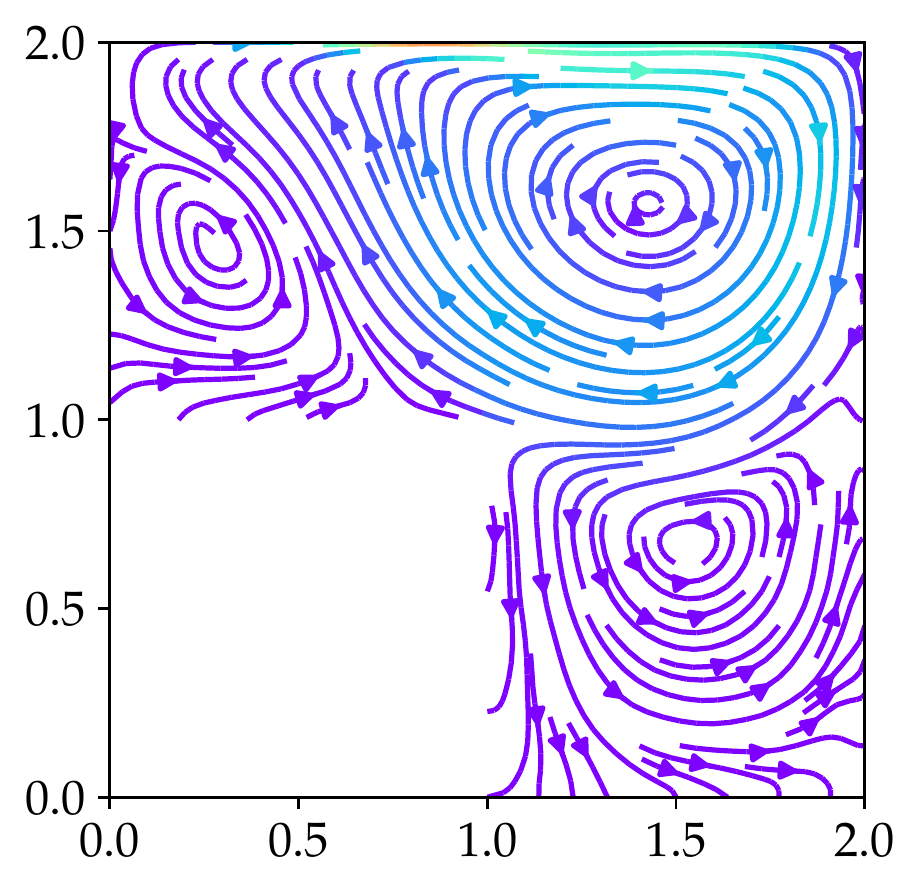}};
\path[fill=gray!70!] 
(0.56,0.48) rectangle +(1.9, 1.9);
\end{tikzpicture}

%% file: fig/error_analysis.tex
\begin{tikzpicture}
\def\a{4.5}; 
\node (train-mae) at (0, 0) {Training MAE};
\node (test-mae)  at (\a, 0) {Test MAE};
\draw[thick, ->] (train-mae) 
                 -- node[below, yshift=-1ex] 
                 {\parbox[t]{2cm}{\centering $\uparrow$ \\ 
                                  generalization \\ 
                                  error}} 
                 (test-mae);
\node[below, yshift=-1ex] at (0, 0) {\parbox[t]{2cm}{\centering $\uparrow$ \\ 
                                                     optimization \\ 
                                                     error}};
                                                     
\def\c{0.7};
\node (mfp-mae) at (2.4*\a, 0) {Final MAE};
\node (mfp) at (1.7*\a, 0) {\parbox[c]{1.4cm}{\centering MF\\Predictor}};
\draw[thick,->] (test-mae) -- (mfp);
\draw[thick,->] (mfp) -- (mfp-mae);
\draw[thick, ->] (2.0*\a,0) -- (2.0*\a, -\c) 
                 --node[below] {\parbox[t]{2.4cm}{\centering accumulate\\assembly error}}
                 (1.4*\a, -\c) -- (1.4*\a, 0);
\end{tikzpicture}

%% file: tab/Operator_Train.tex
  \centering
\begin{tabular}{|l|c|c|c|c|}

\hline
{Operator learner} & {Train MAE} & 
{Validation MAE} & {Residual with FD}  &{Residual with AD }  \\  \hline
	DeepONet & 2.19e-04 & 2.34e-04 & 5.61e+2 & 2.81e+1 \\ 	\hline 
	FNO & 6.95e-05 & 7.58e-05 & 8.48e+2 & -  \\\hline
	LPFC & 1.67e-04 & 1.66e-04 & 1.85e+4 & 5.27e+3 \\ \hline
	LPFC Best & 1.59e-4 & 4.10e-4 & 4.03e-3 & 1.46e-2 \\ \hline
	\end{tabular} 




%% file: tab/MFPredictor_Operator.tex
    
\begin{tabular}{|l|c|c|c|c|}
\hline
 { GFNet Model} & {Optimization Error} &
 {Generalization Error} & {Assembly Error} & {Final MAE} \\ \hline
 DeepONet & 2.19e-04 & 6.87e-04 & 1.97e-03 & 2.87e-03 \\ \hline
 FNO & 6.95e-05 & 5.81e-04 & 4.74e-03 & 5.39e-03   \\  \hline
 LPFC & 1.67e-04 & 4.80e-04 & 1.98e-03 & 2.79e-03 \\ \hline
 LPFC Best & 1.59e-4 & 8.04e-04 & 1.80e-03 & 2.61e-03
 \\ \hline 
\end{tabular}

  
    


%% file: tab/ns_error.tex
\begin{tabular}{|l|cc|cc|cc|cc|cc|}
\hline
\multirow{2}{*}{GFNet Model} & \multicolumn{2}{c|}{Optimization Error} &
 \multicolumn{2}{c|}{Generalization Error} & \multicolumn{2}{c|}{Assembly Error} & \multicolumn{2}{c|}{Final MAE} \\ 
 & $u$ & $v$ & $u$ & $v$ & $u$ & $v$ & $u$ & $v$\\
 \hline
DeepONet   & 6.26e-4 & 5.20e-4 & 1.86e-3 & 2.00e-3 & 6.30e-2 & 6.98e-2 & 6.48e-2 & 7.19e-2 \\
FNO      & 4.01e-4 & 3.87e-4      &       2.39e-3  &  1.76e-3      & 5.02e-2        &   2.74e-2      &  5.30e-2       & 3.31e-2        \\
FC         & 6.74e-4 & 6.54e-4 & 1.87e-3 & 1.65e-3 & 1.10e-2 & 1.02e-2 & 1.35e-2 & 1.24e-2 \\
FC$+$BC    & 4.16e-4 & 3.55e-4 & 1.15e-3 & 1.20e-3 & 6.51e-3 & 8.44e-3 & 8.07e-3 & 1.00e-2 \\ \hline
\end{tabular}

%% file: tab/ns_operator_residual.tex
\begin{tabular}{|l|ccc|ccc|}
\hline
\multirow{2}{*}{GFNet Model} & \multicolumn{3}{c|}{Residual with FD} & \multicolumn{3}{c|}{Residual with AD} \\
 &  $H_1$  & $H_2$  & $H_3$  & $H_1$  & $H_2$  & $H_3$ 
\\ \hline
DeepONet &  4.19e-02 & 6.65e+00 & 9.24e-02 & 1.79e-02 & 9.79e-02 & 2.29e-03  \\ \hline
FNO & 3.34e-01 & 5.73e-01 & 3.72e-01  & - & - & - \\ \hline
FC & 2.30e-03 & 5.96e-03 & 4.27e-04 & 7.62e-04 & 2.96e-04 & 1.37e-04 \\\hline
FC+BC & 4.972e-01 & 1.179e-01 & 9.335e-03 &
1.049e-01 & 
1.960e-03 & 
2.820e-03 \\\hline

\end{tabular}

%% file: fig/step_vel_mfpbc.tex
\begin{tikzpicture}
\node[anchor=south west,inner sep=0] at (0.0,0.0) {\includegraphics[width=0.3\textwidth]{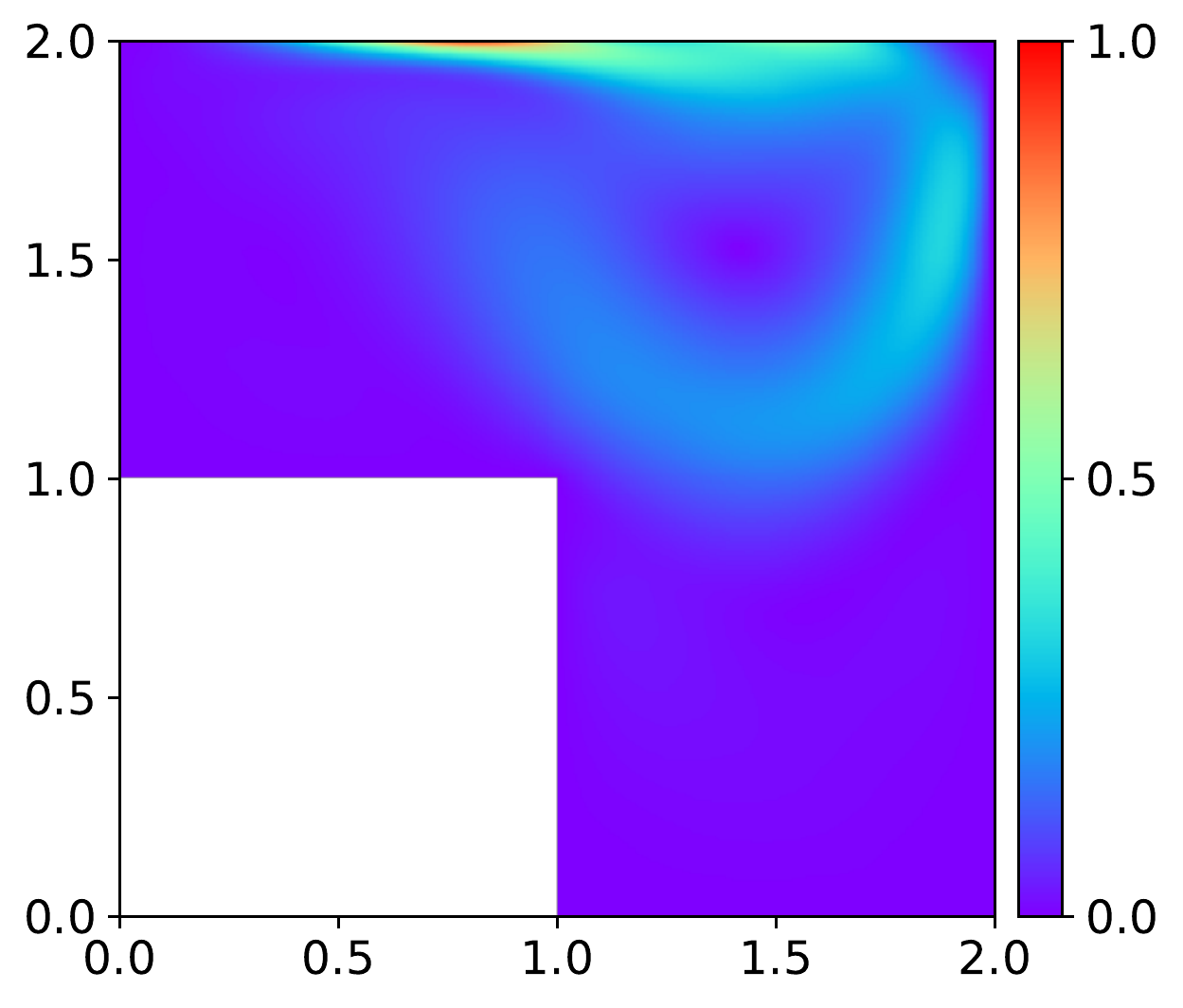}};
\path[fill=gray!70!] 
(0.52,0.40) rectangle +(1.81, 1.81);
\end{tikzpicture}

%% file: fig/step_vel_err_mfpbc.tex
\begin{tikzpicture}
\node[anchor=south west,inner sep=0] at (0.0,0.0) {\includegraphics[width=0.306\textwidth]{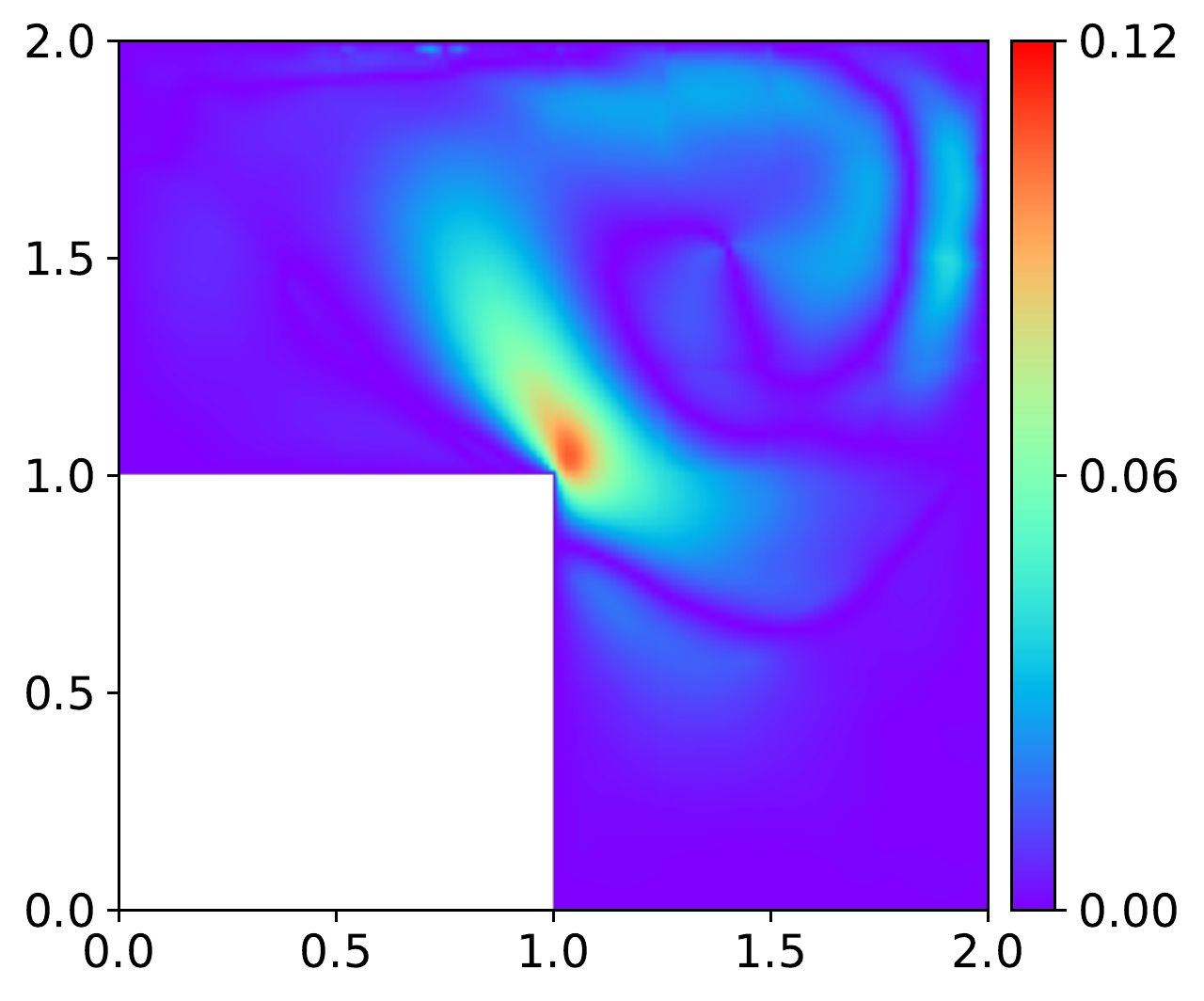}};
\path[fill=gray!70!] 
(0.52,0.40) rectangle +(1.81, 1.81);
\end{tikzpicture}

%% file: fig/step_streamline_mfpbc.tex
\begin{tikzpicture}
\node[anchor=south west,inner sep=0] at (0.0,0.0) {\includegraphics[width=0.28\textwidth]{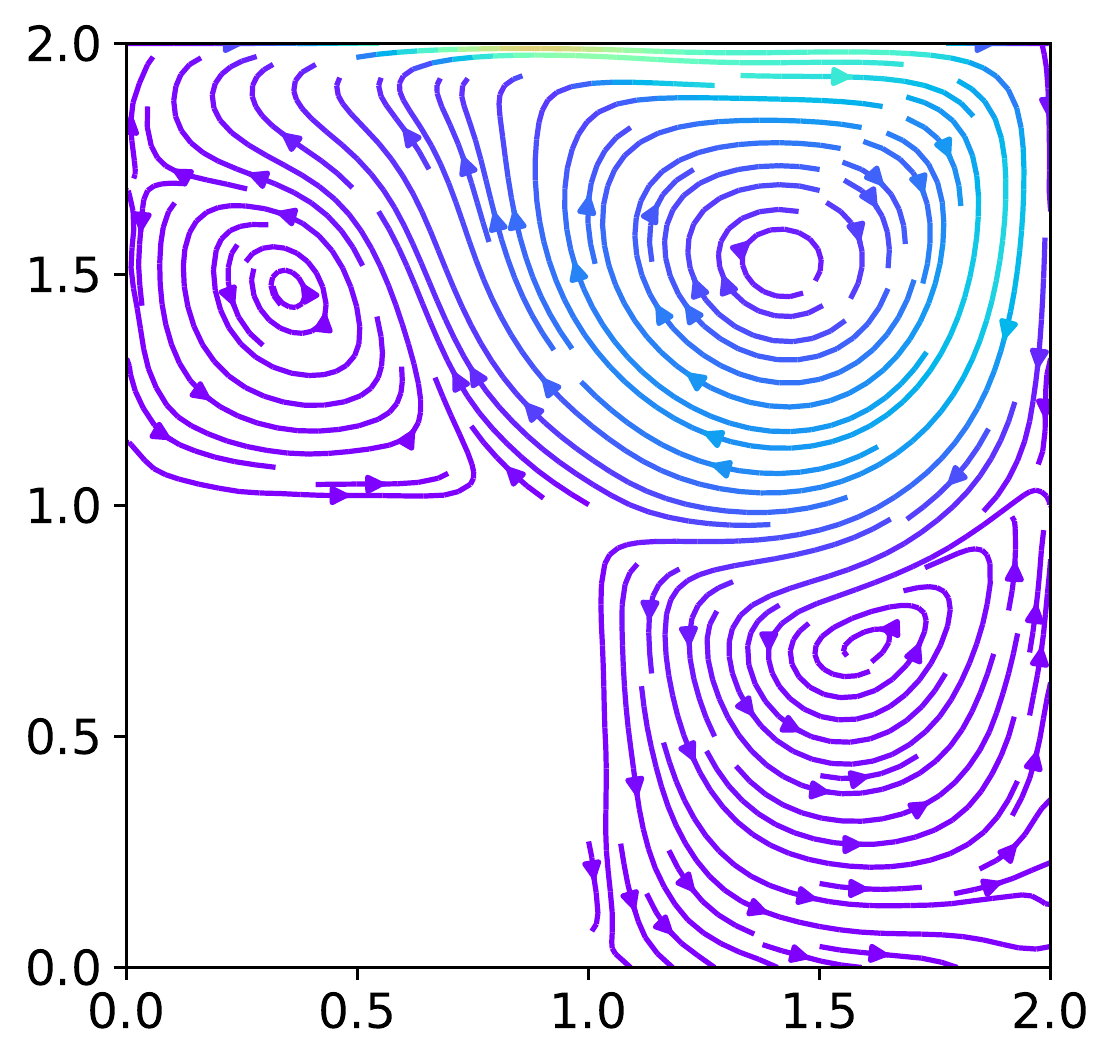}};
\path[fill=gray!70!] 
(0.55,0.40) rectangle +(1.9, 1.9);
\end{tikzpicture}

%% file: conc.tex
\section{Conclusions}\label{sec:conc}
\revise{We demonstrated} that a well-designed and well-trained deep neural network that takes BCs as inputs (GFNet), coupled with a novel iterative algorithm for assembling its predictions (MF predictor) can result in a \emph{transferable} deep learning framework for operator learning. To the best of our knowledge, such a framework is the first of its kind.
The main advantage of this framework is that the model, GFNet, needs to be trained only once and can then be used forever without re-training to solve boundary value problems in larger and complex domains with unseen sizes, shapes, and BCs. 

Our framework demonstrates the capability to infer the solution of Laplace and Navier-Stokes equations on domains that are $1200\times$ and $12\times$ larger that the training domains, respectively. Such scalable predictions are achieved without any re-training. Compared with the state-of-the-art PINN for both PDEs, we demonstrate a remarkable 1-3 orders of magnitude speedups while achieving comparable or better accuracy across a range of BCs and domains unseen during training.
\revise{Moreover, we present an in-depth error analysis of our framework, based on which we compare our GFNet with DeepONet and FNO and demonstrate that our model can effectively reduces the assembly error accumulated in MF predictor.
At last but not least, we improve GFNet to exactly reproduce the input BC, which further reduces all types of errors by up to 84\%.}
We anticipate this research will open new directions for physics-informed surrogate modeling in computational sciences and engineering.

More work remains to be done for GFNet to be widely used for a larger class of problems in engineering. 
For instance, our studies employed square genomes which cannot accurately represent curved objects such as airfoils. 
In addition, we only considered steady flows and boundary value problems whose right-hand-side function can be either zero or a periodic function whose period in each direction equals the size of the genome. The latter choice implies that the BVP solution on one genome is invariant to the genome's position in a large domain which is not a valid assumption for PDEs with space-dependent forcing terms. 
Moreover, the training of GFNet in this paper is a compromise between accuracy and limited time and GPU resources we can access. 
We observed a 20\% improvement in the accuracy of GFNet for the Laplace equation when we trained the model using all the $32\times32$ data points.
However, this will dramatically increase the training cost and prohibit us from fine-tuning the model with 18000 samples in a reasonable amount of time. 
To further improve MF predictor's flexibility and successfully apply our approach to complex geometries and unsteady processes, we can embed boundary coordinates, time, and initial conditions as inputs.
However, such a large input layer will require a GFNet with too many parameters which will increase the training complexities and costs. To consider general non-homogeneous PDEs we anticipate using multiple GFNets to learn the effect of forcing terms on the solution.
Addressing these challenges and leveraging distributed training/inference is an important future exploration.

\section{Acknowledgement}
\revise{We appreciate the support from National Science Foundation (award numbers 2045322 and OAC-2103708) as well as Advanced Research Projects Agency-Energy (award number DE-AR0001209). We would also like to thank the two anonymous reviewers whose comments greatly helped us in improving the quality of our work.}

%% file: appendix.tex
\appendix
\revise{
\section{Non-smooth Boundary Conditions}\label{sec:Non_Smooth_BC}
In section \ref{sec:Error_Analysis} we elaborated on the different error sources in our framework faces where the two most dominant ones were the generalization and the assembly errors. In those analyses we employed smooth and differentiable BCs since all our models were trained with such BCs.
In this section we quantify the generalization error for non-smooth BCs. We evaluate the accuracy of the models described in section \ref{sec:Error_Analysis} by solving BVPs for the Laplace equation subject to non-smooth BCs. We restrict this evaluation to domains of area 1 to prevent any bias that might arise by the smoothing effect of the Laplacian. We consider 100 non-smooth BCs which are generated with a GP whose kernel is a power exponential with randomly chosen power in the range $[1, 2)$. Five sample BCs are demonstrated in Figure \ref{fig:edge_BC}.

\begin{figure}[!htbp]
	\centering
	\includegraphics[width=.35\textwidth]{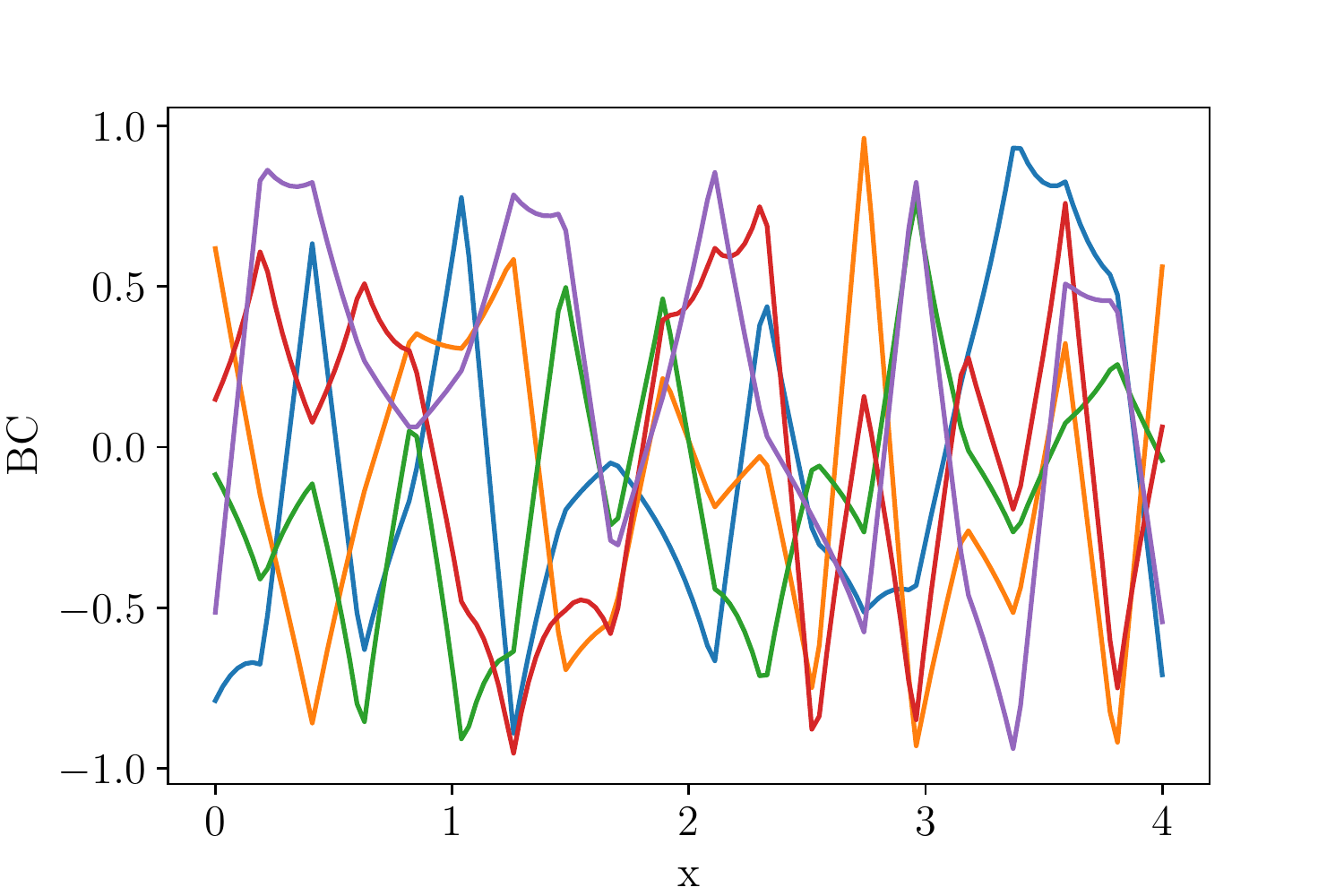}
	\caption{\revise{Sample non-smooth boundary conditions: To quantify the accuracy of the operator learners trained in section \ref{sec:Error_Analysis}, a test dataset with 100 non-smooth BCs are created where the AMG solver is used to solve the Laplace equation.}}
	\label{fig:edge_BC}
\end{figure}

The results are summarized in Table \ref{tab:Non_Standard} and indicate that the accuracy of all models has reduced compared to Table \ref{tab:Operator_Train}). This accuracy reduction is due to the fact that all models have only seen smooth BCs during training. In Table \ref{tab:Non_Standard} the FNO model has the best generalization error and it is followed by the LPFC and DeepONet models. We believe this trend is because FNO predicts the solution for the entire genome at once (rather than a particular spatial location in the genome) which makes it less sensitive to sharp changes in the BCs. In the case of residuals, the LPFC model trained with the PDE loss outperforms the other operator learners. 
A better performance in these cases can be achieved by models that enforce BCs which we have studied for the NS equation in Section \ref{sec:exact_bc}.

\begin{table}[!h]
	\centering
	\vspace{3mm}
	\caption{\revise{Accuracy of different GFNets for non-smooth BCs: 100 random non-smooth BCs are created to evaluate LPFC, LPFC best, FNO, and DeepONet models. We report the test MAE and the residual computed with either FD with a step size of $1/32$ or AD (training MAE and the validation MAE based on smooth BCs are provided in Table \ref{tab:Operator_Train}).}}
	\revise{\input{./tab/Non_Standard.tex}}
	\label{tab:Non_Standard}
\end{table}

\section{Neumann and Robin Boundary condition}
\label{sec:robin_bc}
In this paper we only present tests where the Laplace and NS equations are subject to Dirichlet BCs.
Nonetheless, our framework can also predict PDE solutions with Neumann and Robin BCs for the following two reasons.
First, the Schwarz method that our framework is based on can solve elliptic BVPs with Dirichlet, Neumann, and Robin BCs \cite{xu1992iterative, toselli2004domain, mathew2008domain}. 

Second, in our framework we replace the numerical method used in the Schwarz method to solve genome-wise BVPs with GFNet.
The GFNet that enforces the input BC (Equation \ref{eqn:exact_bc}) can be extended to solve PDE with Neumann and Robin BCs \cite{sukumar2021exact}.
To strictly reproduce the Robin BC, we can follow \cite{sukumar2021exact} and decompose the approximated solution as:
\begin{equation}
    u(\bx) = [1+\phi(c+\pmb{n}\cdot\nabla)]\;\nn(\bx|\hp) - \phi g
    \label{eqn:exact_robin_bc}
\end{equation}
where $\phi(\bx)$ is the signed distance function measuring the distance between $\bx\in\Omega$ to $\partial\Omega$, the underlying Robin BC is $\pmb{n}\cdot\nabla u + cu = g$ at $\partial\Omega$, and $\pmb{n} = -\nabla\phi$ is outward unit normal vector.
When $c=0$, the Robin BC degrades to Neumann BC.
We note that Equation \ref{eqn:exact_robin_bc} simplifies the original formulation in \cite{sukumar2021exact} by omitting the second approximation function (see Equation 25 in \cite{sukumar2021exact}).
We prove that Equation \ref{eqn:exact_robin_bc} satisfies the Robin BC as follow,
\begin{equation}
\begin{split}
\pmb{n}\cdot\nabla u + cu\Big|_{\partial\Omega} &= 
\pmb{n}\cdot\nabla\nn + c\,\nn\pmb{n}\cdot\nabla\phi + 
(\pmb{n}\cdot\nabla\phi)\,\pmb{n}\cdot\nabla\nn - g\,\pmb{n}\cdot\nabla\phi + c\,\nn \\
&=  \pmb{n}\cdot\nabla\nn - c\,\nn - \pmb{n}\cdot\nabla\nn + g + c\,\nn = g.\\
\end{split}
\label{eqn:satisfy_robin}    
\end{equation}
In the above, the first line is based on $\phi=0$ at $\partial\Omega$ and the second line is derived with $\pmb{n}\cdot\nabla\phi = -\pmb{n}\cdot\pmb{n} = -1$.
Despite Equation \ref{eqn:exact_robin_bc}'s complex form, it can be trivially implemented with a neural network.
We refer the readers to \cite{sukumar2021exact} for the details on constructing $\phi(\bx)$.
}

%% file: tab/Non_Standard.tex
  \centering
\begin{tabular}{|l|c|c|c|}

\hline
{Operator learner} & {Test MAE}  & {Residual with FD}  & {Residual with AD }  \\  \hline
	DeepONet & 4.07e-2 & 8.38e+1  &  8.58e+1 \\ 	\hline
	FNO & 7.57e-03  & 1.78e+1 & -  \\\hline
	LPFC & 1.90e-02 & 4.88e+1 & 5.47e+1 \\ \hline
	LPFC Best & 1.88e-02 &  1.70e-2 & 5.4e-3\\ \hline
	\end{tabular} 